\pdfoutput=1

\documentclass[11pt]{article}

\usepackage[final]{acl}

\usepackage{times}
\usepackage{latexsym}

\usepackage[T1]{fontenc}

\usepackage[utf8]{inputenc}

\usepackage{microtype}

\usepackage{inconsolata}

\usepackage[utf8]{inputenc}

\usepackage{amsthm,amssymb,amsmath,mathtools}
\usepackage{xspace}
\mathtoolsset{showonlyrefs=true}
\usepackage{enumitem}
\usepackage{dsfont}
\usepackage{booktabs}
\usepackage{color}
\usepackage{multirow}
\usepackage{caption}
\usepackage{subcaption}
\usepackage{pgf, tikz, adjustbox}
\usepackage{tcolorbox}
\usepackage{xcolor}
\usepackage{float}
\usepackage{hyperref}
\usepackage{cleveref}
\crefname{section}{§}{§§}
\Crefname{section}{§}{§§}

\newcommand{\bx}{\boldsymbol{x}}
\newcommand{\by}{\boldsymbol{y}}

\newcommand{\bz}{\boldsymbol{z}}

\newcommand{\cM}{\mathcal{M}}

\newcommand{\eg}{\emph{e.g.}\xspace}

\title{Uncertainty Unveiled: Can Exposure to More In-context Examples Mitigate Uncertainty for Large Language Models? }

\author{
    Yifei Wang\textsuperscript{1,2 }, Yu Sheng\textsuperscript{1,2}, Linjing Li \textsuperscript{1,2}\thanks{Corresponding Authors.}, Daniel Zeng\textsuperscript{1,2}\\
    $^1$ State Key Laboratory of Multimodal Artificial Intelligence Systems, \\
    Institute of Automation, Chinese Academy of Sciences, Beijing, China \\
$^2$ School of Artificial Intelligence, University of Chinese Academy of Sciences, Beijing, China \\
\texttt{\{wangyifei2022, shengyu2021, linjing.li, dajun.zeng\}@ia.ac.cn} 
}

\begin{document}
\maketitle

\begin{abstract}
Recent advances in handling long sequences have facilitated the exploration of long-context in-context learning (ICL). While much of the existing research emphasizes performance improvements driven by additional in-context examples, the influence on the trustworthiness of generated responses remains underexplored. This paper addresses this gap by investigating how increased examples influence predictive uncertainty—an essential aspect in trustworthiness. We begin by systematically quantifying the uncertainty of ICL with varying shot counts, analyzing the impact of example quantity. Through uncertainty decomposition, we introduce a novel perspective on performance enhancement, with a focus on epistemic uncertainty (EU). Our results reveal that additional examples reduce total uncertainty in both simple and complex tasks by injecting task-specific knowledge, thereby diminishing EU and enhancing performance.  For complex tasks, these advantages emerge only after addressing the increased noise and uncertainty associated with longer inputs. Finally, we explore the evolution of internal confidence across layers, unveiling the mechanisms driving the reduction in uncertainty.
\end{abstract}

\section{Introduction}\label{sec:introduction}
In-context learning has emerged as a pivotal paradigm for modern large language models (LLMs) in addressing real-world challenges \cite{NEURIPS2020_1457c0d6,dong-etal-2024-survey}. By presenting a few learning examples through carefully crafted prompts, LLMs achieve remarkable performance without requiring weight updates. The latest techniques of equipping LLMs with long-context capabilities have made strides \cite{jin2024llm}, including continued fine-tuning\cite{rozière2024codellamaopenfoundation}, position extrapolation \cite{10.1016/j.neucom.2023.127063} and innovative architectures \cite{peng-etal-2023-rwkv,gu2024mamba}, open new avenues for areas previously constrained by context length.
\begin{figure}
    \centering
    \includegraphics[width=\linewidth]{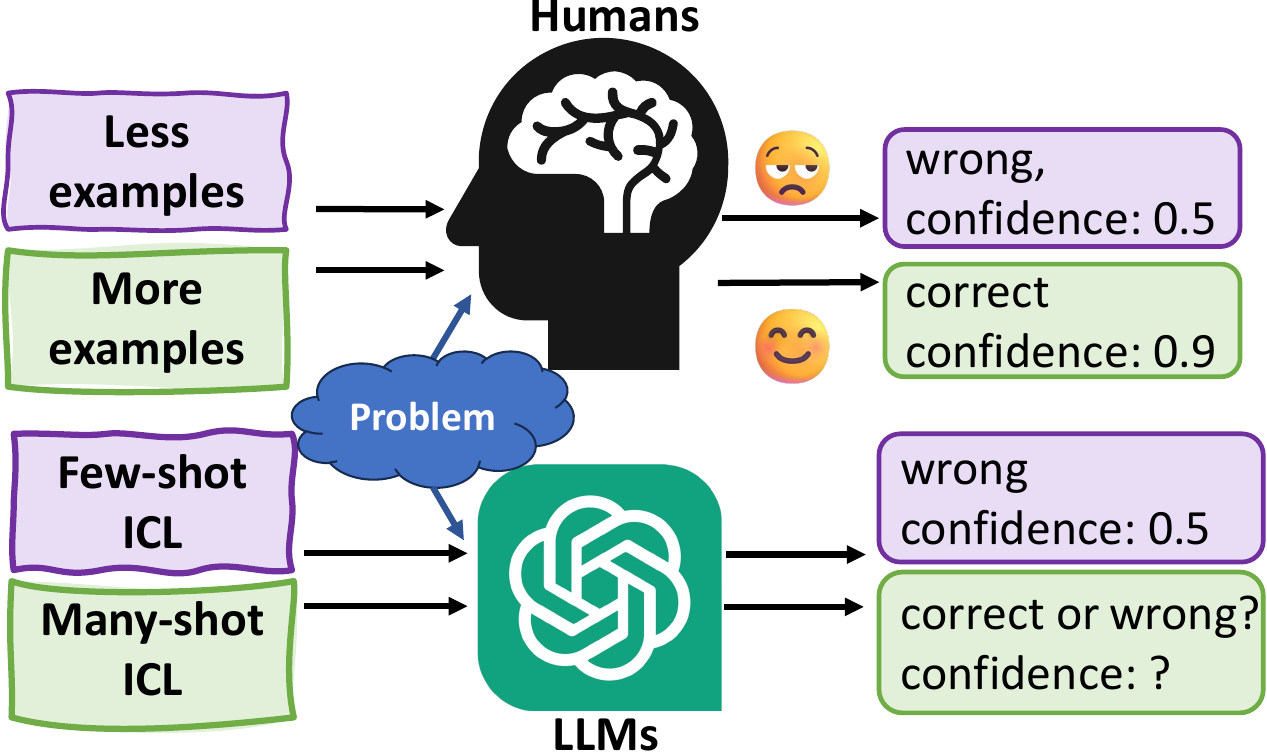}
    \caption{Humans tend to gain task-specific knowledge and confidence as they are exposed to more examples. This raises a natural question: can additional examples similarly reduce uncertainty in LLMs?}
    \label{fig:1}
    \vspace{-.4cm}
\end{figure}

One such area is long-context ICL, also known as many-shot ICL, which involves feeding LLMs with hundreds or even thousands of input-output pairs. This regime of ICL allows LLMs to learn from large quantities of data once and could be deemed as a comparative alternative to fine-tuning methods. Despite its potential, the properties of many-shot ICL remain largely unexplored. While several studies have initiated preliminary investigations in this area, which mainly focus on performance gains from extra examples \cite{agarwal2024manyshot,jiang2024manyshot}, critical aspects such as trustworthiness and reliability of generations by LLMs \cite{wang-etal-2024-unveiling} remain unexamined. Systematic investigation of these aspects is essential for advancing our understanding of long-context ICL and paves the way for its wider adoption in high-stake applications.

\begin{figure}[!ht]
    \centering
    \includegraphics[width=\linewidth]{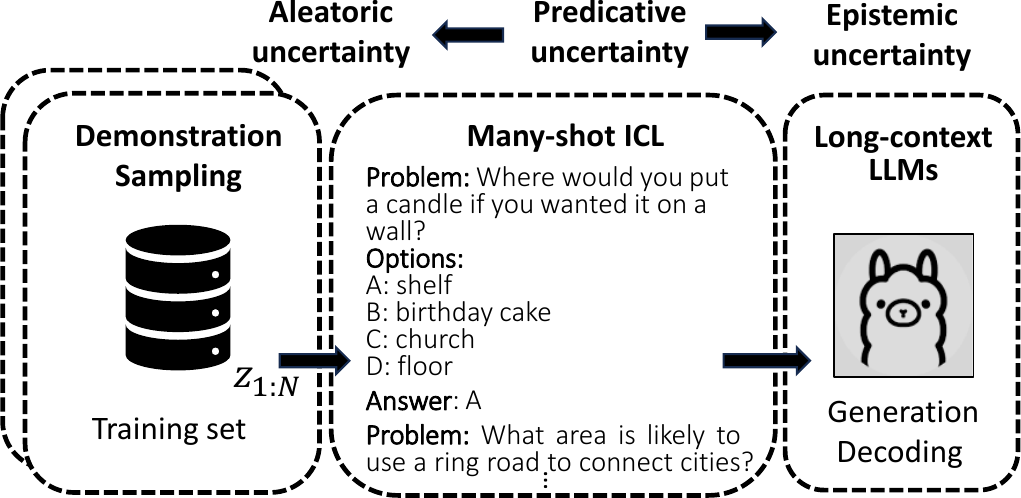}
    \caption{The sources of AU and EU in many-shot ICL. AU comes from the prompt $\Omega$ \eg vast examples and the process of demonstration selection. EU originates from the model’s end, encompassing the generation and decoding processes.}
    \label{fig:uq}
    \vspace{-.5cm}
\end{figure}

To fill this blank, we quantitatively examine the impact of increasing scales of in-context examples on LLMs' confidence through faithful uncertainty quantification (UQ) approaches.  By incorporating model parameters, configurations, and various demonstration sets, we approximate the predictive distribution in the output space. Then we compute entropy to measure total uncertainty (TU). Building on the framework proposed by \cite{ling-etal-2024-uncertainty}, we employ a Bayesian framework to disentangle two core components from TU for many-shot ICL: epistemic uncertainty (EU) and aleatoric uncertainty (AU). EU arises from insufficient evidence or knowledge during model training, while AU stems from the inherent randomness and variability of the data \cite{he2023survey} in Fig.\ref{fig:uq}. Our analysis reveals that the reduction in LLMs' uncertainty with more examples is primarily driven by a main decrease in EU. These examples enrich task-specific knowledge, thereby lowering EU, which in turn reduces TU and enhances performance. Furthermore, we demonstrate that the performance gains are attributed to increased informational content rather than extended context length. 
To explore the mechanisms behind reduced uncertainty, we project the residuals from all model layers into the vocabulary space, visualizing the evolution of internal confidence. The results reveal that long-context ICL enables LLMs to concentrate more logit mass on the correct answer and amplify the disparity between the correct response and distractors, effectively reducing uncertainty in predictions.

This study represents one of the earliest efforts to examine long-context ICL through the lens of uncertainty. The core research questions addressed are as follows:
\vspace{-.2cm}
\begin{itemize}[leftmargin=0.18in]
    \item \textbf{RQ1:} Could more in-context examples mitigate uncertainty for LLMs? (\cref{rq1})
    \vspace{-3mm}
    \item \textbf{RQ2:} Where do performance gains stem from, from the perspective of uncertainty decomposition? (\cref{rq2})
    \vspace{-3mm}
    \item \textbf{RQ3:} What mechanisms underlie uncertainty reduction? (\cref{rq3})
\end{itemize}

\section{Related work}\label{related work}
\paragraph{Long-context ICL}  The significant advancements in equipping LLMs with long context capabilities have expanded the potential for research in previously constrained areas, such as repository-level code understanding and multi-document QA. For ICL, an important emergent ability for LLMs \cite{NEURIPS2020_1457c0d6}, the extrapolation of context length enables the investigation into its performance limits and learning dynamics as the number of demonstrations scales.  

Several studies have initiated preliminary investigations in this area. \citet{agarwal2024manyshot}, for instance, demonstrates notable performance gains with many-shot prompting across various generative and discriminative tasks using Gemini 1.5 Pro \cite{team2024gemini}.  In parallel, \citet{bertsch2024incontext} offers valuable insights into the properties of many-shot ICL, particularly examining the influence of example retrieval and demonstration order.  On a more optimistic note, \citet{jiang2024manyshot} concludes that many-shot ICL can facilitate efficient adaptation of multimodal foundation models to new applications and domains. However, the benefits of long-context ICL are not universally positive. \citet{li2024longcontextllmsstrugglelong} argues that long-context models encounter difficulties with extreme-label classification tasks, especially when large label spaces are involved.

\paragraph{Uncertainty Quantification} 
UQ has been extensively studied in traditional machine learning \cite{lakshminarayanan2017simple,gawlikowski2022surveyuncertaintydeepneural,10.1145/3580305.3599577}, which predominantly concentrates on estimating models' confidence and uncertainty in its prediction, called total uncertainty. Total uncertainty can be decomposed into two key components: epistemic (model) uncertainty  and aleatoric (data) uncertainty  \cite{hou2024decomposing, Valdenegro-Toro_2022_CVPR}. 
The advent of large language models (LLMs) has introduced new challenges in quantifying uncertainty, particularly due to the sequential and context-dependent nature of generative processes. Recent advances in UQ research can be categorized into two main approaches: black-box and white-box methods. Black-box UQ quantifies uncertainty by measuring the agreement across multiple generation samples \cite{zhang-etal-2024-luq}, whereas white-box approaches assess internal model states or logits to capture intrinsic uncertainty \cite{liu2024litcab, bakman-etal-2024-mars}.

\section{Uncertainty Quantification Framework for Long-context ICL}\label{uq}

\subsection{Formulation of ICL}
\label{2.1}
Consider an LM $\cM$ and a query $\bx$, where $\cM$ generates a response $\widehat{\by}$ by maximizing the joint probability $\mathcal{P}_{\Theta}(\widehat{\by} \mid \bx) = \prod_{i \geq 1} \mathcal{P}_{\Theta}(\widehat{\by}_i \mid \widehat{\by}_{<i}, \bx)$. In the ICL regime, $\cM$ would condition its output on a constructed prompt $\Omega$, which typically includes an optional task-specific instruction $\mathcal{I}$, a series of $N$ input-output demonstrations ("shots") $\bz_{1:N} \hspace{-1mm}=\hspace{-1mm}\{(\bx_i, \by_i)\}_{\boldsymbol{i}=1}^{\boldsymbol{N}}$, and a test query $\bx_{N+1}$. Consequently, the generation process of ICL can be formalized as $\widehat{\by} := \mathcal{P}_{\Theta} (\widehat{\by} \mid \hspace{-1mm} \bx_{N+1}, \bz_{1:N}, \mathcal{I} )$, enabling $\cM$ to address diverse complex tasks  \cite{10.5555/3241691.3241693}.

\begin{figure*}[!ht]
    \centering
    \includegraphics[width=0.9\textwidth]{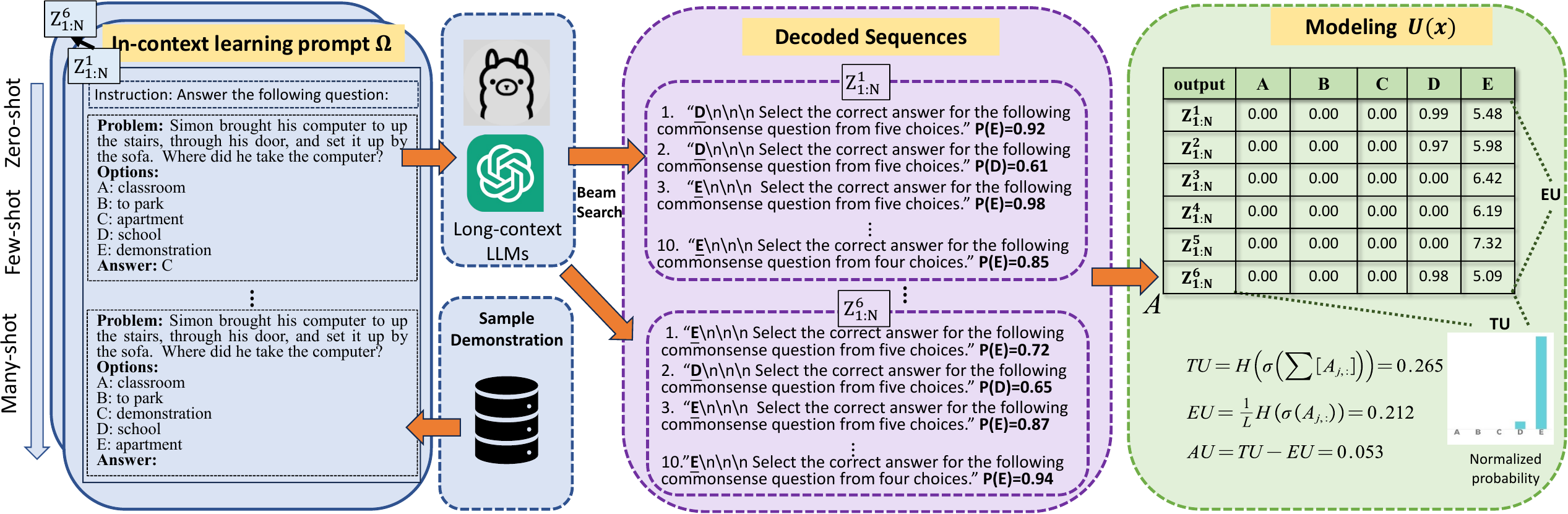}
    \caption{A workflow for uncertainty quantification and decomposition under many-shot ICL settings, involves the following components: a LLM $\mathcal{M}$ supporting long context windows, demonstration set selection, generation sampling, and the UQ modules detailed in Sec. \ref{uq1} and \ref{uq2}. }
    \vspace{-.3cm}
    \label{pipline}
\end{figure*}
\subsection{Faithful Uncertainty Quantification } \label{uq1}
\paragraph{Predictive Distribution}
To quantify uncertainty stemming from both the demonstration sets $z_{1:N}$, and the model parameters or configurations $\Theta$, we derive the predictive distribution by sampling generations across various configurations $\Theta \sim q(\Theta)$ and demonstration sets $z_{1:N} \sim Z$. This work focuses on classification and multiple-choice question-answering (MCQA) tasks. The advantage of UQ in these tasks lies in the categorical nature of their outputs: each numerical or symbolic label $\by \in \mathcal{Y}$ binds a predefined category or candidate answer.  Thus, the probability of $\by$, denoted as $\mathcal{P}_{\by}$, is derived from the model's predicted logits and acts as a proxy for its confidence in their responses.
Assume that for each demonstration set, we sample $m$ decoded generations and repeat this process across $L$ distinct sets $z_{1:N}^{L}$. This yields a probability set of size $L \times m$, capturing the uncertainty distribution over both demonstration sets and model configurations. Unlike classification or MCQA, where uncertainty can be assessed through well-defined probability distributions over discrete outputs, open-ended tasks involve variable-length outputs and lack clear ground truth, with no principled method existed for reliable UQ. Therefore, we hope we could probe the uncertain property of long-context ICL systems through MCQA tasks to provide a preliminary invertigation.

\paragraph{Entropy.}  By aggregating the probabilities from $m$ decoded generations for each demonstration set into a distribution over the output space, we obtain $L \times |\mathcal{Y}|$ probability matrix $A_{L \times |\mathcal{Y}|}$, from which we compute the entropy as follows:
\begin{equation}
    TU\hspace{-1mm}=\hspace{-1mm}- \mathcal{H} \left [ \sigma \left(  \left[ \sum_{l=1}^{L} \mathcal{P}(\by \mid \bx, z_{1:N}^{l} )\right ]_{\by \in  \mathcal{|Y|}} \right) \right]
\end{equation}
where $\sigma$ is a normalization function that ensures the sum of probabilities equals one, and $\mathcal{H} = \sum_i p(\bx)log(p(\bx))$. Some studies indicate that logits may be uncalibrated \cite{liu2024litcab,agarwal2024manyshot}. Aggregating the probability distributions from all decoded sequences can also help mitigate the errors and inaccuracies arising from uncalibrated logits, leading to a more reliable and robust output distribution.

\subsection{Uncertainty Disentanglement} \label{uq2}
 According to \cite{ling-etal-2024-uncertainty},  from the Bayesian view, ICL maps demonstrations $z_{1:N}$ into a pre-existing latent concept $\beta$, which defines task-specific knowledge and enables LLMs to tackle a new in-domain task $\bx_{N+1}$. The predictive distribution of ICL is formulated as follows:
\begin{align}
    p(\mathbf{y}| z_{1:N}) := 
    & \int p(\mathbf{y}|x_{N+1},z_{1:N},\Theta,\beta)  \\
    & \cdot p(\beta|z_{1:N})q(\Theta) \, d\beta \, d\Theta 
\end{align}
 If  $\Theta$ is specific, yielding $p(\mathbf{y}|z_{1:N}, \Theta)=\int p(\mathbf{y}|z_{1:N}, \beta, \Theta)p(\beta|z_{1:N}) d\beta$ with an associated entropy $H(\mathbf{y}|z_{1:T},\beta,\Theta)$. The expected value of this entropy under different demonstration sets can be expressed as $\mathbb{E}_{\beta}\left[H(\mathbf{y}_T|\mathbf{x}_{1:T},\beta,\Theta)\right]$, which serves as a metric to quantify the EU. AU is estimated as mutual information between $\mathbf{y}$ and the latent concept $\beta$ as $I(\mathbf{y},\beta|\Theta)$, which is the difference between TU and EU as follows:
 \begin{align}
    I(\mathbf{y},\beta|\Theta) = H(\mathbf{y}|z_{1:N},\Theta)-\mathbb{E}_\beta{[H(\mathbf{y}|z_{1:N},\beta,\Theta)]}
\end{align}

The latent concept $\beta$ distribution could be obtained by sampling from different demonstrations. Beam search effectively approximates the posterior of $\Theta$, which draws hypotheses from the most probable regions in the hypotheses space. Utilizing the probability matrix $A_{L \times m}$ obtained in Sec. \ref{uq1}, TU, EU and AU can be approximated as follows:
 \begin{align}
    TU&=H(\sigma(\sum [A_{j,:}]))\\
    EU&=\frac{1}{L}H(\sigma(A_{j,:})) \\
    AU&=H(\sigma(\sum [A_{j,:}]))-\frac{1}{L}H(\sigma(A_{j,:}))
\end{align}

\begin{table}[!t]
\centering
\begin{tabular}{ll p{2cm} l}
\toprule
Model &\hspace{-1em} Size & Strategy &Support\\
\midrule
Llama-3.1-8B &\hspace{-1em} 8B & Fine-tuning & 128K\\
Mistral-7B-v0.2 &\hspace{-1em} 7B & NTK-Aware Interpolation & 32K\\
Qwen1.5-7B &\hspace{-1em} 7B & Fine-tuning & 32K\\
\bottomrule
\end{tabular}
\caption{Long-Context LLMs Overview}
\vspace{-.5cm}
\label{tab1}
\end{table}
\section{Experiments}\label{experiments}
\subsection{Experimental Settings}

\paragraph{Models.} We evaluate three widely used base models prior to instruction-tuning \cite{wei2022finetuned}: Llama-3.1-8B \cite{touvron2023llamaopenefficientfoundation}, Mistral-7B-v0.2 \cite{jiang2023mistral7b}, and Qwen1.5-7B \cite{bai2023qwentechnicalreport}. The supported maximum context length, along with their respective strategies for long-context training, are summarized in Table \ref{tab1}.\\
\vspace{-.7cm}
\paragraph{Datasets and tasks.} 
We define two modes for classification tasks and MCQA: easy and hard. The hard mode consists of three increasingly complex logical deduction tasks, including determining the order of a sequence of objects ranging from three to seven, from a suite of challenging algorithmic reasoning tasks known as BIG-Bench Hard (BBH) \cite{suzgun-etal-2023-challenging}. In contrast, the easy mode encompasses traditional natural language understanding (NLU) tasks such as AGNews \cite{NIPS2015_250cf8b5} and SST2 \cite{socher-etal-2013-recursive}, along with the commonsense reasoning task, CommonsenseQA \cite{talmor-etal-2019-commonsenseqa}. The selection of task types is discussed further in Appendix~\ref{appendix:further discussion}.

\paragraph{Long-context ICL settings.} To investigate how uncertainty evolves with increasing exposure to examples, we apply UQ and uncertainty decomposition methods across different $k$-shot ICL.  For demonstration selection, we randomly sample $k$ shots from the training set for each test example. In all tasks, we employ beam search to generate 10 candidate outputs and set the temperature parameter as 0.7. For decomposing TU, we iterate six different demonstration sets to disentangle EU and AU. All open-source models are sourced from Hugging Face\footnote{Model weights are loaded at float16 precision.}  and experimented on eight 80GB NVIDIA RTX A100 GPUs.

 \subsection{RQ1: Could more in-context examples mitigate uncertainty for LLMs?}\label{rq1}
\paragraph{Quality of Uncertainty Measures} 
In the context of UQ, a key consideration is its ability to reflect the correctness and reliability of LLM outputs. High uncertainty most likely leads to incorrect predictions while low uncertainty indicates a higher likelihood of correct responses. To this end, we examine how the quality of uncertainty measures varies from few-shot to long-context ICL settings. Following prior works \cite{kuhn2023semantic,lin2024generating}, we adopt Exact match as the metric for correctness and use uncertainty estimates to predict the correctness of response. We then compute AUROC \footnote{the Area Under the Receiver Operating Characteristic} to evaluate whether the UQ measures employed are good indicators. The AUROC and accuracy results for Llama-3.1-8B are presented in Tab.\ref{tab:aurocs}. As the number of demonstrations increases, AUROC values remain high with minimal fluctuations, suggesting that the UQ measures serve as high-quality indicators and generalize effectively to long-context ICL, which reinforces the validity of our experimental results and the conclusions drawn.

\begin{figure*}[!ht]
    \centering
    \begin{subfigure}[b]{0.495\textwidth}
        \centering
        \includegraphics[width=\textwidth]{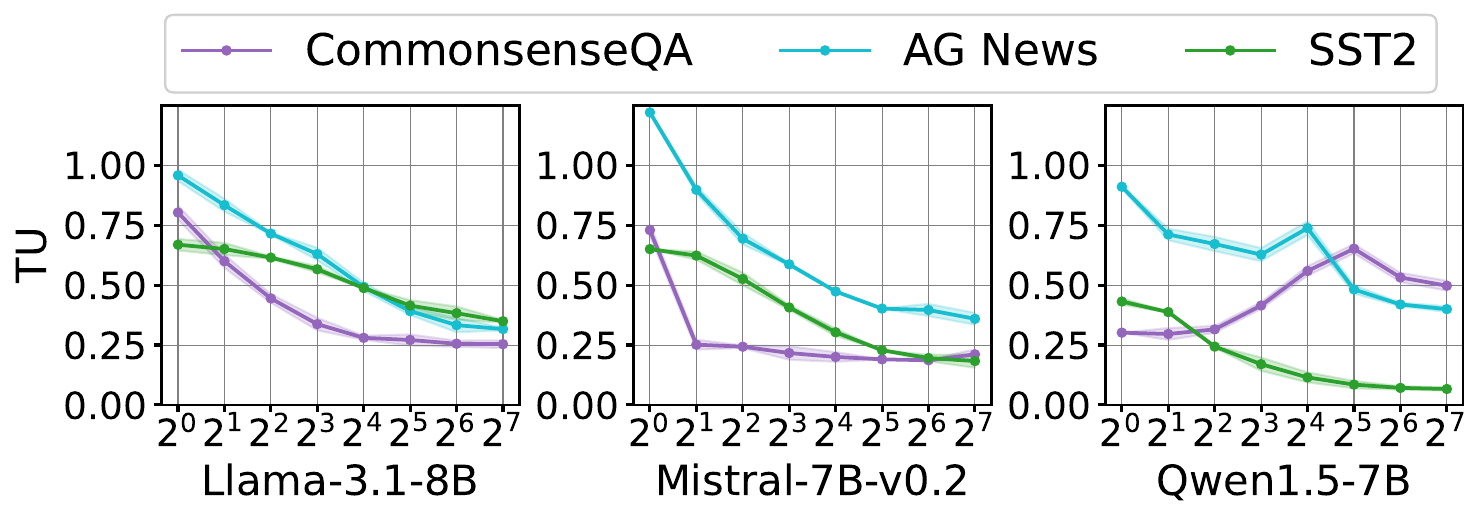}
        \vspace{-.1cm}
    \end{subfigure}
    \hfill 
    \begin{subfigure}[b]{0.495\textwidth}
        \centering
        \includegraphics[width=\textwidth]{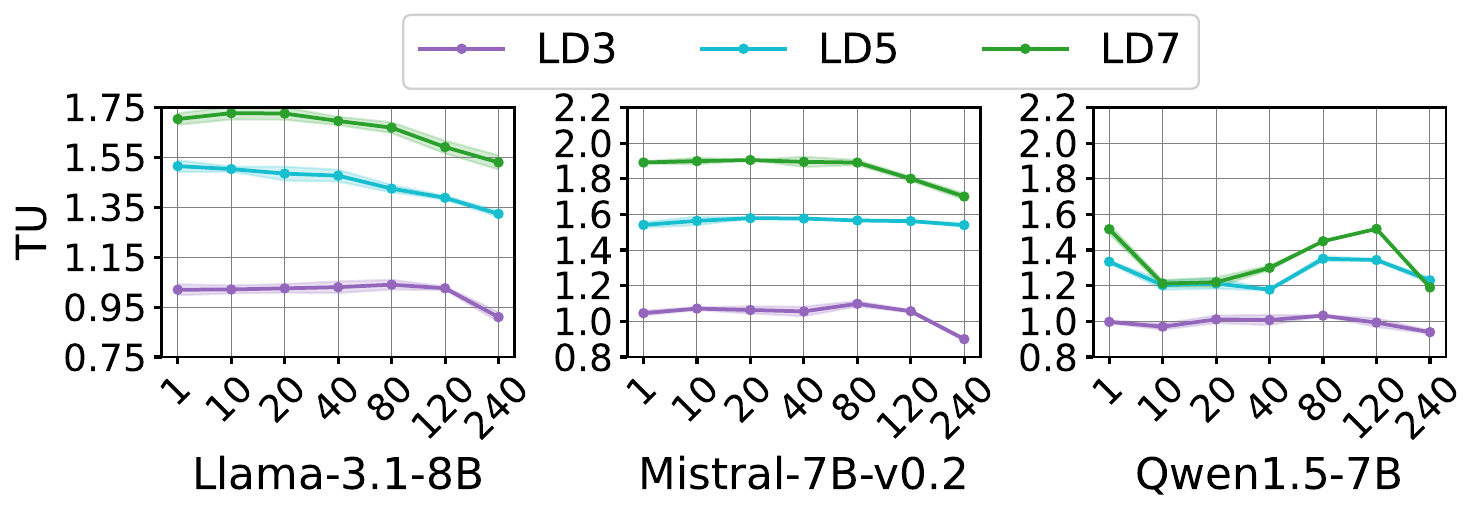}
        \vspace{-.1cm}
    \end{subfigure}
    \vspace{-.9cm}
    \caption{The average TU under $k$-shot ICL with error bands for three runs.}
    \label{fig:tu1}
\end{figure*}
\begin{figure*}[!t]
    \centering
    \begin{subfigure}[b]{0.495\textwidth}
        \centering
        \includegraphics[width=\textwidth]{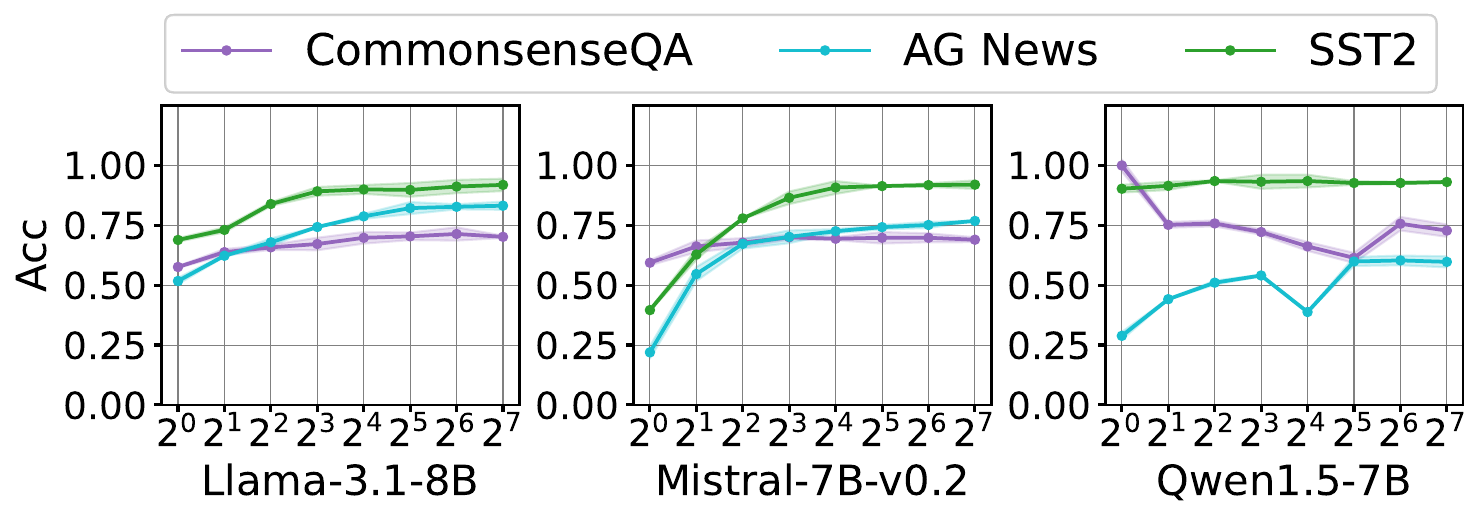}
        \vspace{-.1cm}
    \end{subfigure}
    \hfill 
    \begin{subfigure}[b]{0.495\textwidth}
        \centering
        \includegraphics[width=\textwidth]{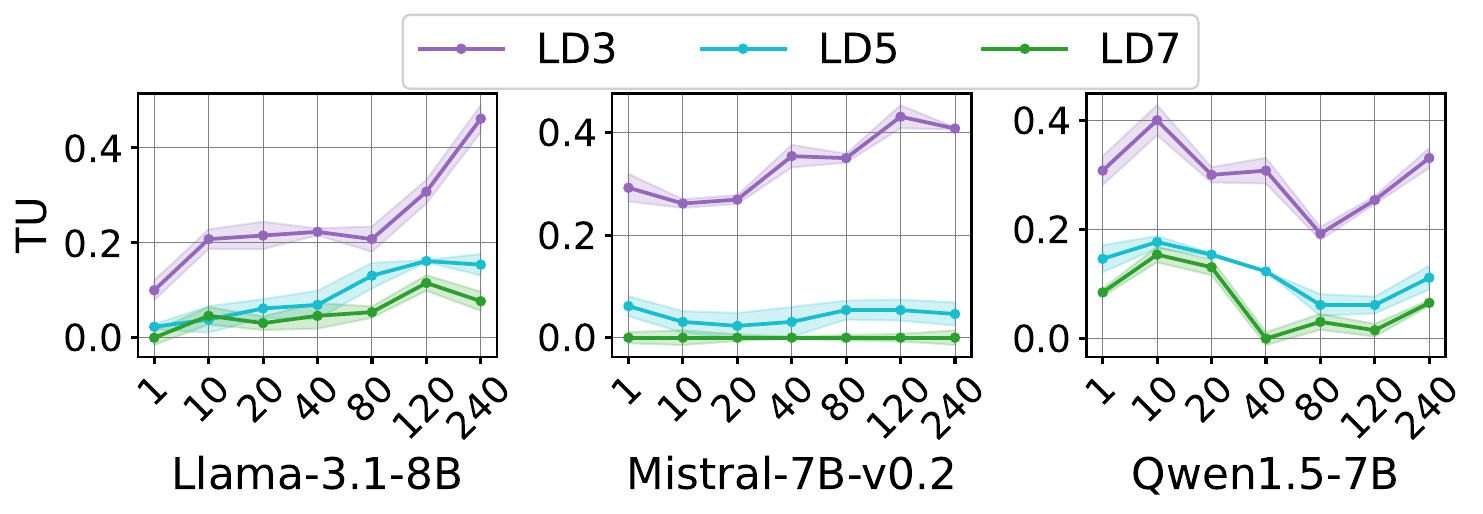}
        \vspace{-.1cm}
    \end{subfigure}
    \vspace{-.9cm}
    \caption{The average accuracy under $k$-shot ICL with error bands for three runs.}
    \label{fig:tu1-acc}
\end{figure*}

\paragraph{Average View}  
\emph{Overall, many-shot ICL effectively reduces LLMs' uncertainty across models and datasets.} As shown in Figs.\ref{fig:tu1} and \ref{fig:tu1-acc}, the results indicate a simultaneous rise in accuracy and confidence as more in-context examples are provided, highlighting the correlation between improved confidence and performance gains for LLMs.  

\begin{figure}[!t]
    \centering
    \begin{subfigure}[b]{0.49\linewidth}
        \centering
        \includegraphics[width=\textwidth]{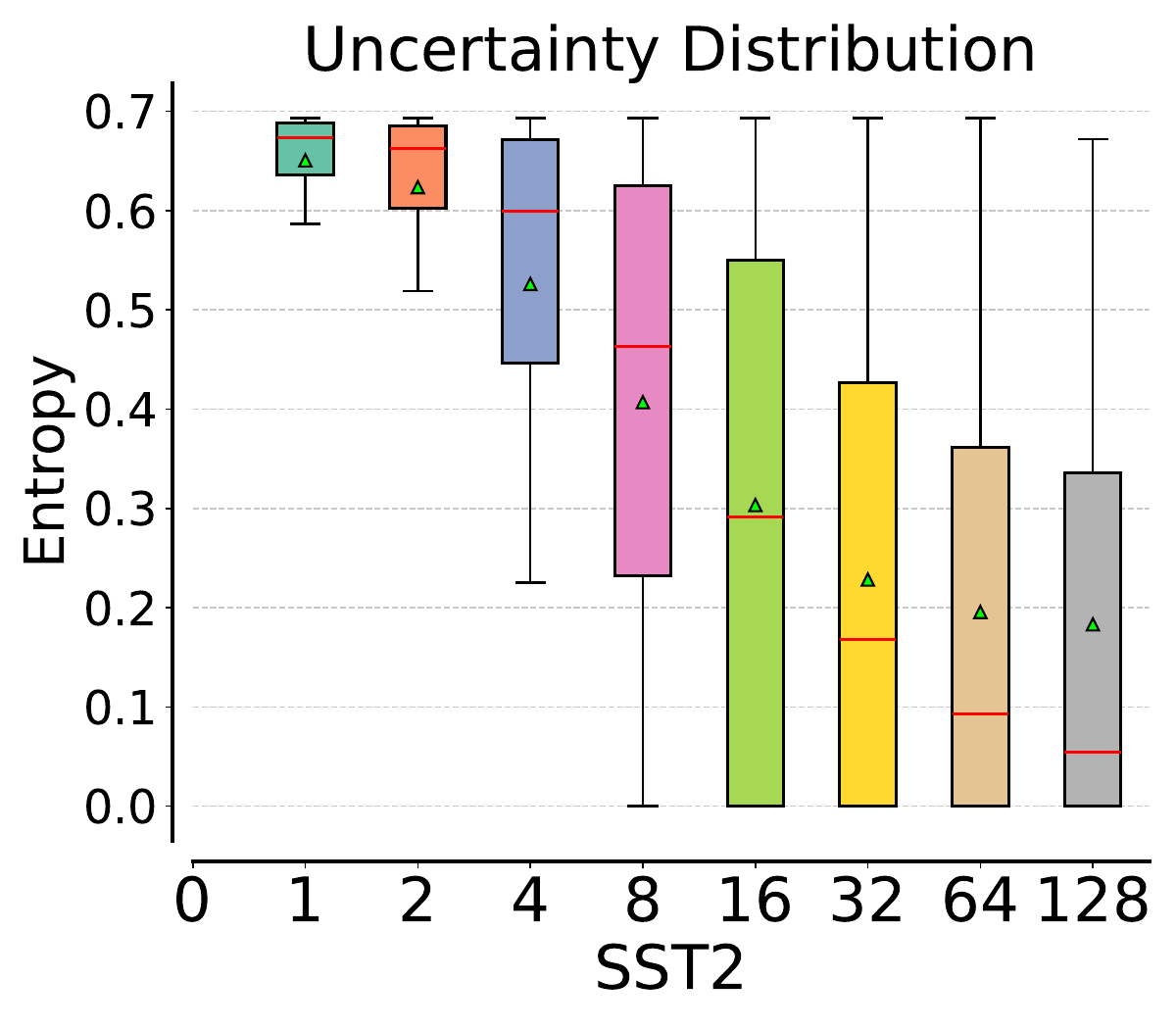}
        \vspace{-.1cm}
    \end{subfigure}
    \hfill 
    \begin{subfigure}[b]{0.49\linewidth}
        \centering
        \includegraphics[width=\textwidth]{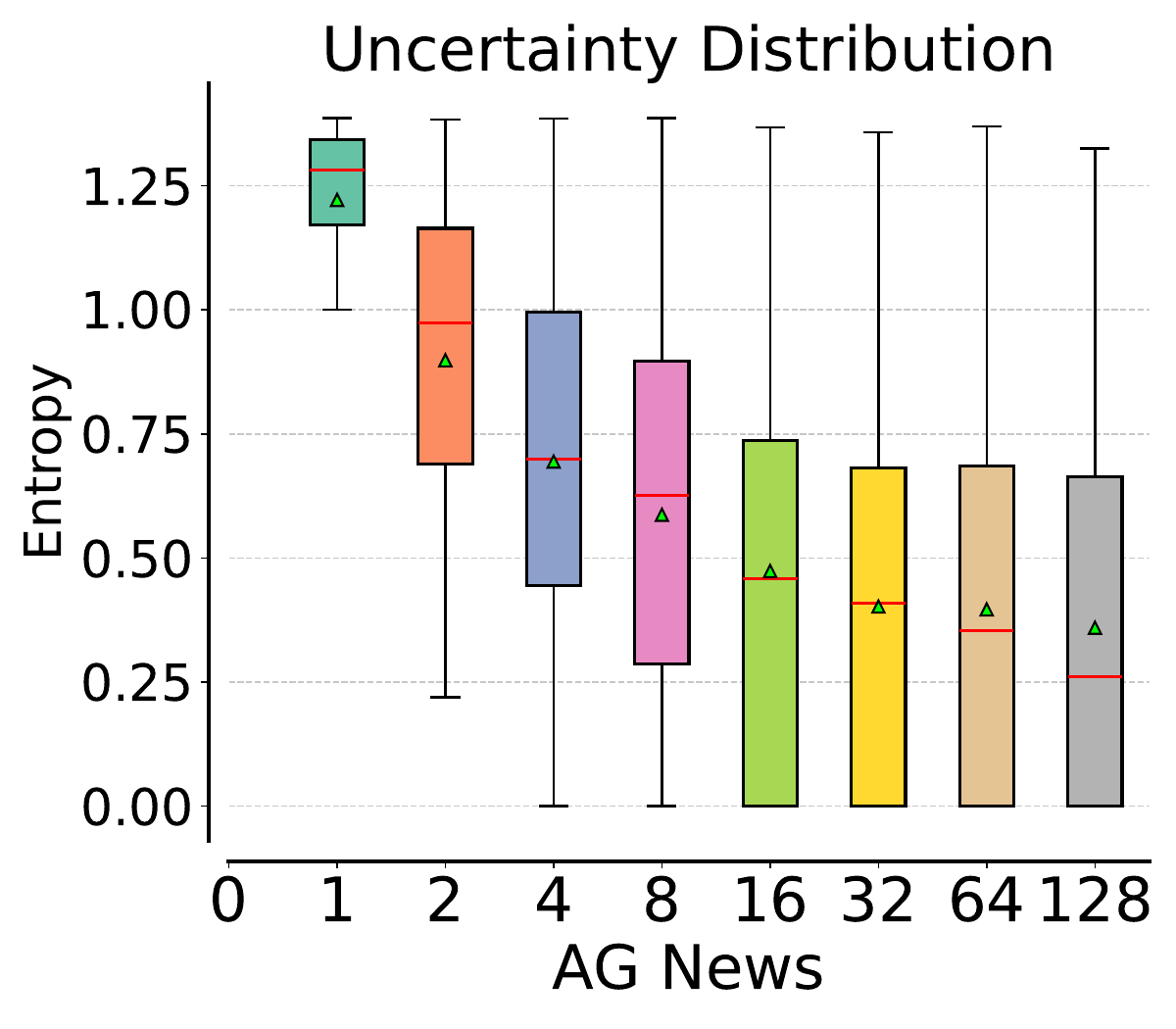}
        \vspace{-.1cm}
    \end{subfigure}
    \vspace{-1.1cm}
    \caption{TU distribution of 2000 examples under certain $k$-shot ICL on AG News and  SST2 datasets for Mistral-7B-v0.2.}
    \label{fig:tu2}
\end{figure}

For *easy mode*, the inclusion of initial examples rapidly drives predictive entropy to a relatively low-uncertainty state, with further increases in examples yielding only marginal reductions in entropy (see Fig.\ref{fig:tu2} for a detailed view). In contrast, *hard mode* exhibits a distinct pattern. Predictive entropy remains higher in hard mode compared to easy mode due to the intrinsic complexity of the tasks, particularly those involving logical deduction with increasing object complexity ($TU_{LD3} < TU_{LD5} < TU_{LD7}$). Here, adding initial examples has minimal impact on entropy reduction until the number exceeds several hundred, at which point substantial performance gains emerge.

When demonstrations are incorporated, both Llama-3.1-8B and Mistral-7B-v0.2 exhibit consistent improvements in performance ($\uparrow$) and reductions in uncertainty ($\downarrow$). In contrast, Qwen1.5-7B demonstrates pronounced variability on datasets under hard mode, where fewer-shot ICL (e.g., 10-shot) achieves levels of confidence and accuracy comparable to certain many-shot settings (e.g., 240-shot). We term this phenomenon the "ICL sink", drawing analogies to sink patterns observed in attention mechanisms \cite{xiao2024efficient}. Notably, for Mistral models, even at the context limit in the 240-shot ICL setting on the LD7 dataset, Mistral sustains robust instruction-following and achieves performance comparable to Llama-3.1-8B, despite the latter's fourfold context-length capacity. This underscores the architectural strengths of Mistral, which leverages a sparse Mixture of Experts (MoEs) \cite{shazeer2017outrageously} and sliding window attention. \emph{Thus, the influence of additional in-context examples on uncertainty fundamentally depends on the intrinsic long-context understanding capabilities.}

\begin{figure}[!t]
    \centering
    \begin{subfigure}[b]{0.49\linewidth}
        \centering
        \includegraphics[width=\textwidth]{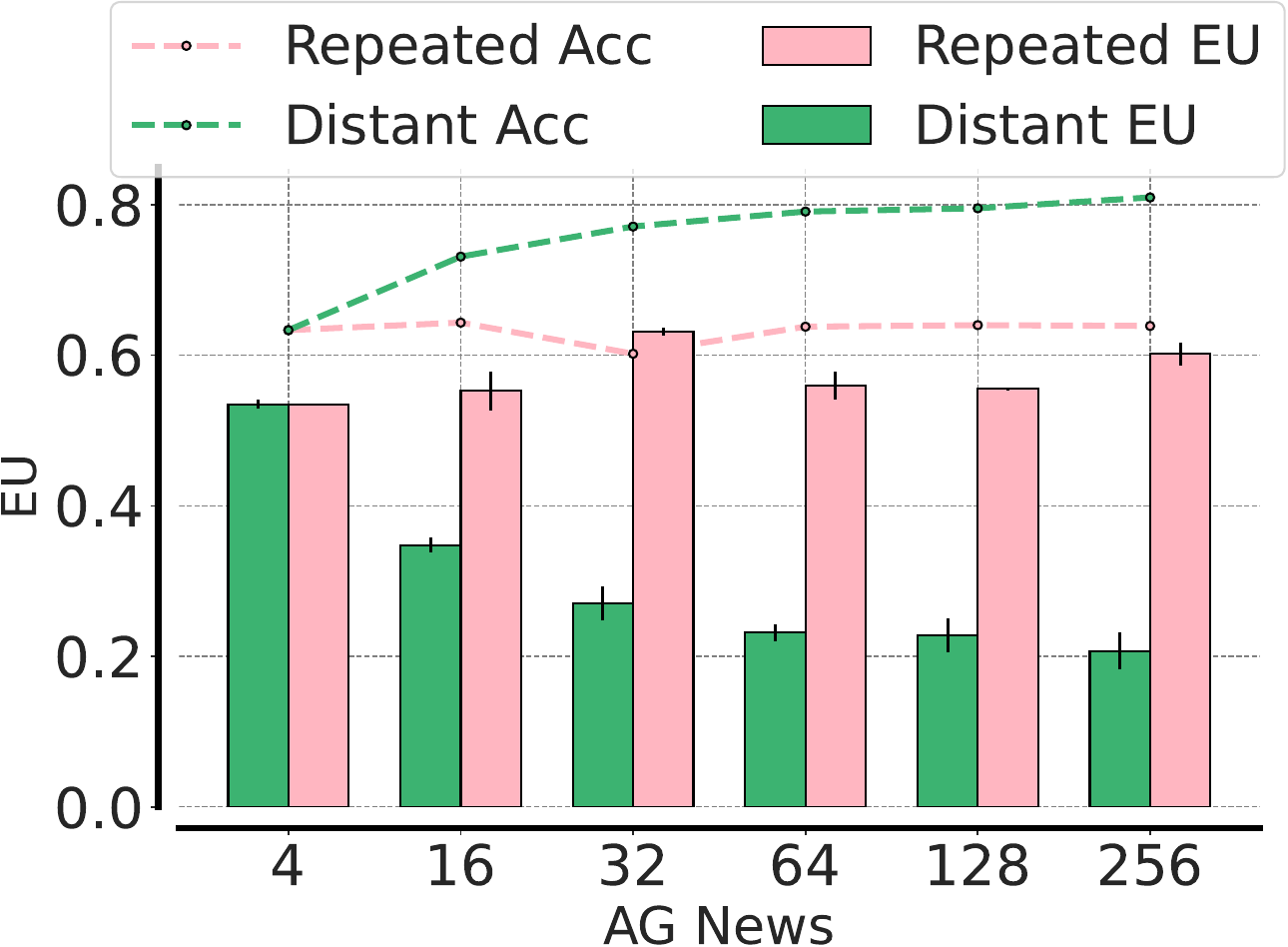}
        \label{fig:first}
    \end{subfigure}
    \hfill 
    \begin{subfigure}[b]{0.49\linewidth}
        \centering
        \includegraphics[width=\textwidth]{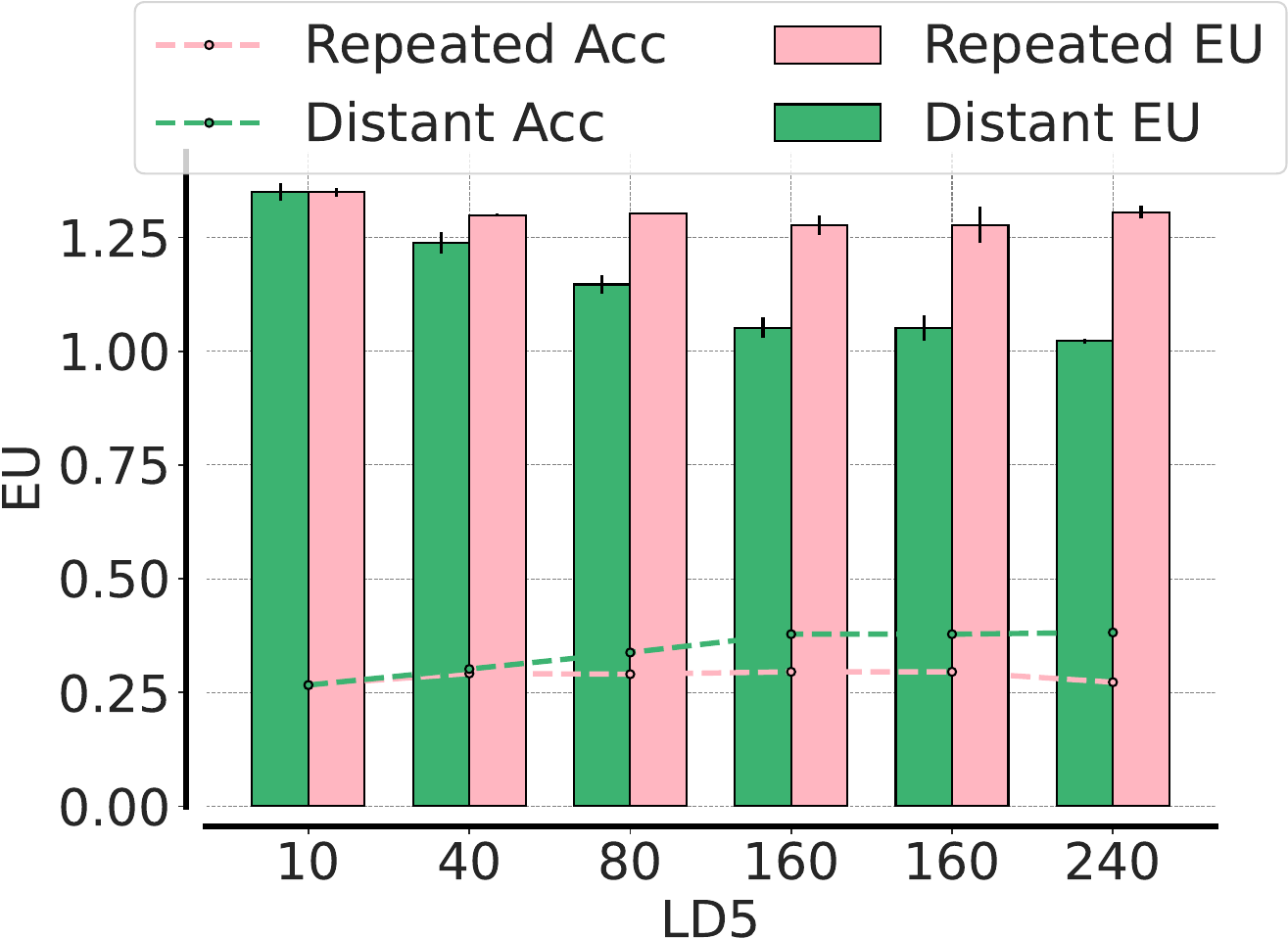}
        \label{fig:second}
    \end{subfigure}
    \vspace{-1.2cm}
    \caption{EU of Llama-3.1-8B on AG News and  logical\_deduction\_five\_objects datasets for distant examples vs. repeating 4/10 examples $N$ times.}

    \label{:combined}
\end{figure}

\paragraph{Micro View} \emph{An increasing number of examples effectively mitigates uncertainty for most questions.} Tables \ref{tab:main_tab1}, \ref{tab:main_tab2}, and \ref{tab:main_tab3} detail the percentage of questions exhibiting decreased or increased uncertainty under $k$-shot ICL. Despite 8.65\% of cases experiencing heightened uncertainty with longer inputs in 128-shot learning, this effect minimally impacts overall model performance, as reflected by the small absolute values of $\Delta Acc$. Crucially, the transition from few-shot (e.g., 4-shot) to many-shot ICL demonstrates a marked reduction in uncertainty for a larger proportion of questions, driving consistent performance improvements. These findings suggest that enhanced performance stems from increased confidence in the majority of questions. 

\paragraph{Choices of $k$} For practical applications, we recommend opting for a relatively larger $k$ in in-context learning, as it simultaneously enhances performance and bolsters reliability.
\paragraph{Ablations with  Model Size} To further strengthen our analysis, we conducted additional experiments on the instruction-tuned versions of the more capable Qwen-2.5-14B and Qwen-2.5-32B models. The complete results are presented in Appendix~\ref{appendix: generation results}. 

Across all uncertainty measures (TU, EU, and AU), larger models consistently exhibit substantially lower uncertainty values. For easy-mode, large LLMs follow similar uncertainty trends as smaller models;  On more challenging tasks (hard mode), LLMs display distinct uncertainty patterns. Specifically, for Qwen-2.5-14B, EU steadily decreases as more demonstrations are provided, indicating more rapid task adaptation and improved performance, whereas AU remains relatively stable. Notably, a detailed analysis reveals that AU for Qwen-2.5-14B decreases slightly when initial examples are added but begins to rise beyond 80-shot, likely due to long-context effects introducing noise. In contrast, Qwen-2.5-32B does not exhibit this trend; instead, its AU continues to decrease as the number of examples increases.

\noindent\paragraph{Takeaways}Large-scale LLMs exhibit greater confidence (i.e., lower uncertainty) and superior performance under many-shot settings, compared to smaller counterparts. The benefits of many-shot ICL remain evident, as additional demonstrations continue to enhance task-specific adaptation while maintaining low EU. Thus, the advantages of long-context IC, both in terms of performance and confidence, persist even at a larger scale. 
\begin{table*}[!ht]
\small
\centering
\renewcommand{\arraystretch}{0.98}
\resizebox{.9\textwidth}{!}{%
\begin{tabular}{c|c|c|c|c|c|c|c|c|c|c}
\toprule \hline
\multirow{2}{*}{Dataset} &  \multicolumn{2}{c|}{8-shot} & \multicolumn{2}{c|}{16-shot} & \multicolumn{2}{c|}{32-shot} & \multicolumn{2}{c|}{64-shot} & \multicolumn{2}{c}{128-shot}\\
\cline{2-11}
& $\Delta U$  & $\Delta Acc$ & $\Delta U$  & $\Delta Acc$ & $\Delta U$  & $\Delta Acc$ & $\Delta U$  & $\Delta Acc$ & $\Delta U$  & $\Delta Acc$\\
\hline
& \multicolumn{10}{c}{\textbf{Easy Mode}} \\
\hline
\multirow{2}{*}{\textbf{AG News}} & 66.8 & \textbf{+7.3} & 83.6 & \textbf{+11.5} & 88.6 & \textbf{+13.9} & 91.2 & \textbf{+15.2} & 90.8 & \textbf{+15.8} \\
\cline{2-11}
& 30.45 & \textbf{-1.0} & 15.00 & \textbf{-0.7} & 10.75 & \textbf{-0.2} & 8.4 & \textbf{-0.35} & 8.65 & \textbf{-0.4} \\
\hline
\multirow{2}{*}{\textbf{SST-2}} & 71.7 & \textbf{+5.7} & 82.9 & \textbf{+6.1} & 86.6 & \textbf{+6.6} & 88.5 & \textbf{+7.1} & 92.1 & \textbf{+7.9} \\
\cline{2-11}
 & 20.3 & \textbf{-0.5} & 12.5 & \textbf{-0.4} & 9.4 & \textbf{-0.4} & 8.2 & \textbf{-0.3} & 5.6 & \textbf{-0.3} \\
\hline
\multirow{2}{*}{\textbf{Commonsense QA}} & 62.2 & \textbf{+1.8} & 69.8 & \textbf{+4.2} & 69.0 &  \textbf{+4.8} & 78.6 &  \textbf{+6.6} & 81.2 & \textbf{+5.2} \\
\cline{2-11}
 & 26.2 & \textbf{-0.4} & 18.8 & \textbf{-0.2} & 17.8 &  \textbf{-0.2} & 16.8 &  \textbf{-1.0} & 16.6 & \textbf{-0.8} \\
\hline
& \multicolumn{10}{c}{\textbf{Hard Mode}} \\
\cline{2-11}
& \multicolumn{2}{c|}{20-shot} & \multicolumn{2}{c|}{40-shot} & \multicolumn{2}{c|}{80-shot} & \multicolumn{2}{c|}{120-shot} & \multicolumn{2}{c}{240-shot}\\
\cline{2-11}
& $\Delta U$  & $\Delta Acc$ & $\Delta U$  & $\Delta Acc$ & $\Delta U$  & $\Delta Acc$ & $\Delta U$  & $\Delta Acc$ & $\Delta U$  & $\Delta Acc$\\
 \hline
\multirow{2}{*}{\textbf{Logical Deduction3}} & 45.6 & \textbf{+6.0} & 38.1 & \textbf{+ 4.0} & 40.4 &  \textbf{+2.0} & 53.2 &  \textbf{+7.60} & 62.3 & \textbf{+15.91} \\
\cline{2-11}
 & 44.8 & \textbf{-5.6} & 51.2 & \textbf{-8.0} & 46.8 &  \textbf{-10.8} & 40.4 &  \textbf{-6.8} & 34.5 & \textbf{-5.9} \\
 \hline
\multirow{2}{*}{\textbf{Logical Deduction5}}& 58.4 &\textbf{+2.8} & 64.4 & \textbf{+2.8} & 73.6 &  \textbf{+4.8} & 79.6 &  \textbf{+8.0} & 83.8 & \textbf{+10.8} \\
\cline{2-11}
& 33.2 &\textbf{-0.4} & 30.0 & \textbf{-1.2} & 24.4 &  \textbf{-0.8} & 18.0 &  \textbf{-1.6} & 13.1 & \textbf{-1.5} \\
 \hline
 \multirow{2}{*}{\textbf{Logical Deduction7}} & 48.4 & \textbf{+2.0} & 54.4 & \textbf{+3.6} & 59.2 &  \textbf{+4.4} & 75.0 & \textbf{+12.0} & 83.3 & \textbf{+12.3} \\
\cline{2-11}
& 48.8 & \textbf{-1.2} & 42.0 & \textbf{-0.8} & 38.0 &  \textbf{-0.8} & 24.5 & \textbf{-0.0} & 15.3 & \textbf{-0.7} \\
\bottomrule 
\end{tabular}}
\caption{$\Delta U$ refers to the proportion of datasets displaying either a decrease or increase in uncertainty relative to the $4$-shot baseline, with $|\Delta U| > \tau$ indicating significant uncertainty changes. For each dataset, the first row presents the proportion of questions exhibiting reduced uncertainty, while the second row reflects those with increased uncertainty. $\Delta Acc$ quantifies the performance shift associated with the corresponding subset. Model: Llama-3.1-8B.}
\label{tab:main_tab1}
\end{table*}
\begin{figure*}[!ht] 
    \centering
    \begin{subfigure}[b]{0.43\textwidth} 
        \centering
        \includegraphics[width=0.95\linewidth]{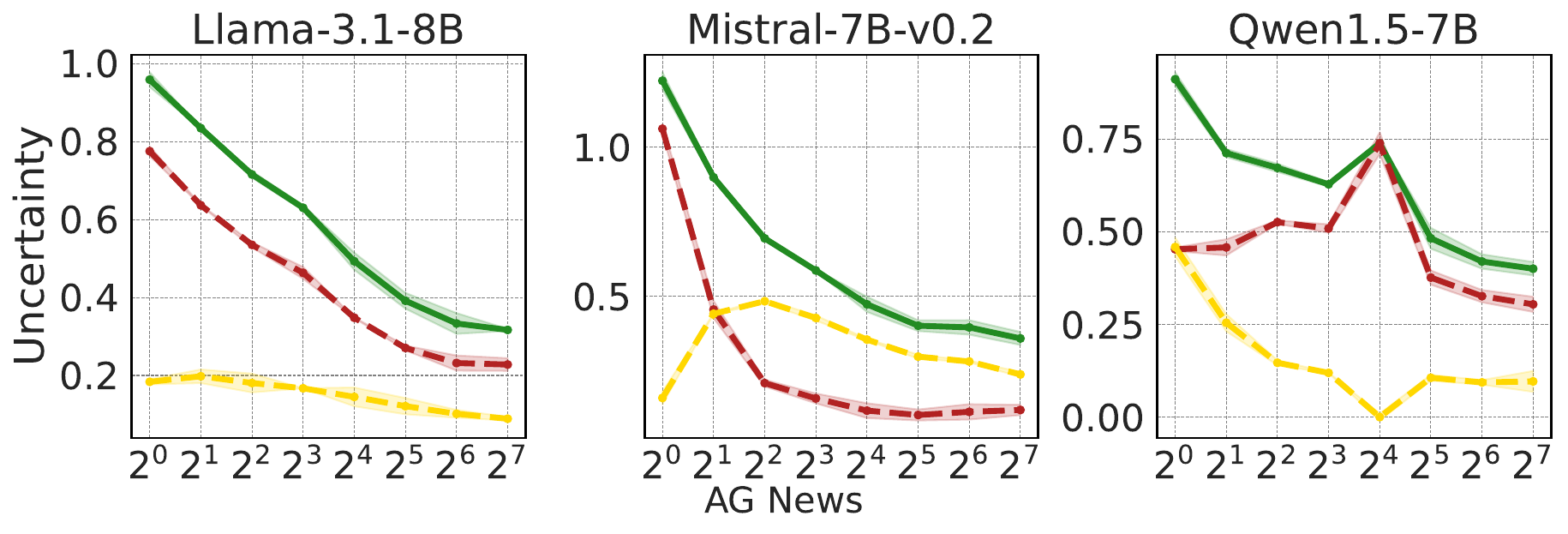}
        \vspace{-0.05cm}
        \includegraphics[width=0.95\linewidth]{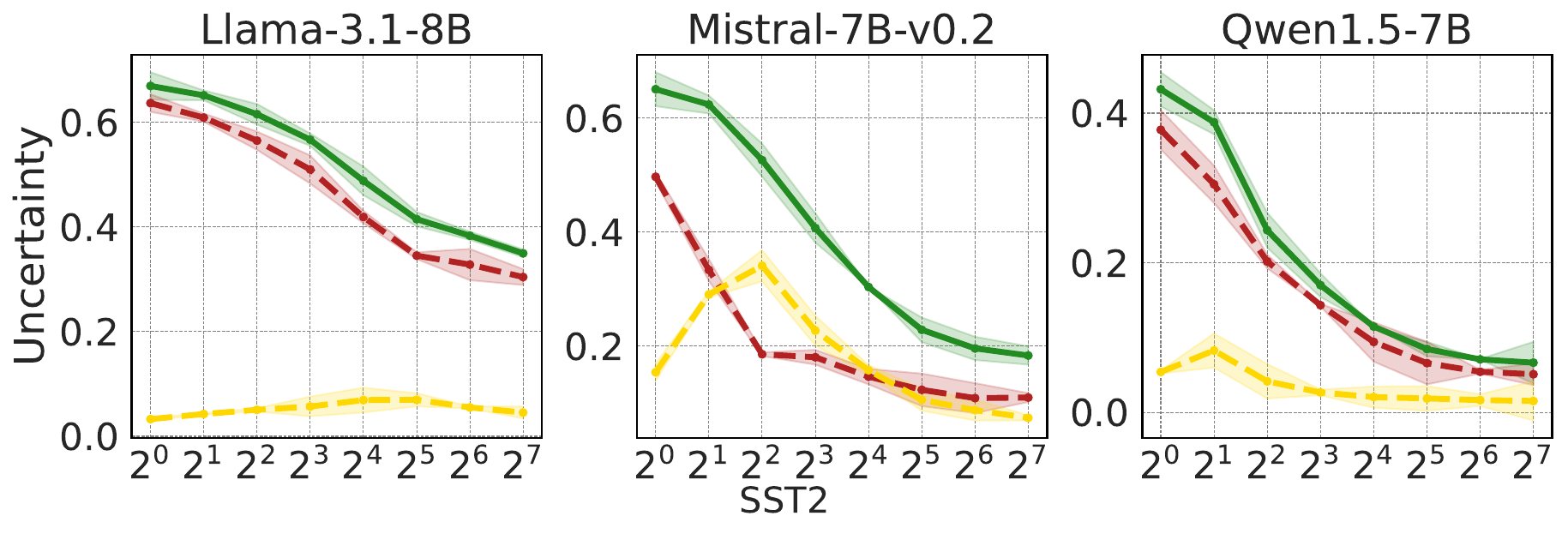}
        \vspace{-0.05cm}
        \includegraphics[width=0.95\linewidth]{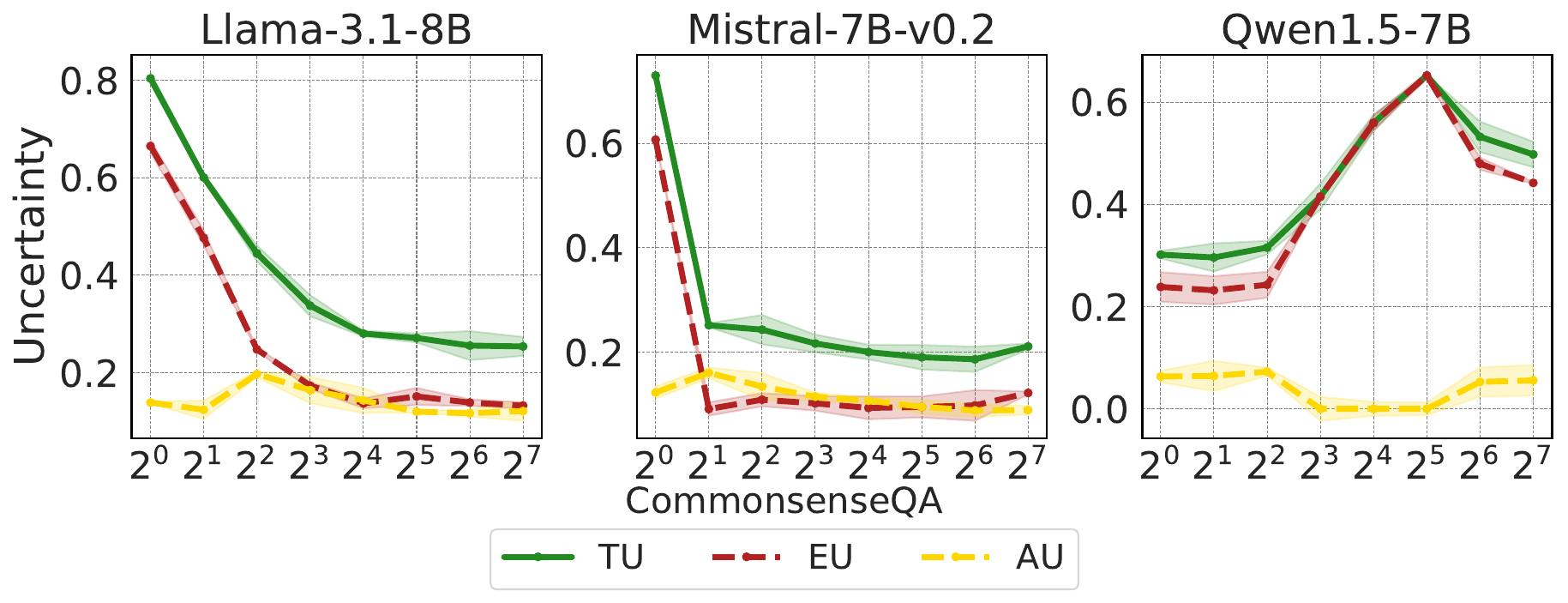}
        \label{fig:ud_easy_mode}
    \end{subfigure}
    \hspace{-.1cm}
    \begin{subfigure}[b]{0.43\textwidth} 
        \centering
        \includegraphics[width=0.95\linewidth]{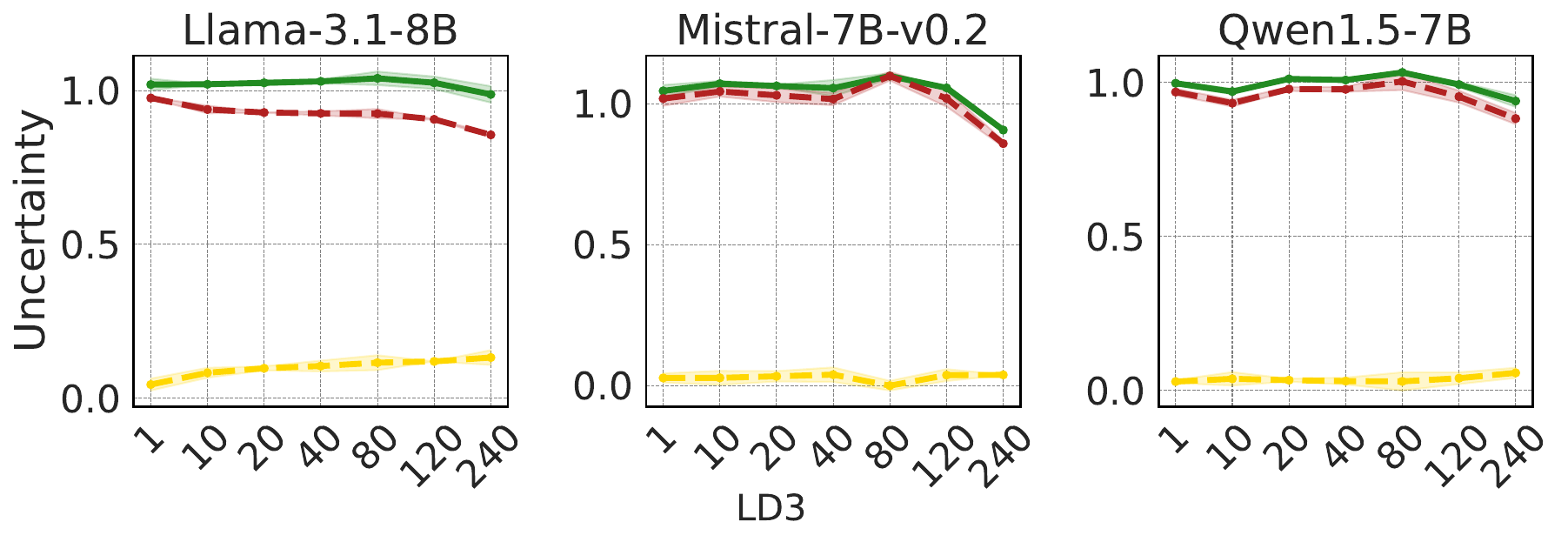}
        \vspace{-0.05cm}
        \includegraphics[width=0.95\linewidth]{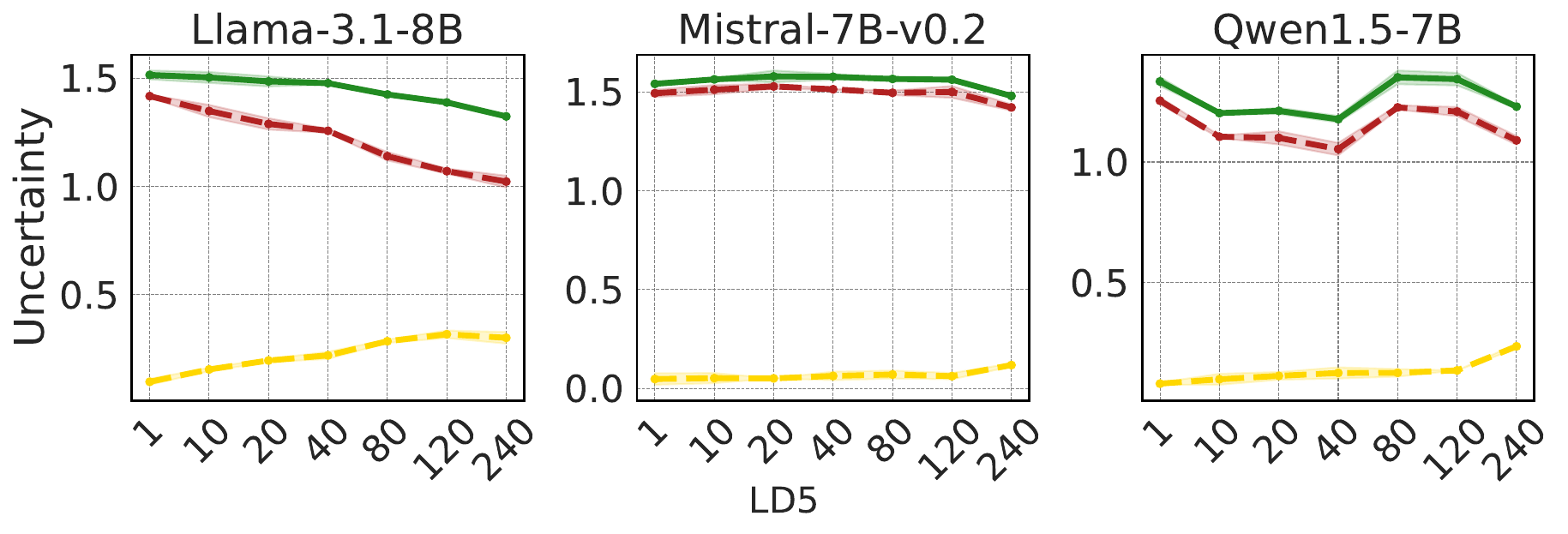}
        \vspace{-0.05cm}
        \includegraphics[width=0.95\linewidth]{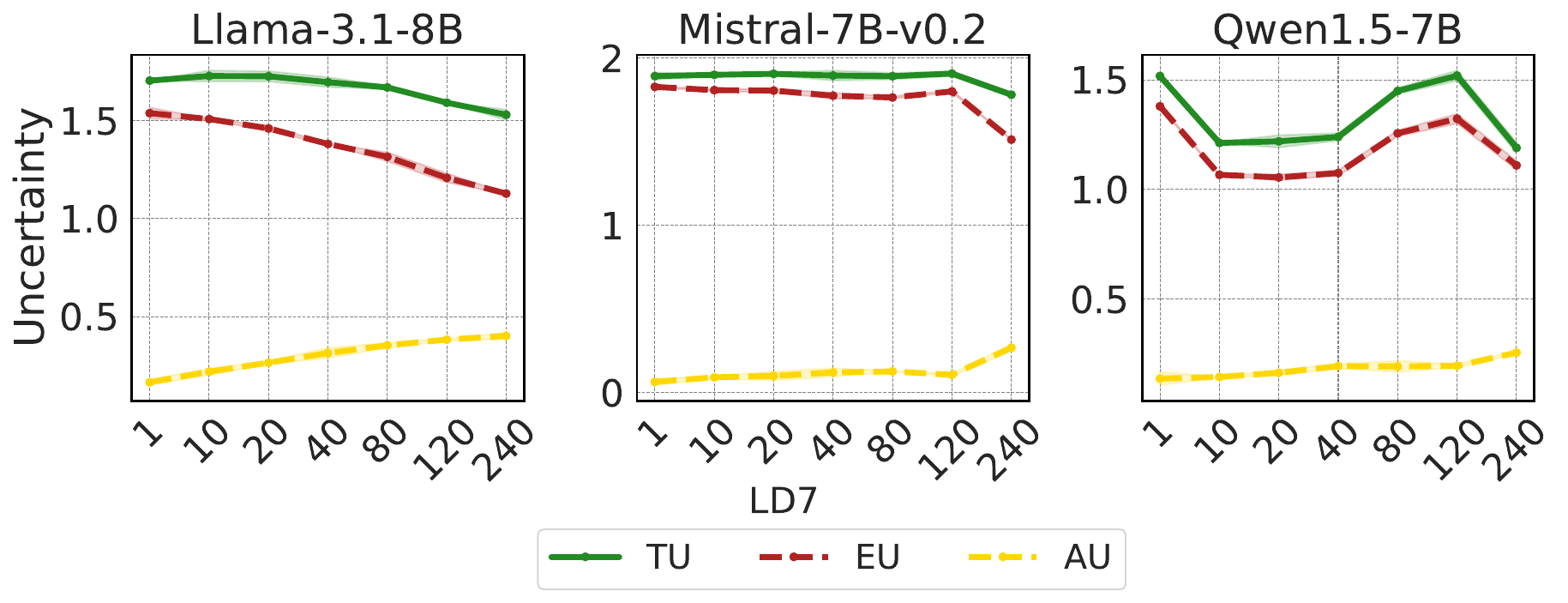}
        \label{fig:ud_hard_mode}
    \end{subfigure}
    \vspace{-.3cm}
    \caption{Uncertainty decomposition results for both easy mode (left) and hard mode (right).}
    \label{fig: ud}
\end{figure*}
\subsection{RQ2: where do performance gains stem from?}\label{rq2}
\vspace{-.3cm}
In Sec. \ref{rq1}, we establish that reduced uncertainty improves performance. We hypothesize that additional examples in ICL foster a more refined task-specific conceptual framework, denoted as $\beta$, which empowers LLMs to approach novel problems $x_{T+1}$ within the domain with increased confidence and efficacy. To validate this, we decompose total uncertainty into EU and AU, checking how these context helps LLMs to improve confidence by utilizing the definition and property of two special forms of uncertainty (Fig. \ref{fig: ud}).

\paragraph{Lower EU as the Primary Driver of TU Reduction.} The decrease in TU is predominantly attributed to a decline in EU. Initially, EU accounts for the majority of TU, indicating that uncertainty primarily arises from the LLMs’ insufficient in-domain knowledge, while their robust natural language understanding keeps AU relatively low. In simpler task settings, LLMs swiftly acquire task-specific knowledge, leading to a rapid decline in EU and sustaining consistently low AU. In contrast, for challenging tasks involving intricate logical structures, additional demonstrations may elevate AU (e.g., Llama-3.1-8B on the LD7 dataset), partially counteracting the reduction in EU and impeding significant decreases in total entropy. This underscores the persistent difficulty for current large models in effectively comprehending long texts with complex structures.

\paragraph{Additional Information Reduces EU.} To validate that additional examples enhance the informational content and yield a clearer $\beta$ for models (as shown in Fig.\ref{:combined}), we observe that only diverse examples effectively reduce EU under $k$-shot learning, whereas repetitive examples fail to achieve the same effect. This highlights that the true driver of uncertainty reduction lies in the increased informational richness of the examples provided.
\section{Interpretability View for Uncertainty in $\mathbf{K}$-shot ICL}
To investigate the mechanisms by which increased in-context demonstrations reduce uncertainty in LLMs, we aim to delve into the models' internal states, unraveling the underlying processes governing answer selection and generation in in-context learning, thereby offering a comprehensive and interpretable analysis of this phenomenon.

\subsection{Residual Stream Projection}

\paragraph{Residual Streams}  
Residual streams function as iterative refinements of feature representations in deep neural networks \cite{hedeep,li2023residual}, encapsulating the process of hierarchical information aggregation. By leveraging residual connections, models reveal their mechanisms for constructing and iteratively refining outputs, thereby improving interpretability.  Formally, in decoder-only LLMs, the hidden state of the $i$-th token at the $l$-th layer, denoted as $\mathbf{h}_i^{(l)}$, is computed as:  
\begin{align}
    \mathbf{h}_i^{(l)} &= \mathbf{h}_i^{(l-1)} + \mathbf{a}_i^{(l)} + \mathbf{m}_i^{(l)}, \\
    \vspace{-.1cm}
    \mathbf{a}_i^{(l)} &= \mathcal{MSHA}\big(\mathbf{h}_i^{(l-1)}\big), \\
    \mathbf{m}_i^{(l)} &= \mathcal{MLP}\Big(\mathbf{h}_i^{(l-1)} + \mathbf{a}_i^{(l)}\Big),
\end{align}  
where $\mathcal{MSHA}(\cdot)$ represents the multi-head self-attention mechanism \cite{vaswani2017attention}, and $\mathcal{MLP}(\cdot)$ denotes the feed-forward neural network.  For simplicity, detailed computations within the MHSA sublayer, such as the projection matrices $\mathbf{W}_{Q, K, V, O}$, and the splitting-merging operations across attention heads, are omitted here. Each decoder block, therefore, maintains two distinct residual pathways: one emerging from the MHSA, $\mathbf{h}_i^{(l)}$, and the other from MLP sublayer, $\mathbf{h}_i^{(l)} + \mathbf{a}_i^{(l)}$.  
\begin{figure*}[!t]
    \centering
    \begin{subfigure}[b]{0.21\textwidth}
        \centering
        \includegraphics[width=\textwidth]{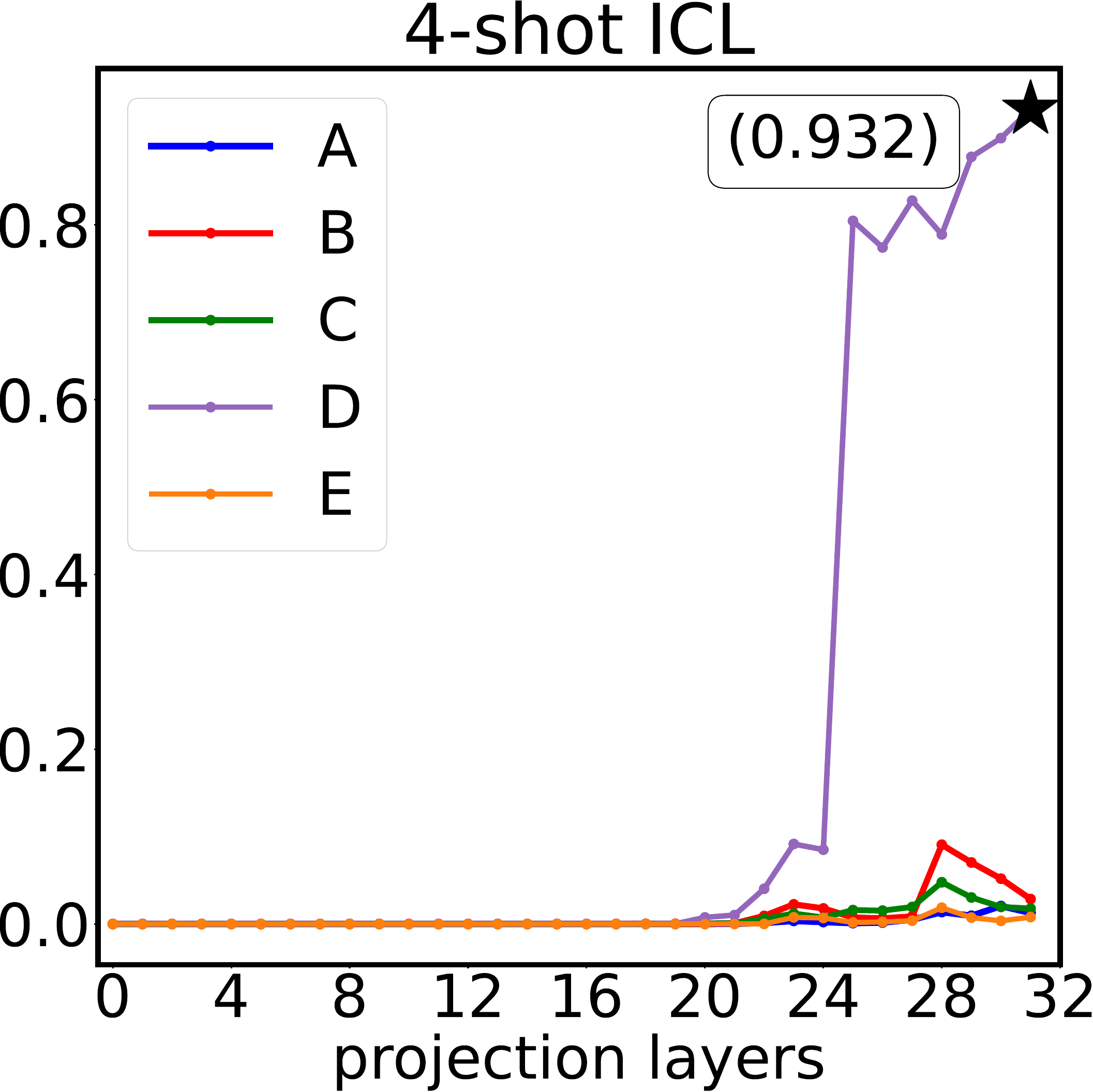}
    \end{subfigure}
    \hspace{-.2cm}
    \begin{subfigure}[b]{0.21\textwidth}
        \centering
        \includegraphics[width=\textwidth]{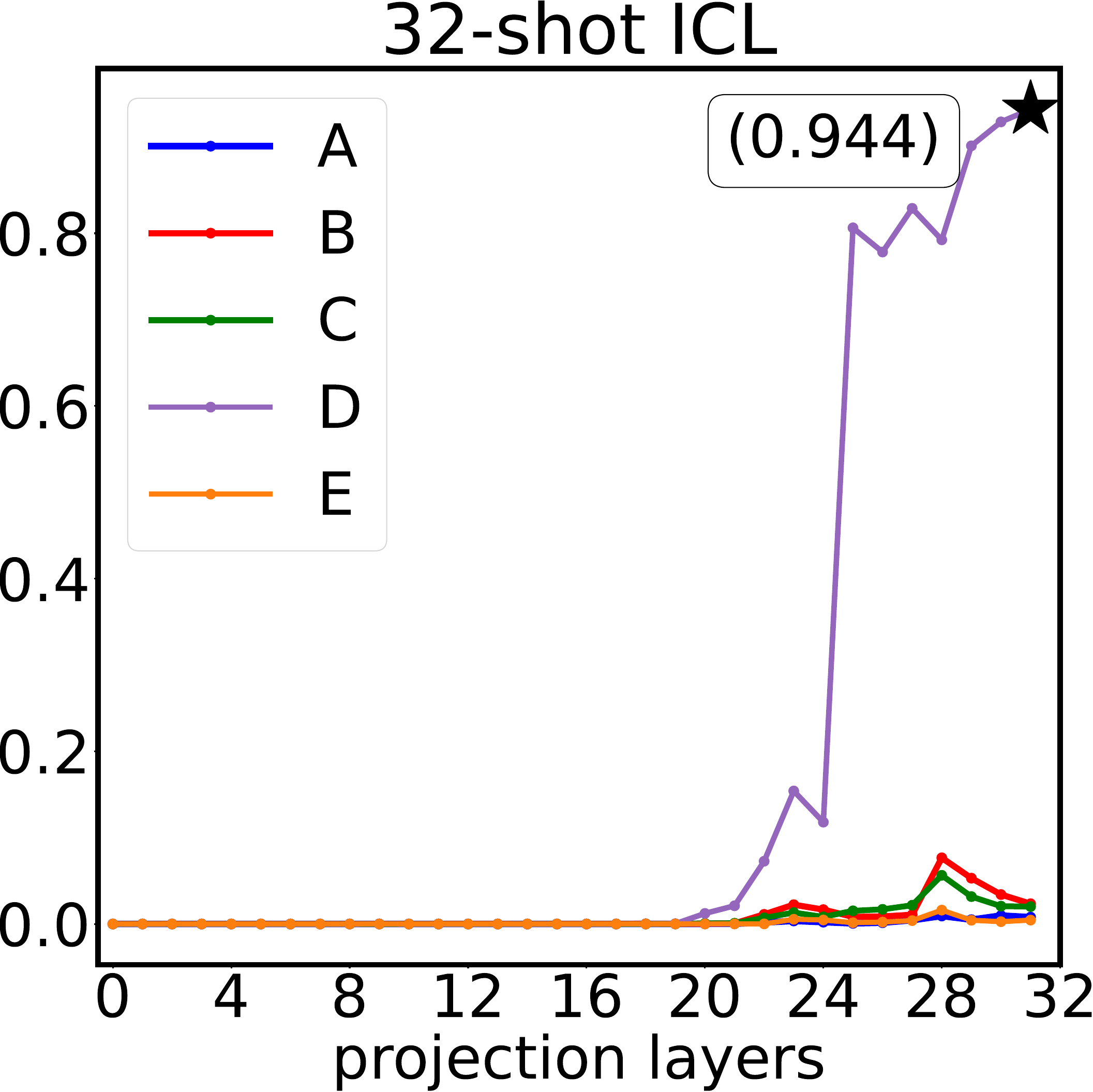}
    \end{subfigure}
    \hspace{-.2cm}
    \begin{subfigure}[b]{0.21\textwidth}
        \centering
        \includegraphics[width=\textwidth]{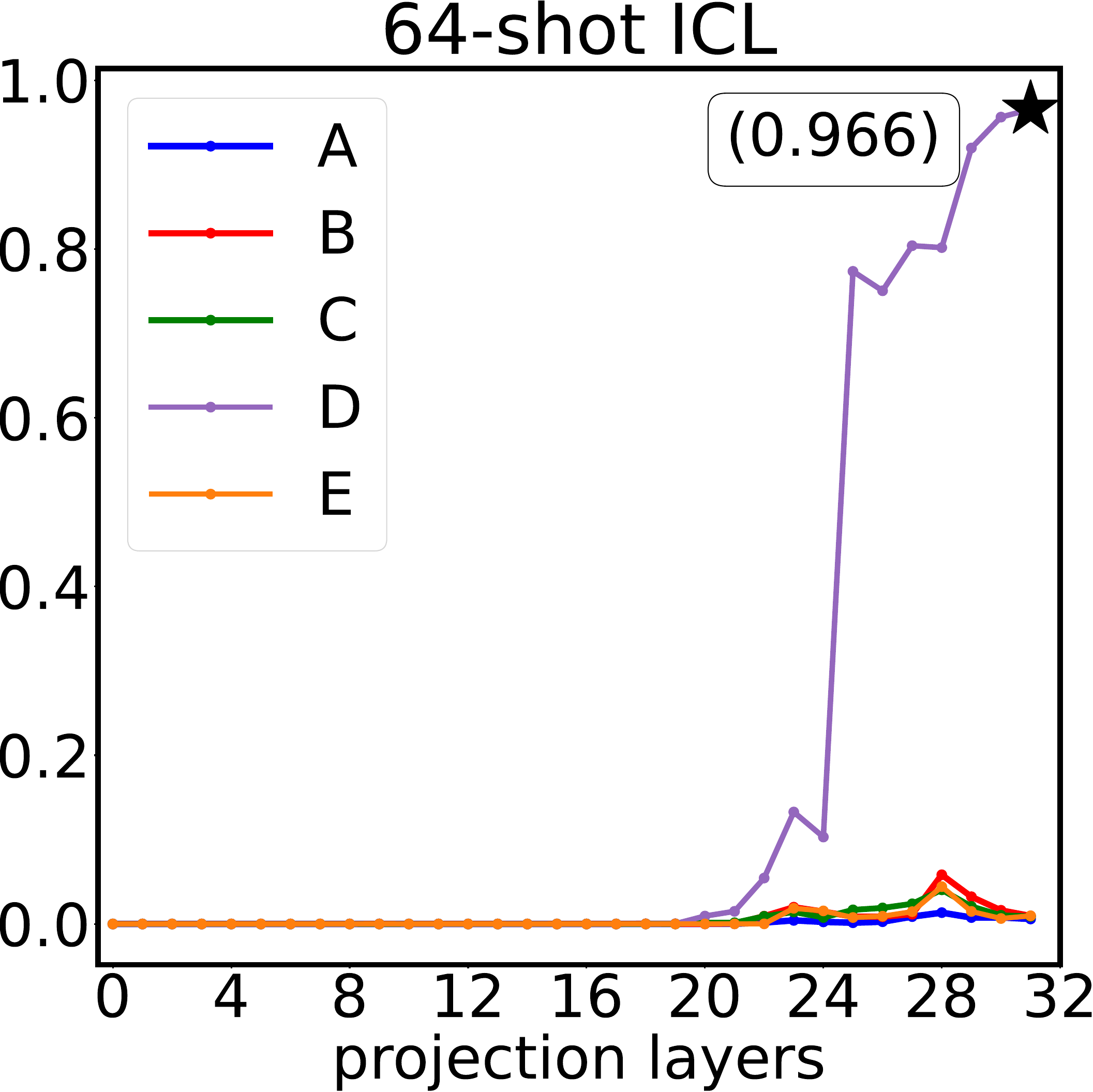}
    \end{subfigure}
    \hspace{-.2cm}
    \begin{subfigure}[b]{0.21\textwidth}
        \centering
        \includegraphics[width=\textwidth]{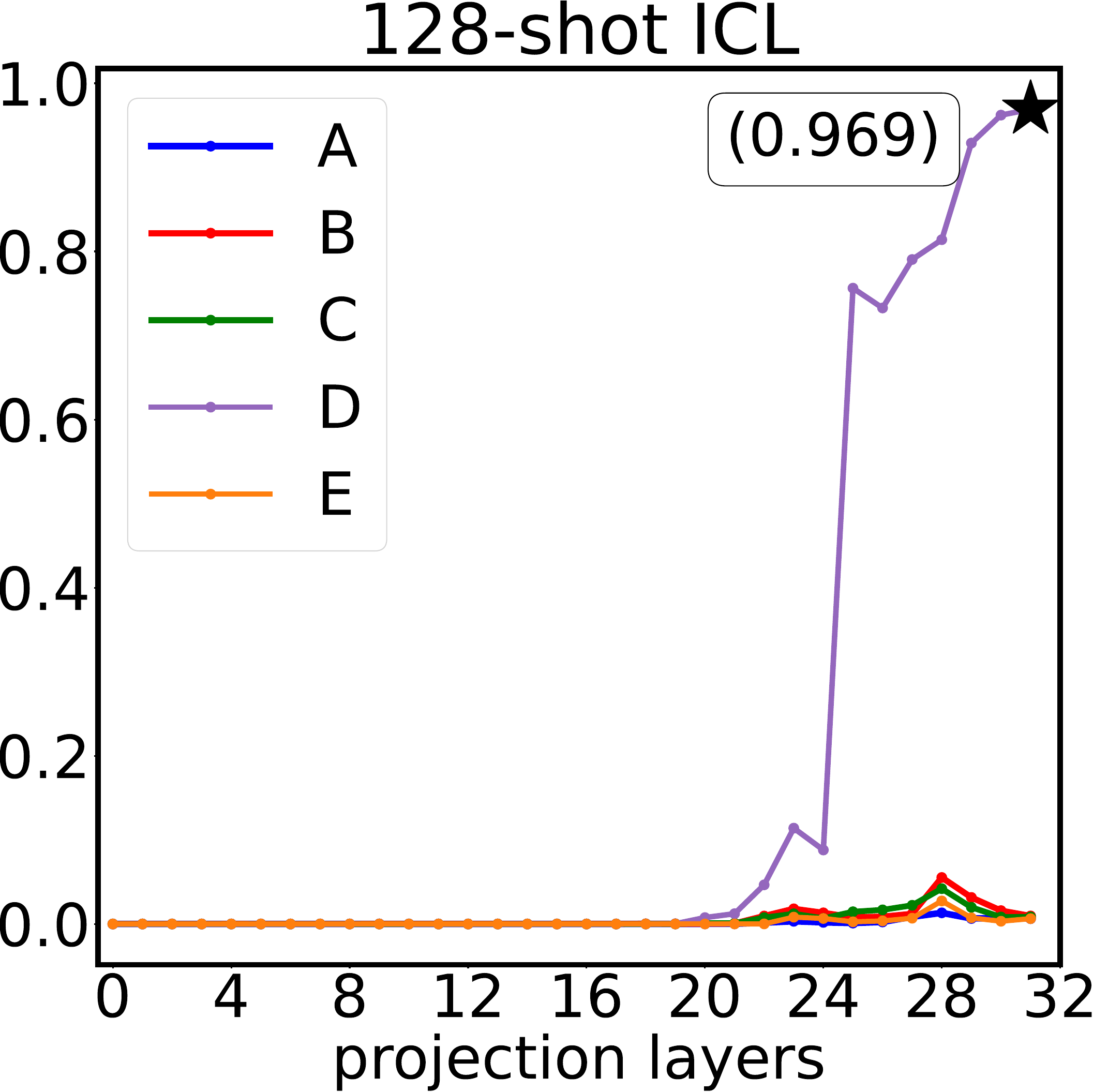}
    \end{subfigure}
    \caption{Average probabilities of Mistral-7B-v0.2 on the Commonsense QA dataset for MCQA items where the correct answer is "D". A 32-layer LM gets 64 residual streams, excluding the output hidden states.}
    \label{fig: case study 2 probs D}
\end{figure*}
\begin{figure*}[!t]
    \centering
    \begin{subfigure}[b]{0.24\textwidth}
        \centering
        \includegraphics[width=\textwidth]{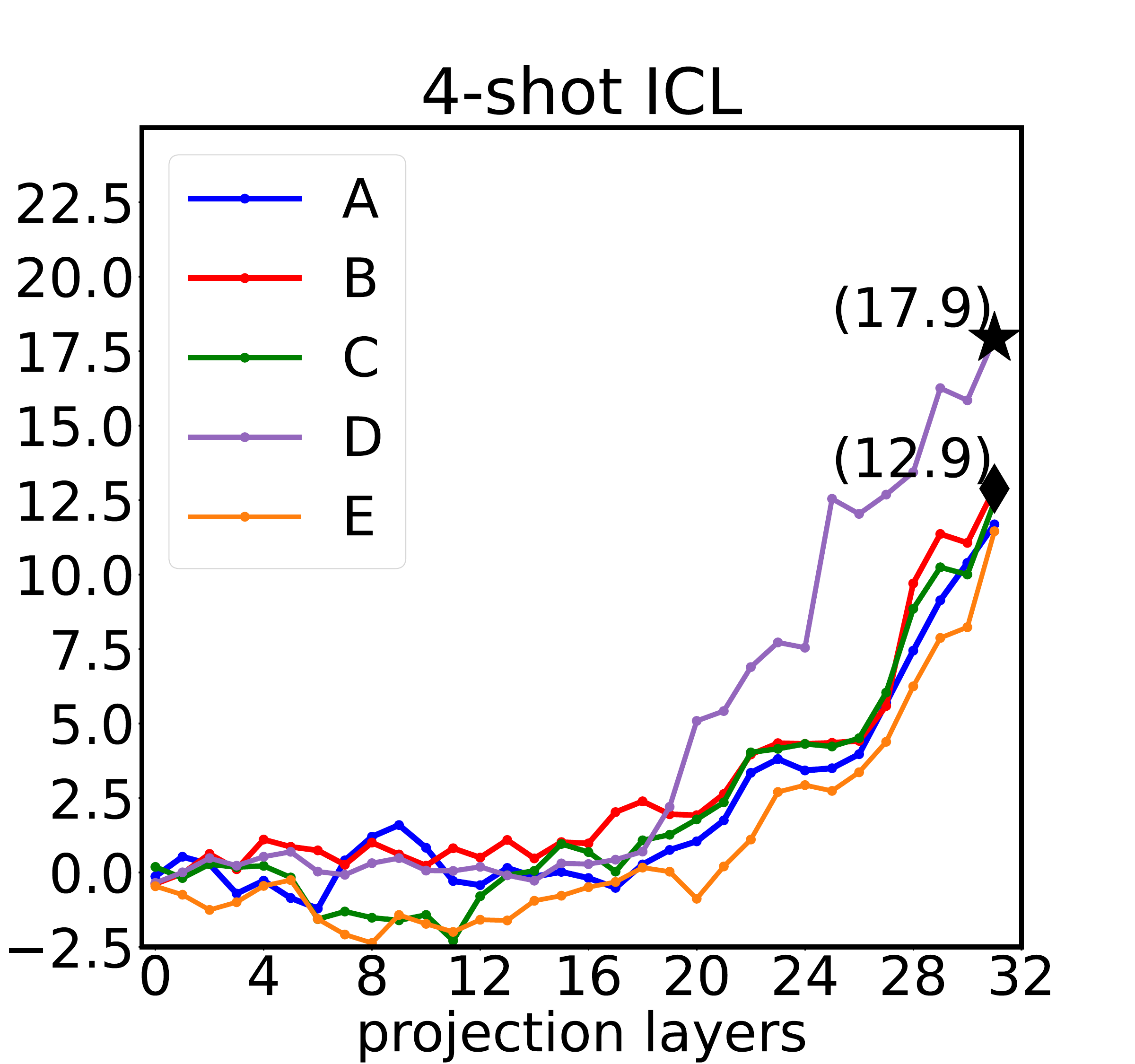}
    \end{subfigure}
    \hspace{-.3cm}
    \begin{subfigure}[b]{0.24\textwidth}
        \centering
        \includegraphics[width=\textwidth]{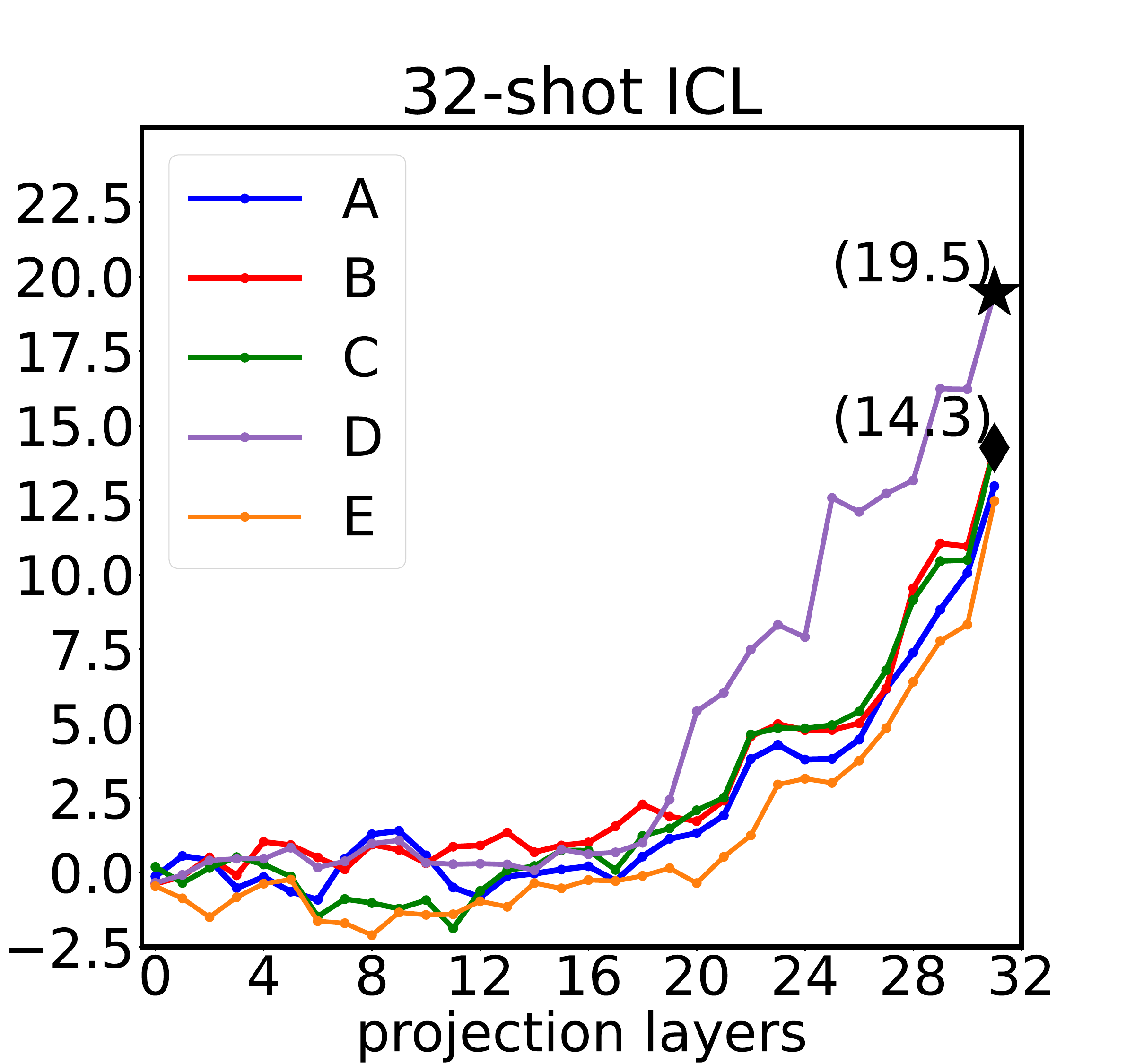}
    \end{subfigure}
    \hspace{-.2cm}
    \begin{subfigure}[b]{0.24\textwidth}
        \centering
        \includegraphics[width=\textwidth]{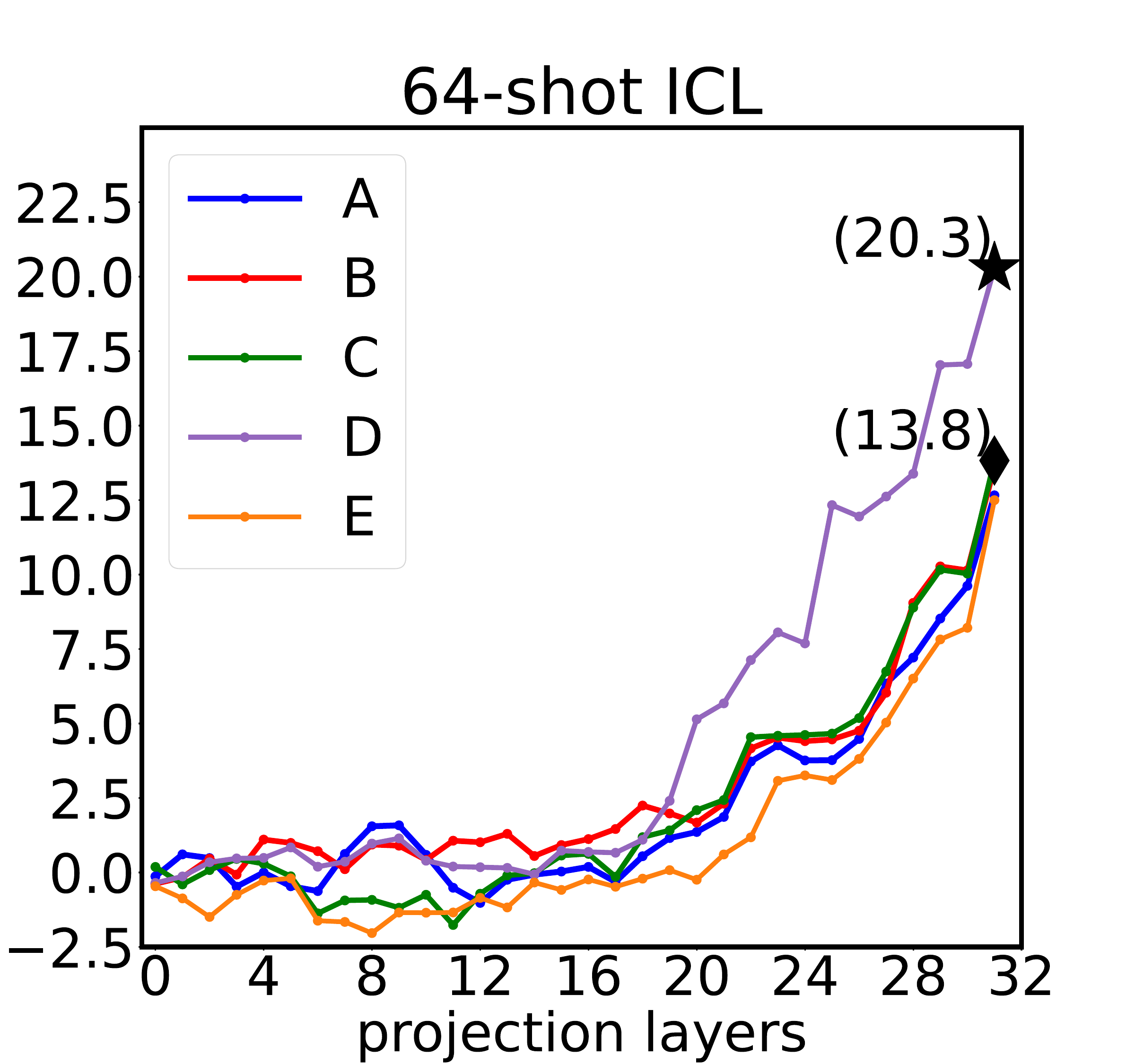}
    \end{subfigure}
    \hspace{-.2cm}
    \begin{subfigure}[b]{0.24\textwidth}
        \centering
        \includegraphics[width=\textwidth]{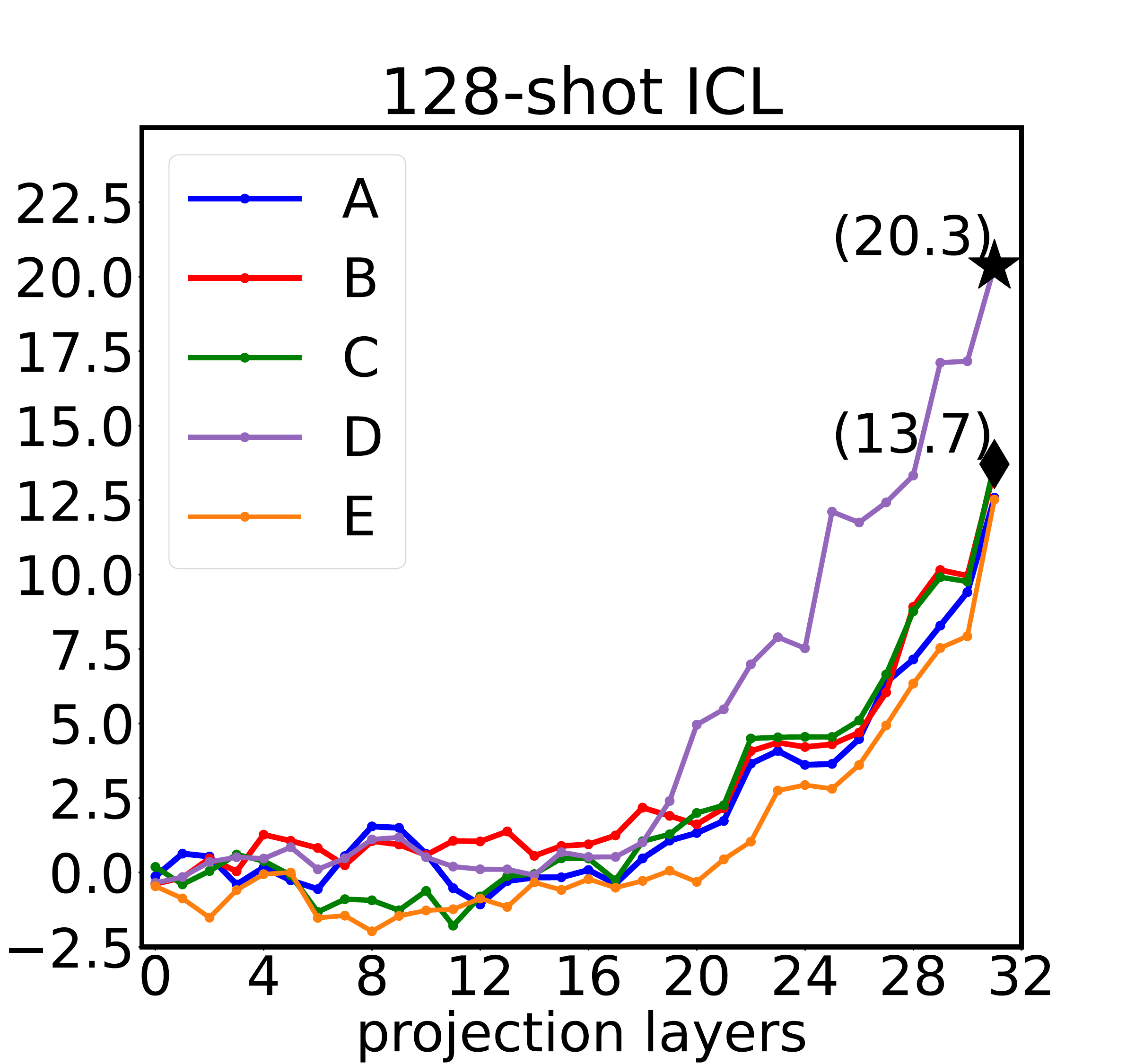}
    \end{subfigure}
    \caption{Average logits of Mistral-7B-v0.2 on the Commonsense QA dataset for MCQA items with the correct answer "D". Increasing in-context examples amplifies the logit of the correct option, thereby magnifying the difference between the logits of correct and incorrect options. $\star$ represent the highest logit and $\blacklozenge$ the second highest logit. Refer to Appendix \ref{Additional results: Average logits and probabilities} for additional results.}
    \label{fig: case study 2 logits D}
\end{figure*}
\paragraph{Projection into Vocabulary} To uncover the latent information encoded within residual streams, projecting intermediate states onto a probability distribution over the vocabulary space $V$ provides critical insights into the temporal and spatial dynamics of how these networks construct and refine their outputs \cite{geva-etal-2021-transformer,belrose2023eliciting,dar-etal-2023-analyzing}. Analogous to token generation, each residual stream $\mathbf{r}_{i}\hspace{-1pt} \in \hspace{-1pt} \mathbb{R}^d$ at the final position—where $i$ indexes the $i$-th residual in the model—undergoes transformation via an unembedding matrix $W_U \in \mathbb{R}^{d \times |V|}$ post layer normalization. This process yields calibrated logits $\mathbf{l}_i= {W}_{U} \mathbf{LayerNormalization}(\mathbf{r}_i)$ and the corresponding probabilities $\mathbf{p}_i=\mathbf{Softmax}(\mathbf{l}_i)$.

\paragraph{Correlation with Uncertainty in ICL} For $k$-shot in-context learning, consider projecting the residual representations at the answer position into the probability simplex $\Delta^{|V|}$ over the vocabulary $V$. Denote the resulting logits and probabilities of candidate symbols (e.g., "A", "B", "C") as $\ell_i$ and $p_i$, respectively. These logits ${\ell_i}$ or probabilities ${p_i}$ serve as proxies for confidence levels associated with each candidate. Analyzing the evolution of $\ell_i$ across model layers reveals the hierarchical development of inner confidence throughout $k$-shot learning, offering a profound understanding of the underlying uncertainty dynamics.

\subsection{RQ3: What mechanisms underlie uncertainty reduction?}\label{rq3}
\paragraph{Qualitative Analysis} To begin with, we present a case study in Fig.\ref{fig:case_2}, offering an intuitive and qualitative demonstration of how the number of shots influences uncertainty. In this case, the Mistral-7B model struggles to distinguish the correct answer, option "E", under a 4-shot ICL setting, as the other options continuously mislead the model throughout the process. This is evidenced by the fluctuating confidence levels, which rise and fall erratically. In contrast, as the number of shots increases (32-, 64-, and 128-shot settings), many-shot ICL consistently boosts the probability of selecting "E" as the correct answer from about 22nd layer onward, maintaining this highest probability thereafter. Simultaneously, it demonstrates robustness by maintaining near-zero probabilities for incorrect options, effectively eliminating the influence of distractors on the model’s final prediction.
\begin{table}[!h]
\centering
\resizebox{\linewidth}{!}{%
\begin{tabular}{c|c|c|c|c}
\toprule \hline
\multirow{1}{*}{\textbf{CMQA}}  & \multicolumn{1}{c|}{4-shot} & \multicolumn{1}{c|}{32-shot} & \multicolumn{1}{c|}{64-shot} & \multicolumn{1}{c}{128-shot}\\
\hline
\multirow{1}{*}{Llama-3.1} & 2.86 / 24.98 & 2.75 / 27.03 & 2.55 / 27.66 & 2.53 / 28.01\\
\hline
Mistral-v0.2  & 2.78 / 17.14 & 2.24 / 19.60 & 2.57 / 20.38 & 2.75 / 20.84 \\
\hline

Qwen1.5 & 3.51 / 29.11 &  3.62 / 30.49 & 3.73 / 30.97 & 3.76 / 30.94 \\
\hline
\hline
\textbf{LD3} & \multicolumn{1}{c|}{4-shot} & \multicolumn{1}{c|}{40-shot} & \multicolumn{1}{c|}{120-shot} & \multicolumn{1}{c}{240-shot}\\
 \hline
Llama-3.1 & 0.51 / 15.93 &  0.77 / 17.15 & 0.65 / 16.6 & 0.77 / 16.87 \\
 \hline
Mistral-7B-v0.2 & 0.26 / 11.07  &   0.48 / 11.92 & 0.46 / 12.05 & 0.59 / 11.87 \\
 \hline
Qwen1.5-7B  & 0.45 / 15.98 & 0.46 / 16.49 & 0.43 / 16.54 & 0.49 / 16.72 \\
 \hline
\bottomrule 
\end{tabular}
}
\caption{Average logit difference / the largest logit.}
\label{tab: logit diff}
\end{table}
\vspace{-.3cm}
\paragraph{Extended Examples Amplify Logit Disparity.}  
We compute the average logits $\ell_i$ and probabilities $p_i$ (Figs. \ref{fig: case study 2 logits D} and \ref{fig: case study 2 probs D}) across varying shot counts for groups sharing the same answer. The analysis reveals that extended ICL enhances the precision of LLMs, concentrating greater logit mass on the correct symbol while effectively suppressing alternatives. This dynamic, driven by the interplay between an amplified logit disparity and increased absolute logit values (Tab.\ref{tab: logit diff}), leverages the exponential sensitivity of the $\mathbf{Softmax}$ function to propel the probability of the correct symbol toward 1. Consequently, \textbf{the uncertainty in LLM predictions is significantly reduced.}

\section{Further Discussion}
\paragraph{Clarifications}
While our work builds upon the framework in \cite{ling-etal-2024-uncertainty}, our research specifically investigates the evolution of uncertainty in long-context ICL, a topic that has not been  examined to date. In contrast, \citeauthor{ling-etal-2024-uncertainty} primarily focus on introducing a framework for decomposing uncertainty in few-shot ICL. By shifting the focus to long-context scenarios, our study explores how uncertainty evolves as the number of in-context examples increases, thereby addressing an important yet understudied dimension of ICL.
\section{Conclusion}
\vspace{-.2cm}
This study investigates the impact of extra demonstrations on the confidence of LLMs in their responses. Experimental results demonstrate that additional examples significantly reduce TU across both simple and complex tasks by integrating task-specific knowledge. This reduction is primarily attributed to decreased model uncertainty, which enhances overall performance. However, in complex tasks, many-shot ICL faces challenges in reducing TU due to a concurrent increase in AU. Analysis of the internal representations of LLMs reveals that many-shot ICL not only reallocates greater logit mass toward correct responses but also enlarges the logit margin between correct answers and distractors, reflecting an increase in model confidence.
\section*{Limitation}
Our study is the first systematic investigation into uncertainty evolution in long-context ICL, addressing a critical research gap. These foundational experiments hope to provide a basis for future UQ studies on open-ended tasks. However, several limitations must be acknowledged.

\paragraph{Exclusion of Open-Ended Tasks}
The scope of this work does not encompass the uncertainty analysis of open-ended tasks, such as abstractive summarization \cite{hasan-etal-2021-xl} and machine translation \cite{costa2022no}, owing to the lack of robust UQ techniques for free-form generative scenarios. Nevertheless, applying ICL to rationale-intensive reasoning and generative contexts remains a promising direction. Future investigations should assess the reliability and trustworthiness of ICL in these domains, as advancements in this area could not only enhance task-solving performance but also broaden the applicability of UQ methodologies to more diverse and complex settings.

\paragraph{Limited Exploration of ICL Configurations}
This study also excludes several influential ICL paradigms, such as unsupervised ICL \cite{yu-etal-2024-rethinking}, reinforced ICL \cite{jiang2024manyshot}, and CoT prompting \cite{wei2022chain}, the latter of which is widely adopted in reasoning tasks to elicit step-by-step rationales. Existing UQ methods fall short of capturing the logical complexity intrinsic to reasoning-intensive contexts. Furthermore, practical challenges, including the computational overhead and context-length constraints of current open-source LLMs, prevented us from investigating extreme-shot ICL scenarios involving thousands of demonstrations. These limitations underscore promising directions for future research, particularly in applying UQ methodologies to better accommodate the unique challenges posed by reasoning tasks. More discussion in Appendix~\ref{appendix:related work}.

\section*{Broader Impact}
Despite these limitations, this study marks a pivotal advancement in understanding the reliability of ICL by harnessing recent breakthroughs in uncertainty quantification and decomposition, an essential yet underexplored aspect of LLM research. The research on uncertainty in ICL enriches the field of uncertainty quantification, providing novel perspectives on the trustworthiness of many-shot ICL. These contributions lay a solid foundation for broadening ICL's applicability in high-stakes domains. Ultimately, these findings could catalyze the development of more dependable and interpretable AI systems, offering profound societal impact.
\section*{Acknowledgements}
This work was supported in part by the National Natural Science Foundation of China under Grant 72293575, and in part by the Strategic Priority Research Program of Chinese Academy of Sciences under Grant XDA0480301 and the Excellent Youth Program of State Key Laboratory of Multimodal Artificial Intelligence Systems MAIS2024310.
\newpage
\bibliography{cal_references}

\begin{thebibliography}{54}
\expandafter\ifx\csname natexlab\endcsname\relax\def\natexlab#1{#1}\fi

\bibitem[{Agarwal et~al.(2024)Agarwal, Singh, Zhang, Bohnet, Rosias, Chan, Zhang, Faust, and Larochelle}]{agarwal2024manyshot}
Rishabh Agarwal, Avi Singh, Lei~M Zhang, Bernd Bohnet, Luis Rosias, Stephanie~C.Y. Chan, Biao Zhang, Aleksandra Faust, and Hugo Larochelle. 2024.
\newblock \href {https://openreview.net/forum?id=goi7DFHlqS} {Many-shot in-context learning}.
\newblock In \emph{ICML 2024 Workshop on In-Context Learning}.

\bibitem[{Aky{\"u}rek et~al.()Aky{\"u}rek, Schuurmans, Andreas, Ma, and Zhou}]{akyureklearning}
Ekin Aky{\"u}rek, Dale Schuurmans, Jacob Andreas, Tengyu Ma, and Denny Zhou.
\newblock What learning algorithm is in-context learning? investigations with linear models.
\newblock In \emph{The Eleventh International Conference on Learning Representations}.

\bibitem[{Bai et~al.(2023)Bai, Bai, Chu, Cui, Dang, Deng, Fan, Ge, Han, Huang, Hui, Ji, Li, Lin, Lin, Liu, Liu, Lu, Lu, Ma, Men, Ren, Ren, Tan, Tan, Tu, Wang, Wang, Wang, Wu, Xu, Xu, Yang, Yang, Yang, Yang, Yao, Yu, Yuan, Yuan, Zhang, Zhang, Zhang, Zhang, Zhou, Zhou, Zhou, and Zhu}]{bai2023qwentechnicalreport}
Jinze Bai, Shuai Bai, Yunfei Chu, Zeyu Cui, Kai Dang, Xiaodong Deng, Yang Fan, Wenbin Ge, Yu~Han, Fei Huang, Binyuan Hui, Luo Ji, Mei Li, Junyang Lin, Runji Lin, Dayiheng Liu, Gao Liu, Chengqiang Lu, Keming Lu, Jianxin Ma, Rui Men, Xingzhang Ren, Xuancheng Ren, Chuanqi Tan, Sinan Tan, Jianhong Tu, Peng Wang, Shijie Wang, Wei Wang, Shengguang Wu, Benfeng Xu, Jin Xu, An~Yang, Hao Yang, Jian Yang, Shusheng Yang, Yang Yao, Bowen Yu, Hongyi Yuan, Zheng Yuan, Jianwei Zhang, Xingxuan Zhang, Yichang Zhang, Zhenru Zhang, Chang Zhou, Jingren Zhou, Xiaohuan Zhou, and Tianhang Zhu. 2023.
\newblock \href {http://arxiv.org/abs/2309.16609} {Qwen technical report}.

\bibitem[{Bakman et~al.(2024)Bakman, Yaldiz, Buyukates, Tao, Dimitriadis, and Avestimehr}]{bakman-etal-2024-mars}
Yavuz~Faruk Bakman, Duygu~Nur Yaldiz, Baturalp Buyukates, Chenyang Tao, Dimitrios Dimitriadis, and Salman Avestimehr. 2024.
\newblock \href {https://doi.org/10.18653/v1/2024.acl-long.419} {{MARS}: Meaning-aware response scoring for uncertainty estimation in generative {LLM}s}.
\newblock In \emph{Proceedings of the 62nd Annual Meeting of the Association for Computational Linguistics (Volume 1: Long Papers)}, pages 7752--7767, Bangkok, Thailand. Association for Computational Linguistics.

\bibitem[{Belrose et~al.(2023)Belrose, Furman, Smith, Halawi, Ostrovsky, McKinney, Biderman, and Steinhardt}]{belrose2023eliciting}
Nora Belrose, Zach Furman, Logan Smith, Danny Halawi, Igor Ostrovsky, Lev McKinney, Stella Biderman, and Jacob Steinhardt. 2023.
\newblock Eliciting latent predictions from transformers with the tuned lens.
\newblock \emph{arXiv preprint arXiv:2303.08112}.

\bibitem[{Bertsch et~al.(2024)Bertsch, Ivgi, Alon, Berant, Gormley, and Neubig}]{bertsch2024incontext}
Amanda Bertsch, Maor Ivgi, Uri Alon, Jonathan Berant, Matthew~R. Gormley, and Graham Neubig. 2024.
\newblock \href {https://openreview.net/forum?id=4KAmc7vUbq} {In-context learning with long-context models: An in-depth exploration}.
\newblock In \emph{First Workshop on Long-Context Foundation Models @ ICML 2024}.

\bibitem[{Besta et~al.(2024)Besta, Blach, Kubicek, Gerstenberger, Podstawski, Gianinazzi, Gajda, Lehmann, Niewiadomski, Nyczyk et~al.}]{besta2024graph}
Maciej Besta, Nils Blach, Ales Kubicek, Robert Gerstenberger, Michal Podstawski, Lukas Gianinazzi, Joanna Gajda, Tomasz Lehmann, Hubert Niewiadomski, Piotr Nyczyk, et~al. 2024.
\newblock Graph of thoughts: Solving elaborate problems with large language models.
\newblock In \emph{Proceedings of the AAAI Conference on Artificial Intelligence}, volume~38, pages 17682--17690.

\bibitem[{Brown et~al.(2020)Brown, Mann, Ryder, Subbiah, Kaplan, Dhariwal, Neelakantan, Shyam, Sastry, Askell, Agarwal, Herbert-Voss, Krueger, Henighan, Child, Ramesh, Ziegler, Wu, Winter, Hesse, Chen, Sigler, Litwin, Gray, Chess, Clark, Berner, McCandlish, Radford, Sutskever, and Amodei}]{NEURIPS2020_1457c0d6}
Tom Brown, Benjamin Mann, Nick Ryder, Melanie Subbiah, Jared~D Kaplan, Prafulla Dhariwal, Arvind Neelakantan, Pranav Shyam, Girish Sastry, Amanda Askell, Sandhini Agarwal, Ariel Herbert-Voss, Gretchen Krueger, Tom Henighan, Rewon Child, Aditya Ramesh, Daniel Ziegler, Jeffrey Wu, Clemens Winter, Chris Hesse, Mark Chen, Eric Sigler, Mateusz Litwin, Scott Gray, Benjamin Chess, Jack Clark, Christopher Berner, Sam McCandlish, Alec Radford, Ilya Sutskever, and Dario Amodei. 2020.
\newblock \href {https://proceedings.neurips.cc/paper_files/paper/2020/file/1457c0d6bfcb4967418bfb8ac142f64a-Paper.pdf} {Language models are few-shot learners}.
\newblock In \emph{Advances in Neural Information Processing Systems}, volume~33, pages 1877--1901. Curran Associates, Inc.

\bibitem[{Chan et~al.(2022)Chan, Santoro, Lampinen, Wang, Singh, Richemond, McClelland, and Hill}]{chan2022data}
Stephanie Chan, Adam Santoro, Andrew Lampinen, Jane Wang, Aaditya Singh, Pierre Richemond, James McClelland, and Felix Hill. 2022.
\newblock Data distributional properties drive emergent in-context learning in transformers.
\newblock \emph{Advances in neural information processing systems}, 35:18878--18891.

\bibitem[{Costa-juss{\`a} et~al.(2022)Costa-juss{\`a}, Cross, {\c{C}}elebi, Elbayad, Heafield, Heffernan, Kalbassi, Lam, Licht, Maillard et~al.}]{costa2022no}
Marta~R Costa-juss{\`a}, James Cross, Onur {\c{C}}elebi, Maha Elbayad, Kenneth Heafield, Kevin Heffernan, Elahe Kalbassi, Janice Lam, Daniel Licht, Jean Maillard, et~al. 2022.
\newblock No language left behind: Scaling human-centered machine translation.
\newblock \emph{arXiv e-prints}, pages arXiv--2207.

\bibitem[{Dar et~al.(2023)Dar, Geva, Gupta, and Berant}]{dar-etal-2023-analyzing}
Guy Dar, Mor Geva, Ankit Gupta, and Jonathan Berant. 2023.
\newblock \href {https://doi.org/10.18653/v1/2023.acl-long.893} {Analyzing transformers in embedding space}.
\newblock In \emph{Proceedings of the 61st Annual Meeting of the Association for Computational Linguistics (Volume 1: Long Papers)}, pages 16124--16170, Toronto, Canada. Association for Computational Linguistics.

\bibitem[{Dong et~al.(2024)Dong, Li, Dai, Zheng, Ma, Li, Xia, Xu, Wu, Chang, Sun, Li, and Sui}]{dong-etal-2024-survey}
Qingxiu Dong, Lei Li, Damai Dai, Ce~Zheng, Jingyuan Ma, Rui Li, Heming Xia, Jingjing Xu, Zhiyong Wu, Baobao Chang, Xu~Sun, Lei Li, and Zhifang Sui. 2024.
\newblock \href {https://doi.org/10.18653/v1/2024.emnlp-main.64} {A survey on in-context learning}.
\newblock In \emph{Proceedings of the 2024 Conference on Empirical Methods in Natural Language Processing}, pages 1107--1128, Miami, Florida, USA. Association for Computational Linguistics.

\bibitem[{Gatt and Krahmer(2018)}]{10.5555/3241691.3241693}
Albert Gatt and Emiel Krahmer. 2018.
\newblock Survey of the state of the art in natural language generation: core tasks, applications and evaluation.
\newblock \emph{J. Artif. Int. Res.}, 61(1):65–170.

\bibitem[{Gawlikowski et~al.(2022)Gawlikowski, Tassi, Ali, Lee, Humt, Feng, Kruspe, Triebel, Jung, Roscher, Shahzad, Yang, Bamler, and Zhu}]{gawlikowski2022surveyuncertaintydeepneural}
Jakob Gawlikowski, Cedrique Rovile~Njieutcheu Tassi, Mohsin Ali, Jongseok Lee, Matthias Humt, Jianxiang Feng, Anna Kruspe, Rudolph Triebel, Peter Jung, Ribana Roscher, Muhammad Shahzad, Wen Yang, Richard Bamler, and Xiao~Xiang Zhu. 2022.
\newblock \href {http://arxiv.org/abs/2107.03342} {A survey of uncertainty in deep neural networks}.

\bibitem[{Geva et~al.(2021)Geva, Schuster, Berant, and Levy}]{geva-etal-2021-transformer}
Mor Geva, Roei Schuster, Jonathan Berant, and Omer Levy. 2021.
\newblock \href {https://doi.org/10.18653/v1/2021.emnlp-main.446} {Transformer feed-forward layers are key-value memories}.
\newblock In \emph{Proceedings of the 2021 Conference on Empirical Methods in Natural Language Processing}, pages 5484--5495, Online and Punta Cana, Dominican Republic. Association for Computational Linguistics.

\bibitem[{Gu and Dao(2024)}]{gu2024mamba}
Albert Gu and Tri Dao. 2024.
\newblock \href {https://openreview.net/forum?id=AL1fq05o7H} {Mamba: Linear-time sequence modeling with selective state spaces}.

\bibitem[{Hasan et~al.(2021)Hasan, Bhattacharjee, Islam, Mubasshir, Li, Kang, Rahman, and Shahriyar}]{hasan-etal-2021-xl}
Tahmid Hasan, Abhik Bhattacharjee, Md.~Saiful Islam, Kazi Mubasshir, Yuan-Fang Li, Yong-Bin Kang, M.~Sohel Rahman, and Rifat Shahriyar. 2021.
\newblock \href {https://doi.org/10.18653/v1/2021.findings-acl.413} {{XL}-sum: Large-scale multilingual abstractive summarization for 44 languages}.
\newblock In \emph{Findings of the Association for Computational Linguistics: ACL-IJCNLP 2021}, pages 4693--4703, Online. Association for Computational Linguistics.

\bibitem[{He et~al.()He, Zhang, Ren, and Sun}]{hedeep}
Kaiming He, Xiangyu Zhang, Shaoqing Ren, and Jian Sun.
\newblock Deep residual learning for image recognition.

\bibitem[{He et~al.(2023)He, Jiang, Xiao, Xu, and Li}]{he2023survey}
Wenchong He, Zhe Jiang, Tingsong Xiao, Zelin Xu, and Yukun Li. 2023.
\newblock A survey on uncertainty quantification methods for deep learning.
\newblock \emph{arXiv preprint arXiv:2302.13425}.

\bibitem[{Hou et~al.(2024)Hou, Liu, Qian, Andreas, Chang, and Zhang}]{hou2024decomposing}
Bairu Hou, Yujian Liu, Kaizhi Qian, Jacob Andreas, Shiyu Chang, and Yang Zhang. 2024.
\newblock \href {https://openreview.net/forum?id=byxXa99PtF} {Decomposing uncertainty for large language models through input clarification ensembling}.
\newblock In \emph{Forty-first International Conference on Machine Learning}.

\bibitem[{Jiang et~al.(2023)Jiang, Sablayrolles, Mensch, Bamford, Chaplot, de~las Casas, Bressand, Lengyel, Lample, Saulnier, Lavaud, Lachaux, Stock, Scao, Lavril, Wang, Lacroix, and Sayed}]{jiang2023mistral7b}
Albert~Q. Jiang, Alexandre Sablayrolles, Arthur Mensch, Chris Bamford, Devendra~Singh Chaplot, Diego de~las Casas, Florian Bressand, Gianna Lengyel, Guillaume Lample, Lucile Saulnier, Lélio~Renard Lavaud, Marie-Anne Lachaux, Pierre Stock, Teven~Le Scao, Thibaut Lavril, Thomas Wang, Timothée Lacroix, and William~El Sayed. 2023.
\newblock \href {http://arxiv.org/abs/2310.06825} {Mistral 7b}.

\bibitem[{Jiang et~al.(2024)Jiang, Irvin, Wang, Chaudhry, Chen, and Ng}]{jiang2024manyshot}
Yixing Jiang, Jeremy~Andrew Irvin, Ji~Hun Wang, Muhammad~Ahmed Chaudhry, Jonathan~H Chen, and Andrew~Y. Ng. 2024.
\newblock \href {https://openreview.net/forum?id=j2rKwWXdcz} {Many-shot in-context learning in multimodal foundation models}.
\newblock In \emph{ICML 2024 Workshop on In-Context Learning}.

\bibitem[{Jin et~al.(2024)Jin, Han, Yang, Jiang, Liu, Chang, Chen, and Hu}]{jin2024llm}
Hongye Jin, Xiaotian Han, Jingfeng Yang, Zhimeng Jiang, Zirui Liu, Chia-Yuan Chang, Huiyuan Chen, and Xia Hu. 2024.
\newblock \href {http://arxiv.org/abs/2401.01325} {Llm maybe longlm: Self-extend llm context window without tuning}.

\bibitem[{Kong et~al.(2023)Kong, Kamarthi, Chen, Prakash, and Zhang}]{10.1145/3580305.3599577}
Lingkai Kong, Harshavardhan Kamarthi, Peng Chen, B.~Aditya Prakash, and Chao Zhang. 2023.
\newblock \href {https://doi.org/10.1145/3580305.3599577} {Uncertainty quantification in deep learning}.
\newblock In \emph{Proceedings of the 29th ACM SIGKDD Conference on Knowledge Discovery and Data Mining}, KDD '23, page 5809–5810, New York, NY, USA. Association for Computing Machinery.

\bibitem[{Kuhn et~al.(2023)Kuhn, Gal, and Farquhar}]{kuhn2023semantic}
Lorenz Kuhn, Yarin Gal, and Sebastian Farquhar. 2023.
\newblock \href {https://openreview.net/forum?id=VD-AYtP0dve} {Semantic uncertainty: Linguistic invariances for uncertainty estimation in natural language generation}.
\newblock In \emph{The Eleventh International Conference on Learning Representations}.

\bibitem[{Lakshminarayanan et~al.(2017)Lakshminarayanan, Pritzel, and Blundell}]{lakshminarayanan2017simple}
Balaji Lakshminarayanan, Alexander Pritzel, and Charles Blundell. 2017.
\newblock Simple and scalable predictive uncertainty estimation using deep ensembles.
\newblock \emph{Advances in neural information processing systems}, 30.

\bibitem[{Li and Papyan(2023)}]{li2023residual}
Jianing Li and Vardan Papyan. 2023.
\newblock \href {https://openreview.net/forum?id=zOCIKYVaF5} {Residual alignment: Uncovering the mechanisms of residual networks}.
\newblock In \emph{Thirty-seventh Conference on Neural Information Processing Systems}.

\bibitem[{Li et~al.(2024)Li, Zhang, Do, Yue, and Chen}]{li2024longcontextllmsstrugglelong}
Tianle Li, Ge~Zhang, Quy~Duc Do, Xiang Yue, and Wenhu Chen. 2024.
\newblock \href {http://arxiv.org/abs/2404.02060} {Long-context llms struggle with long in-context learning}.

\bibitem[{Lin et~al.(2024)Lin, Trivedi, and Sun}]{lin2024generating}
Zhen Lin, Shubhendu Trivedi, and Jimeng Sun. 2024.
\newblock \href {https://openreview.net/forum?id=DWkJCSxKU5} {Generating with confidence: Uncertainty quantification for black-box large language models}.
\newblock \emph{Transactions on Machine Learning Research}.

\bibitem[{Ling et~al.(2024)Ling, Zhao, Zhang, Cheng, Liu, Sun, Oishi, Osaki, Matsuda, Ji, Bai, Zhao, and Chen}]{ling-etal-2024-uncertainty}
Chen Ling, Xujiang Zhao, Xuchao Zhang, Wei Cheng, Yanchi Liu, Yiyou Sun, Mika Oishi, Takao Osaki, Katsushi Matsuda, Jie Ji, Guangji Bai, Liang Zhao, and Haifeng Chen. 2024.
\newblock \href {https://doi.org/10.18653/v1/2024.naacl-long.184} {Uncertainty quantification for in-context learning of large language models}.
\newblock In \emph{Proceedings of the 2024 Conference of the North American Chapter of the Association for Computational Linguistics: Human Language Technologies (Volume 1: Long Papers)}, pages 3357--3370, Mexico City, Mexico. Association for Computational Linguistics.

\bibitem[{Liu et~al.(2024)Liu, Khalifa, and Wang}]{liu2024litcab}
Xin Liu, Muhammad Khalifa, and Lu~Wang. 2024.
\newblock \href {https://openreview.net/forum?id=jH67LHVOIO} {Litcab: Lightweight language model calibration over short- and long-form responses}.
\newblock In \emph{The Twelfth International Conference on Learning Representations}.

\bibitem[{Min et~al.(2022)Min, Lyu, Holtzman, Artetxe, Lewis, Hajishirzi, and Zettlemoyer}]{min-etal-2022-rethinking}
Sewon Min, Xinxi Lyu, Ari Holtzman, Mikel Artetxe, Mike Lewis, Hannaneh Hajishirzi, and Luke Zettlemoyer. 2022.
\newblock \href {https://doi.org/10.18653/v1/2022.emnlp-main.759} {Rethinking the role of demonstrations: What makes in-context learning work?}
\newblock In \emph{Proceedings of the 2022 Conference on Empirical Methods in Natural Language Processing}, pages 11048--11064, Abu Dhabi, United Arab Emirates. Association for Computational Linguistics.

\bibitem[{Peng et~al.(2023)Peng, Alcaide, Anthony, Albalak, Arcadinho, Biderman, Cao, Cheng, Chung, Derczynski, Du, Grella, Gv, He, Hou, Kazienko, Kocon, Kong, Koptyra, Lau, Lin, Mantri, Mom, Saito, Song, Tang, Wind, Wo{\'z}niak, Zhang, Zhou, Zhu, and Zhu}]{peng-etal-2023-rwkv}
Bo~Peng, Eric Alcaide, Quentin Anthony, Alon Albalak, Samuel Arcadinho, Stella Biderman, Huanqi Cao, Xin Cheng, Michael Chung, Leon Derczynski, Xingjian Du, Matteo Grella, Kranthi Gv, Xuzheng He, Haowen Hou, Przemyslaw Kazienko, Jan Kocon, Jiaming Kong, Bart{\l}omiej Koptyra, Hayden Lau, Jiaju Lin, Krishna Sri~Ipsit Mantri, Ferdinand Mom, Atsushi Saito, Guangyu Song, Xiangru Tang, Johan Wind, Stanis{\l}aw Wo{\'z}niak, Zhenyuan Zhang, Qinghua Zhou, Jian Zhu, and Rui-Jie Zhu. 2023.
\newblock \href {https://doi.org/10.18653/v1/2023.findings-emnlp.936} {{RWKV}: Reinventing {RNN}s for the transformer era}.
\newblock In \emph{Findings of the Association for Computational Linguistics: EMNLP 2023}, pages 14048--14077, Singapore. Association for Computational Linguistics.

\bibitem[{Robinson and Wingate(2023)}]{robinson2023leveraging}
Joshua Robinson and David Wingate. 2023.
\newblock \href {https://openreview.net/forum?id=yKbprarjc5B} {Leveraging large language models for multiple choice question answering}.
\newblock In \emph{The Eleventh International Conference on Learning Representations}.

\bibitem[{Rozière et~al.(2024)Rozière, Gehring, Gloeckle, Sootla, Gat, Tan, Adi, Liu, Sauvestre, Remez, Rapin, Kozhevnikov, Evtimov, Bitton, Bhatt, Ferrer, Grattafiori, Xiong, Défossez, Copet, Azhar, Touvron, Martin, Usunier, Scialom, and Synnaeve}]{rozière2024codellamaopenfoundation}
Baptiste Rozière, Jonas Gehring, Fabian Gloeckle, Sten Sootla, Itai Gat, Xiaoqing~Ellen Tan, Yossi Adi, Jingyu Liu, Romain Sauvestre, Tal Remez, Jérémy Rapin, Artyom Kozhevnikov, Ivan Evtimov, Joanna Bitton, Manish Bhatt, Cristian~Canton Ferrer, Aaron Grattafiori, Wenhan Xiong, Alexandre Défossez, Jade Copet, Faisal Azhar, Hugo Touvron, Louis Martin, Nicolas Usunier, Thomas Scialom, and Gabriel Synnaeve. 2024.
\newblock \href {http://arxiv.org/abs/2308.12950} {Code llama: Open foundation models for code}.

\bibitem[{Shazeer et~al.(2017)Shazeer, Mirhoseini, Maziarz, Davis, Le, Hinton, and Dean}]{shazeer2017outrageously}
Noam Shazeer, Azalia Mirhoseini, Krzysztof Maziarz, Andy Davis, Quoc Le, Geoffrey Hinton, and Jeff Dean. 2017.
\newblock Outrageously large neural networks: The sparsely-gated mixture-of-experts layer.
\newblock \emph{arXiv preprint arXiv:1701.06538}.

\bibitem[{Socher et~al.(2013)Socher, Perelygin, Wu, Chuang, Manning, Ng, and Potts}]{socher-etal-2013-recursive}
Richard Socher, Alex Perelygin, Jean Wu, Jason Chuang, Christopher~D. Manning, Andrew Ng, and Christopher Potts. 2013.
\newblock \href {https://aclanthology.org/D13-1170} {Recursive deep models for semantic compositionality over a sentiment treebank}.
\newblock In \emph{Proceedings of the 2013 Conference on Empirical Methods in Natural Language Processing}, pages 1631--1642, Seattle, Washington, USA. Association for Computational Linguistics.

\bibitem[{Su et~al.(2024)Su, Ahmed, Lu, Pan, Bo, and Liu}]{10.1016/j.neucom.2023.127063}
Jianlin Su, Murtadha Ahmed, Yu~Lu, Shengfeng Pan, Wen Bo, and Yunfeng Liu. 2024.
\newblock \href {https://doi.org/10.1016/j.neucom.2023.127063} {Roformer: Enhanced transformer with rotary position embedding}.
\newblock \emph{Neurocomput.}, 568(C).

\bibitem[{Suzgun et~al.(2023)Suzgun, Scales, Sch{\"a}rli, Gehrmann, Tay, Chung, Chowdhery, Le, Chi, Zhou, and Wei}]{suzgun-etal-2023-challenging}
Mirac Suzgun, Nathan Scales, Nathanael Sch{\"a}rli, Sebastian Gehrmann, Yi~Tay, Hyung~Won Chung, Aakanksha Chowdhery, Quoc Le, Ed~Chi, Denny Zhou, and Jason Wei. 2023.
\newblock \href {https://doi.org/10.18653/v1/2023.findings-acl.824} {Challenging {BIG}-bench tasks and whether chain-of-thought can solve them}.
\newblock In \emph{Findings of the Association for Computational Linguistics: ACL 2023}, pages 13003--13051, Toronto, Canada. Association for Computational Linguistics.

\bibitem[{Talmor et~al.(2019)Talmor, Herzig, Lourie, and Berant}]{talmor-etal-2019-commonsenseqa}
Alon Talmor, Jonathan Herzig, Nicholas Lourie, and Jonathan Berant. 2019.
\newblock \href {https://doi.org/10.18653/v1/N19-1421} {{C}ommonsense{QA}: A question answering challenge targeting commonsense knowledge}.
\newblock In \emph{Proceedings of the 2019 Conference of the North {A}merican Chapter of the Association for Computational Linguistics: Human Language Technologies, Volume 1 (Long and Short Papers)}, pages 4149--4158, Minneapolis, Minnesota. Association for Computational Linguistics.

\bibitem[{Team et~al.(2024)Team, Georgiev, Lei, Burnell, Bai, Gulati, Tanzer, Vincent, Pan, Wang et~al.}]{team2024gemini}
Gemini Team, Petko Georgiev, Ving~Ian Lei, Ryan Burnell, Libin Bai, Anmol Gulati, Garrett Tanzer, Damien Vincent, Zhufeng Pan, Shibo Wang, et~al. 2024.
\newblock Gemini 1.5: Unlocking multimodal understanding across millions of tokens of context.
\newblock \emph{arXiv preprint arXiv:2403.05530}.

\bibitem[{Touvron et~al.(2023)Touvron, Lavril, Izacard, Martinet, Lachaux, Lacroix, Rozière, Goyal, Hambro, Azhar, Rodriguez, Joulin, Grave, and Lample}]{touvron2023llamaopenefficientfoundation}
Hugo Touvron, Thibaut Lavril, Gautier Izacard, Xavier Martinet, Marie-Anne Lachaux, Timothée Lacroix, Baptiste Rozière, Naman Goyal, Eric Hambro, Faisal Azhar, Aurelien Rodriguez, Armand Joulin, Edouard Grave, and Guillaume Lample. 2023.
\newblock \href {http://arxiv.org/abs/2302.13971} {Llama: Open and efficient foundation language models}.

\bibitem[{Valdenegro-Toro and Mori(2022)}]{Valdenegro-Toro_2022_CVPR}
Matias Valdenegro-Toro and Daniel~Saromo Mori. 2022.
\newblock A deeper look into aleatoric and epistemic uncertainty disentanglement.
\newblock In \emph{Proceedings of the IEEE/CVF Conference on Computer Vision and Pattern Recognition (CVPR) Workshops}, pages 1509--1517.

\bibitem[{Vaswani(2017)}]{vaswani2017attention}
A~Vaswani. 2017.
\newblock Attention is all you need.
\newblock \emph{Advances in Neural Information Processing Systems}.

\bibitem[{Wang et~al.(2024)Wang, Chen, Wen, Sheng, Li, and Zeng}]{wang-etal-2024-unveiling}
Yifei Wang, Yuheng Chen, Wanting Wen, Yu~Sheng, Linjing Li, and Daniel~Dajun Zeng. 2024.
\newblock \href {https://doi.org/10.18653/v1/2024.emnlp-main.420} {Unveiling factual recall behaviors of large language models through knowledge neurons}.
\newblock In \emph{Proceedings of the 2024 Conference on Empirical Methods in Natural Language Processing}, pages 7388--7402, Miami, Florida, USA. Association for Computational Linguistics.

\bibitem[{Wei et~al.(2022{\natexlab{a}})Wei, Bosma, Zhao, Guu, Yu, Lester, Du, Dai, and Le}]{wei2022finetuned}
Jason Wei, Maarten Bosma, Vincent Zhao, Kelvin Guu, Adams~Wei Yu, Brian Lester, Nan Du, Andrew~M. Dai, and Quoc~V Le. 2022{\natexlab{a}}.
\newblock \href {https://openreview.net/forum?id=gEZrGCozdqR} {Finetuned language models are zero-shot learners}.
\newblock In \emph{International Conference on Learning Representations}.

\bibitem[{Wei et~al.(2022{\natexlab{b}})Wei, Wang, Schuurmans, Bosma, Xia, Chi, Le, Zhou et~al.}]{wei2022chain}
Jason Wei, Xuezhi Wang, Dale Schuurmans, Maarten Bosma, Fei Xia, Ed~Chi, Quoc~V Le, Denny Zhou, et~al. 2022{\natexlab{b}}.
\newblock Chain-of-thought prompting elicits reasoning in large language models.
\newblock \emph{Advances in neural information processing systems}, 35:24824--24837.

\bibitem[{Wiegreffe et~al.(2024)Wiegreffe, Tafjord, Belinkov, Hajishirzi, and Sabharwal}]{wiegreffe2024answerassembleaceunderstanding}
Sarah Wiegreffe, Oyvind Tafjord, Yonatan Belinkov, Hannaneh Hajishirzi, and Ashish Sabharwal. 2024.
\newblock \href {http://arxiv.org/abs/2407.15018} {Answer, assemble, ace: Understanding how transformers answer multiple choice questions}.

\bibitem[{Xiao et~al.(2024)Xiao, Tian, Chen, Han, and Lewis}]{xiao2024efficient}
Guangxuan Xiao, Yuandong Tian, Beidi Chen, Song Han, and Mike Lewis. 2024.
\newblock \href {https://openreview.net/forum?id=NG7sS51zVF} {Efficient streaming language models with attention sinks}.
\newblock In \emph{The Twelfth International Conference on Learning Representations}.

\bibitem[{Yao et~al.(2023)Yao, Yu, Zhao, Shafran, Griffiths, Cao, and Narasimhan}]{yao2023tree}
Shunyu Yao, Dian Yu, Jeffrey Zhao, Izhak Shafran, Tom Griffiths, Yuan Cao, and Karthik Narasimhan. 2023.
\newblock Tree of thoughts: Deliberate problem solving with large language models.
\newblock \emph{Advances in neural information processing systems}, 36:11809--11822.

\bibitem[{Yu et~al.(2024)Yu, Liu, Yu, Yu, and Ao}]{yu-etal-2024-rethinking}
Guoxin Yu, Lemao Liu, Mo~Yu, Yue Yu, and Xiang Ao. 2024.
\newblock \href {https://doi.org/10.18653/v1/2024.emnlp-main.779} {Rethinking the evaluation of in-context learning for {LLM}s}.
\newblock In \emph{Proceedings of the 2024 Conference on Empirical Methods in Natural Language Processing}, pages 14068--14082, Miami, Florida, USA. Association for Computational Linguistics.

\bibitem[{Zhang et~al.(2024{\natexlab{a}})Zhang, Liu, Basaldella, and Collier}]{zhang-etal-2024-luq}
Caiqi Zhang, Fangyu Liu, Marco Basaldella, and Nigel Collier. 2024{\natexlab{a}}.
\newblock \href {https://aclanthology.org/2024.emnlp-main.299} {{LUQ}: Long-text uncertainty quantification for {LLM}s}.
\newblock In \emph{Proceedings of the 2024 Conference on Empirical Methods in Natural Language Processing}, pages 5244--5262, Miami, Florida, USA. Association for Computational Linguistics.

\bibitem[{Zhang et~al.(2024{\natexlab{b}})Zhang, Deng, Liu, Pan, and Bing}]{zhang-etal-2024-sentiment}
Wenxuan Zhang, Yue Deng, Bing Liu, Sinno Pan, and Lidong Bing. 2024{\natexlab{b}}.
\newblock \href {https://doi.org/10.18653/v1/2024.findings-naacl.246} {Sentiment analysis in the era of large language models: A reality check}.
\newblock In \emph{Findings of the Association for Computational Linguistics: NAACL 2024}, pages 3881--3906, Mexico City, Mexico. Association for Computational Linguistics.

\bibitem[{Zhang et~al.(2015)Zhang, Zhao, and LeCun}]{NIPS2015_250cf8b5}
Xiang Zhang, Junbo Zhao, and Yann LeCun. 2015.
\newblock \href {https://proceedings.neurips.cc/paper_files/paper/2015/file/250cf8b51c773f3f8dc8b4be867a9a02-Paper.pdf} {Character-level convolutional networks for text classification}.
\newblock In \emph{Advances in Neural Information Processing Systems}, volume~28. Curran Associates, Inc.

\end{thebibliography}
\newpage
\onecolumn
\appendix
\section{Related Work}\label{appendix:related work}
\paragraph{Overview.} Estimating uncertainty in generation tasks presents greater challenges \cite{kuhn2023semantic} compared to tasks with a predefined candidate set like classification tasks \cite{zhang-etal-2024-sentiment} and multiple-choice question answering (MCQA) \cite{robinson2023leveraging}. This is primarily due to the vast, high-dimensional semantic space inherent in natural language, which results in an effectively infinite generation space \cite{lin2024generating,ling-etal-2024-uncertainty,liu2024litcab}. In contrast, classification tasks provide LLMs with a finite set of discrete candidates, where the model's task is limited to selecting the most probable answer from a predefined set \cite{wiegreffe2024answerassembleaceunderstanding}. 
\section{Further Discussion}\label{appendix:further discussion}
\subsection{Limitations of UQ in Open-ended Tasks}
 UQ in open-ended tasks primarily focuses on knowledge-intensive QA tasks, which differs fundamentally from the typical ICL paradigm.
ICL primarily relies on: pattern matching \cite{min-etal-2022-rethinking}; distribution alignment \cite{chan2022data} ;implicit fine-tuning \cite{akyureklearning}. In contrast, knowledge-intensive QA depends on retrieving from external knowledge and parametric knowledge, rather than adapting through in-context distribution learning. As a result, many-shot ICL is not well-suited for knowledge-intensive QA scenarios, making existing UQ methods for this domain inapplicable. Moreover, prior research on the performance of many-shot ICL has primarily focused on reasoning tasks and extreme-label classification \cite{li2024longcontextllmsstrugglelong}, rather than knowledge-intensive tasks.

\paragraph{Challenges in Extending UQ to Open-ended Tasks}Open-ended tasks encompass summarization, intermediate reasoning, code generation, program synthesis, and planning. However, existing UQ methods struggle to generalize effectively to these tasks, particularly in long-context ICL settings. For instance, semantic entropy \cite{kuhn2023semantic}, a widely used UQ approach, measures uncertainty based on semantic dispersion. However, in summarization tasks, summary quality is judged primarily by its fidelity to the source content, rather than semantic variability alone. This presents key limitations:
A summary may deviate semantically yet still provide a valid abstraction of the original text.
Summarization evaluation involves coverage, conciseness, and coherence, which semantic entropy alone cannot quantify.
Given these limitations, we focus on classification and multiple-choice tasks, which offer a robust evaluation framework for analyzing uncertainty evolution in long-context ICL.
\subsection{Limitations of UQ for CoT}

In CoT tasks, uncertainty accumulates throughout the reasoning process, influencing the final answer. This uncertainty propagation occurs in intermediate reasoning steps, and the final answer generation. Current UQ techniques primarily focus on single-step inference or static tasks, whereas CoT relies on multi-step reasoning. This multi-stage nature makes it difficult for existing methods to effectively capture uncertainty propagation across reasoning steps. While research on CoT uncertainty is still in its early stages, some prior works have explored possible approaches. For instance, some work proposed a stepwise scoring mechanism which assigns a confidence score to each intermediate explanation. However, this approach has notable limitations:(1) \emph{Overconfidence}: LLMs tend to be overconfident in their predictions, making single-step confidence scores unreliable; (2). \emph{Lack of global coherence}: stepwise scoring ignores dependencies across reasoning steps, failing to capture uncertainty propagation across the entire reasoning chain; (3). \emph{Step mismatch}: the reasoning steps generated may not align with the logical steps required for complex reasoning tasks, limiting the effectiveness in capturing uncertainty flow.
\paragraph{Potential Strategies: A Topological Perspective} To better model uncertainty propagation in CoT reasoning, we propose leveraging topological structures. CoT reasoning typically involves problem decomposition, backtracking and correction, evaluation and verification, and final integration. While current models generate reasoning in an autoregressive (linear) manner, actual human reasoning follows a more complex topological structure. Inspired by Tree-based CoT \cite{yao2023tree} and Graph-based CoT \cite{besta2024graph}, we propose modeling CoT uncertainty using graph or tree structures. In this framework: each reasoning step is represented as a node; uncertainty from prior steps propagates through the topological structure to influence subsequent steps; the final answer (root node) aggregates the propagated uncertainties from all previous steps. By explicitly modeling uncertainty flow in a structured manner, this approach could overcome the limitations of stepwise scoring and offer a systematic framework for analyzing uncertainty evolution in multi-step reasoning. We believe this direction holds promise for improving uncertainty estimation in CoT-based tasks.
\section{Generalization Results on Larger LLMs}
\label{appendix: generation results}
\subsection{Qwen2.5-14B-Instruct}
\begin{table}[h!]
\centering
\begin{tabular}{lcccccccc}
\toprule
\textbf{AG\_News} & \textbf{2} & \textbf{4} & \textbf{8} & \textbf{16} & \textbf{32} & \textbf{64} & \textbf{128} & \textbf{256} \\
\midrule
TU  & 0.148 & 0.127 & 0.113 & 0.125 & 0.115 & 0.105 & 0.086 & 0.065 \\
EU  & 0.057 & 0.029 & 0.030 & 0.033 & 0.038 & 0.040 & 0.028 & 0.026 \\
AU  & 0.091 & 0.098 & 0.083 & 0.092 & 0.077 & 0.065 & 0.058 & 0.039 \\
ACC & 87.6  & 87.2  & 88.9  & 88.19 & 88.7  & 88.5  & 89.9  & 90.5  \\
\bottomrule
\end{tabular}
\caption{Performance of \textbf{Qwen2.5-14B-Instruct} on AG\_News with varying numbers of in-context examples}
\label{tab:qwen2.5_icl}
\end{table}

\begin{table}[h]
\centering
\begin{tabular}{lcccccccc}
\toprule
\textbf{LD5} & \textbf{1} & \textbf{10} & \textbf{20} & \textbf{40} & \textbf{80} & \textbf{120} & \textbf{240} \\
\midrule
TU  & 0.345 & 0.307 & 0.302 & 0.257 & 0.279 & 0.272 & 0.245 \\
EU  & 0.229 & 0.194 & 0.190 & 0.148 & 0.157 & 0.139 & 0.124 \\
AU  & 0.116 & 0.112 & 0.112 & 0.109 & 0.122 & 0.133 & 0.121 \\
ACC & 62.4  & 63.6  & 64.4  & 68.4  & 67.2  & 72.1  & 72.8  \\
\bottomrule
\end{tabular}
\caption{Performance of \textbf{Qwen2.5-14B-Instruct} on LD5 with varying numbers of in-context examples}
\label{tab:ld5_icl}
\end{table}

\subsection{Qwen2.5-32B-Instruct}
\begin{table}[h]
\centering
\begin{tabular}{lcccccccc}
\toprule
\textbf{AG\_News} & \textbf{2} & \textbf{4} & \textbf{8} & \textbf{16} & \textbf{32} & \textbf{64} & \textbf{128} \\
\midrule
TU  & 0.220 & 0.171 & 0.151 & 0.099 & 0.076 & 0.060 & 0.049 \\
EU  & 0.151 & 0.093 & 0.059 & 0.030 & 0.017 & 0.020 & 0.018 \\
AU  & 0.069 & 0.078 & 0.091 & 0.068 & 0.059 & 0.040 & 0.030 \\
ACC & 88.9  & 86.8  & 87.1  & 89.5  & 89.5  & 92.4  & 92.8  \\
\bottomrule
\end{tabular}
\caption{Performance of \textbf{Qwen2.5-32B-Instruct} on AG\_News with varying numbers of in-context examples}
\label{tab:agnews_icl}
\end{table}

\begin{table}[h!]
\centering
\begin{tabular}{lccccccc}
\toprule
\textbf{LD5} & \textbf{1} & \textbf{10} & \textbf{20} & \textbf{40} & \textbf{80} & \textbf{120} \\
\midrule
TU  & 0.353 & 0.313 & 0.284 & 0.247 & 0.225 & 0.202 \\
EU  & 0.121 & 0.102 & 0.102 & 0.099 & 0.091 & 0.082 \\
AU  & 0.232 & 0.210 & 0.182 & 0.147 & 0.134 & 0.120 \\
ACC & 74.8  & 79.6  & 82.0  & 82.4  & 83.6  & 84.1  \\
\bottomrule
\end{tabular}
\caption{Performance of \textbf{Qwen2.5-32B-Instruct} on LD5 with varying numbers of in-context examples}
\label{tab:ld5_results}
\end{table}

\newpage
\section{Quality of UQ for other LLMs}\label{appendix: uq quality for other LLMs}

\begin{table}[!ht]
\centering
\resizebox{\columnwidth}{!}{%
\begin{tabular}{c|c|c|c|c|c|c|c|c}
\toprule \hline
\multirow{3}{*}{Dataset} & \multicolumn{8}{c}{\large \textbf{Llama-3.1-8B}} \\
\cline{2-9}
& \multicolumn{1}{c|}{1-shot} & \multicolumn{1}{c|}{2-shot} & \multicolumn{1}{c|}{4-shot} & \multicolumn{1}{c|}{8-shot} & \multicolumn{1}{c|}{16-shot} & \multicolumn{1}{c|}{32-shot} & \multicolumn{1}{c|}{64-shot} & \multicolumn{1}{c}{128-shot}\\
\cline{2-9}
& \multicolumn{8}{c}{Easy Mode} \\
\hline
AGNews & 0.686  & 0.704  & 0.725  & 0.735  & 0.780  & 0.804  & 0.822 & 0.837  \\
\hline
SST-2 & 0.714 & 0.751 & 0.751 &  0.748 &  0.740 &  0.741 &  0.742 &  0.750  \\
\hline
Commonsense QA & 0.563  & 0.599 & 0.636  & 0.673 & 0.726 &  0.774 &  0.784 &   0.798  \\
\hline
& \multicolumn{8}{c}{Hard Mode} \\
\hline
& \multicolumn{1}{c|}{1-shot} & \multicolumn{1}{c|}{4-shot} & \multicolumn{1}{c|}{10-shot} & \multicolumn{1}{c|}{20-shot} & \multicolumn{1}{c|}{40-shot} & \multicolumn{1}{c|}{80-shot} & \multicolumn{1}{c|}{120-shot} & \multicolumn{1}{c}{240-shot}\\
\cline{1-9}
Logical Deduction 3 & 0.973  & 0.965  & 0.939 &  0.948 & 0.951 & 0.939 &  0.966 &  0.947   \\
\hline
Logical Deduction 5 & 0.996  & 0.995  & 0.963&  0.983  & 0.971  & 0.983 & 0.959 &  0.974  \\
\hline
Logical Deduction 7 & 0.987  & 0.997  & 0.976  & 0.987  & 0.982 & 0.986 &  0.986 &  0.964  \\
\hline
\bottomrule 
\end{tabular}
}
\caption{\textbf{AUROC} of Llama-3.1-8B model.  High AUROC indicates the good quality of UQ measures. }
\label{tab:aurocs}
\end{table}

\begin{table}[!h]
\centering
\resizebox{\columnwidth}{!}{%
\begin{tabular}{c|c|c|c|c|c|c|c|c}
\toprule \hline
\multirow{3}{*}{Dataset} & \multicolumn{8}{c}{\large \textbf{Mistral-7B-v0.2}} \\
\cline{2-9}
& \multicolumn{1}{c|}{1-shot} & \multicolumn{1}{c|}{2-shot} & \multicolumn{1}{c|}{4-shot} & \multicolumn{1}{c|}{8-shot} & \multicolumn{1}{c|}{16-shot} & \multicolumn{1}{c|}{32-shot} & \multicolumn{1}{c|}{64-shot} & \multicolumn{1}{c}{128-shot}\\
\cline{2-9}
& \multicolumn{8}{c}{Easy Mode} \\
\hline
AGNews & 0.633  & 0.696  & 0.734  & 0.753  & 0.769  & 0.778  & 0.790 & 0.780  \\
\hline
SST-2 & 0.714 & 0.723 & 0.685 &  0.772 &  0.813 &  0.849 &  0.846 &  0.871  \\
\hline
Commonsense QA & 0.739  & 0.728 & 0.731 & 0.733 & 0.743 &  0.711 &  0.710 &   0.728  \\
\hline
& \multicolumn{8}{c}{Hard Mode} \\
\hline
& \multicolumn{1}{c|}{1-shot} & \multicolumn{1}{c|}{4-shot} & \multicolumn{1}{c|}{10-shot} & \multicolumn{1}{c|}{20-shot} & \multicolumn{1}{c|}{40-shot} & \multicolumn{1}{c|}{80-shot} & \multicolumn{1}{c|}{120-shot} & \multicolumn{1}{c}{240-shot}\\
\cline{1-9}
Logical Deduction 3 & 0.956  & 0.987  & 0.951 &  0.976 & 0.951 & 0.986 &  0.966 &  0.947   \\
\hline
Logical Deduction 5 & 0.938  & 0.929  & 0.918 &  0.922  & 0.934  & 0.913 & 0.918 &  0.912  \\
\hline
Logical Deduction 7 & 0.923  & 0.939  & 0.928  & 0.919  & 0.925 & 0.925 &  0.936 &  0.946  \\
\hline
\bottomrule 
\end{tabular}
}
\caption{\textbf{AUROC} of Mistral-7B-v0.2 model.  High AUROC indicates the good quality of UQ measures. }
\label{tab:aurocs2}
\end{table}
\begin{table}[!ht]
\centering
\resizebox{\columnwidth}{!}{%
\begin{tabular}{c|c|c|c|c|c|c|c|c}
\toprule \hline
\multirow{3}{*}{Dataset} & \multicolumn{8}{c}{\large \textbf{Qwen1.5-7B}} \\
\cline{2-9}
& \multicolumn{1}{c|}{1-shot} & \multicolumn{1}{c|}{2-shot} & \multicolumn{1}{c|}{4-shot} & \multicolumn{1}{c|}{8-shot} & \multicolumn{1}{c|}{16-shot} & \multicolumn{1}{c|}{32-shot} & \multicolumn{1}{c|}{64-shot} & \multicolumn{1}{c}{128-shot}\\
\cline{2-9}
& \multicolumn{8}{c}{Easy Mode} \\
\hline
AGNews & 0.634  & 0.716  & 0.743  & 0.744  & 0.688  & 0.739  & 0.731 & 0.741  \\
\hline
SST-2 & 0.742 & 0.766 & 0.842 &  0.854 &  0.872 &  0.870 &  0.870 &  0.879  \\
\hline
Commonsense QA & 0.768  & 0.818 & 0.801  & 0.801 & 0.799 &  0.776 &  0.772 &   0.770  \\
\hline
& \multicolumn{8}{c}{Hard Mode} \\
\hline
& \multicolumn{1}{c|}{1-shot} & \multicolumn{1}{c|}{4-shot} & \multicolumn{1}{c|}{10-shot} & \multicolumn{1}{c|}{20-shot} & \multicolumn{1}{c|}{40-shot} & \multicolumn{1}{c|}{80-shot} & \multicolumn{1}{c|}{120-shot} & \multicolumn{1}{c}{240-shot}\\
\cline{1-9}
Logical Deduction 3 & 0.875  & 0.846  & 0.918&  0.900 & 0.928 & 0.871 &  0.966 &  0.788   \\
\hline
Logical Deduction 5 & 0.935  & 0.918  & 0.903 &  0.849  & 0.962  & 0.934 & 0.921 &  0.912  \\
\hline
Logical Deduction 7 & 0.923  & 0.879  & 0.893  & 0.911  & 0.934 & 0.925 &  0.946 &  0.956  \\
\hline
\bottomrule 
\end{tabular}
}
\caption{\textbf{AUROC} of Qwen1.5-7B model.  High AUROC indicates the good quality of UQ measures.}
\label{tab:aurocs3}
\end{table}

\newpage
\section{Question-level Analysis}
\begin{table*}[!h]
\centering
\resizebox{\textwidth}{!}{%
\begin{tabular}{c|c|c|c|c|c|c|c|c|c|c}
\toprule \hline
\multirow{3}{*}{Dataset} & \multicolumn{10}{c}{\large \textbf{Mistral-7B-v0.2}} \\
\cline{2-11}
&  \multicolumn{2}{c|}{8-shot} & \multicolumn{2}{c|}{16-shot} & \multicolumn{2}{c|}{32-shot} & \multicolumn{2}{c|}{64-shot} & \multicolumn{2}{c}{128-shot}\\
\cline{2-11}
& $\Delta U$  & $\Delta Acc$ & $\Delta U$  & $\Delta Acc$ & $\Delta U$  & $\Delta Acc$ & $\Delta U$  & $\Delta Acc$ & $\Delta U$  & $\Delta Acc$\\
\hline
& \multicolumn{10}{c}{\textbf{Easy Mode}} \\
\hline
\multirow{2}{*}{\textbf{AG News}} & 61.4 & \textbf{+8.1} & 70.9 & \textbf{+9.5} & 77.9 & \textbf{+10.7} & 76.6 & \textbf{+11.4} & 78.8 & \textbf{+12.5} \\
\cline{2-11}
& 34.8 & \textbf{-4.9} & 25.1 & \textbf{-4.2} & 18.75 & \textbf{-3.6} & 19.15 & \textbf{-3.4} & 16.45 & \textbf{-2.8} \\
\hline
\multirow{2}{*}{\textbf{SST-2}} & 67.3 & \textbf{+9.1} & 79.2 & \textbf{+12.8} & 86.8 & \textbf{+13.7} & 87.9 & \textbf{+13.8} & 90.0 & \textbf{+14.6} \\
\cline{2-11}
 & 27.3 & \textbf{-0.7} & 16.3 & \textbf{-0.0} & 10.8 & \textbf{-0.5} & 9.7 & \textbf{-0.3} & 8.6 & \textbf{-0.3} \\
\hline
\multirow{2}{*}{\textbf{Commonsense QA}} & 49.8 & \textbf{+1.8} & 40.0 & \textbf{+1.4} & 37.2 &  \textbf{+3.4} & 38 &  \textbf{+1.6} & 36.0 & \textbf{+2.0} \\
\cline{2-11}
 & 21.8 & \textbf{+0.4} & 20.2 & \textbf{+0.4} & 18.4 &  \textbf{-1.4} & 19.2 &  \textbf{+0.4} & 23.6 & \textbf{-0.8} \\
\hline
& \multicolumn{10}{c}{\textbf{Hard Mode}} \\
\cline{2-11}
& \multicolumn{2}{c|}{20-shot} & \multicolumn{2}{c|}{40-shot} & \multicolumn{2}{c|}{80-shot} & \multicolumn{2}{c|}{120-shot} & \multicolumn{2}{c}{240-shot}\\
\cline{2-11}
& $\Delta U$  & $\Delta Acc$ & $\Delta U$  & $\Delta Acc$ & $\Delta U$  & $\Delta Acc$ & $\Delta U$  & $\Delta Acc$ & $\Delta U$  & $\Delta Acc$\\
 \hline
\multirow{2}{*}{\textbf{Logical Deduction3}} & 80.4 & \textbf{+13.6} & 78.4 & \textbf{+20.4} & 79.3&  \textbf{+20.5} & 76.8 &  \textbf{+18.4} & 81.6 & \textbf{+20.4} \\
\cline{2-11}
 & 9.6 & \textbf{-0.4} & 9.2 & \textbf{-0.8} & 44.4 &  \textbf{-6.0} & 12.4 &  \textbf{-0.4} & 9.6 & \textbf{-1.6} \\
 \hline
\multirow{2}{*}{\textbf{Logical Deduction5}}& 31.6 &\textbf{+0.4} & 39.6 & \textbf{+0.8} & 48.12 &  \textbf{+0.0} & 57.9 &  \textbf{+0.0} & 79.2 & \textbf{+1.5} \\
\cline{2-11}
& 41.2 &\textbf{-0.4} & 36.8 & \textbf{-0.8} & 33.0 &  \textbf{-0.9} & 24.0 &  \textbf{-0.8} & 15.4 & \textbf{-0.5} \\
 \hline
 \multirow{2}{*}{\textbf{Logical Deduction7}} & 46.8 & \textbf{+0.0} & 60 & \textbf{+0.0} & 62.8 &  \textbf{+0.0} & 47.2 & \textbf{+0.0} & 96.5 & \textbf{+0.0} \\
\cline{2-11}
& 38.8 & \textbf{-0.0} & 25.2& \textbf{-0.0} & 28.4 &  \textbf{-0.0} & 34.8 & \textbf{-0.0} & 1.17 & \textbf{-0.0} \\
\bottomrule 
\end{tabular}
}
\caption{$\Delta$ U denotes the proportion of datasets whose uncertainty decreases/increases compared to $4$-shot settings, with the first line for each dataset giving the ratio of decreased uncertainty questions and the second line for each dataset giving the ratio of increased uncertainty questions. $\Delta Acc$ represents the performance changes caused by the corresponding part of examples.}
\label{tab:main_tab2}
\end{table*}
\begin{table*}[!ht]
\centering
\resizebox{\textwidth}{!}{%
\begin{tabular}{c|c|c|c|c|c|c|c|c|c|c}
\toprule \hline
\multirow{3}{*}{Dataset} & \multicolumn{10}{c}{\large \textbf{Qwen1.5-7B}} \\
\cline{2-11}
&  \multicolumn{2}{c|}{8-shot} & \multicolumn{2}{c|}{16-shot} & \multicolumn{2}{c|}{32-shot} & \multicolumn{2}{c|}{64-shot} & \multicolumn{2}{c}{128-shot}\\
\cline{2-11}
& $\Delta U$  & $\Delta Acc$ & $\Delta U$  & $\Delta Acc$ & $\Delta U$  & $\Delta Acc$ & $\Delta U$  & $\Delta Acc$ & $\Delta U$  & $\Delta Acc$\\
\hline
& \multicolumn{10}{c}{\textbf{Easy Mode}} \\
\hline
\multirow{2}{*}{\textbf{AG News}} & 62.2 & \textbf{+3.7} & 40.4 & \textbf{-0.8} & 79.5 & \textbf{+9.3} & 83.9 & \textbf{+9.8} & 83.8 & \textbf{+10.0} \\
\cline{2-11}
& 34.4 & \textbf{-0.8} & 52.0 & \textbf{-10.0} & 17.25 & \textbf{-0.6} & 13.0 & \textbf{-0.5} & 12.6 & \textbf{-1.4} \\
\hline
\multirow{2}{*}{\textbf{SST-2}} & 78.0 & \textbf{+0.2} & 86.6 & \textbf{+0.0} & 82.9 & \textbf{+0.1} & 77.3 & \textbf{-0.5} & 71.8 & \textbf{-0.3} \\
\cline{2-11}
 & 13.0 & \textbf{-0.5} & 4.3 & \textbf{-0.1} & 2.9 & \textbf{-0.5} & 1.8 & \textbf{-0.1} & 2.4 & \textbf{-0.3} \\
\hline
\multirow{2}{*}{\textbf{Commonsense QA}} & 33.6 & \textbf{+0.0} & 12.2 & \textbf{+0.4} & 11.0 &  \textbf{+0.2} & 8.9 &  \textbf{+0.81} & 15.0 & \textbf{+0.4} \\
\cline{2-11}
 & 48.1 & \textbf{-3.6} & 68.6 & \textbf{-10.4} & 76.4 &  \textbf{-14.4} & 79.7 &  \textbf{-1.6} & 72.6 & \textbf{-2.4} \\
\hline
& \multicolumn{10}{c}{\textbf{Hard Mode}} \\
\cline{2-11}
& \multicolumn{2}{c|}{20-shot} & \multicolumn{2}{c|}{40-shot} & \multicolumn{2}{c|}{80-shot} & \multicolumn{2}{c|}{120-shot} & \multicolumn{2}{c}{240-shot}\\
\cline{2-11}
& $\Delta U$  & $\Delta Acc$ & $\Delta U$  & $\Delta Acc$ & $\Delta U$  & $\Delta Acc$ & $\Delta U$  & $\Delta Acc$ & $\Delta U$  & $\Delta Acc$\\
 \hline
\multirow{2}{*}{\textbf{Logical Deduction3}} & 13.6 & \textbf{+2.0} & 13.2 & \textbf{+1.6} & 9.2&  \textbf{+0.4} & 20.8 &  \textbf{+1.6} & 36.0 & \textbf{+6.0} \\
\cline{2-11}
 & 84 & \textbf{-13.6} & 82.4 & \textbf{-16.0} & 88.4 &  \textbf{-22.8} & 77.2 &  \textbf{-16.8} & 60.8 & \textbf{-10.4} \\
 \hline
\multirow{2}{*}{\textbf{Logical Deduction5}}& 73.6 &\textbf{+7.19} & 74.8 & \textbf{+3.6} & 50.8 &  \textbf{+0.0} & 49.2 &  \textbf{+0.0} & 69.2 & \textbf{+6.4} \\
\cline{2-11}
& 24.0 &\textbf{-2.8} & 24.4 & \textbf{-3.6} & 47.2 &  \textbf{-12.0} & 47.6 &  \textbf{-11.2} & 15.4 & \textbf{-5.6} \\
 \hline
 \multirow{2}{*}{\textbf{Logical Deduction7}} & 81.2 & \textbf{+4.8} & 54.4 & \textbf{+2.0} & 44.0 &  \textbf{-0.4} & 41.1 & \textbf{-1.6} & 52.5 & \textbf{+10.3} \\
\cline{2-11}
& 17.6 & \textbf{-0.8} & 100 & \textbf{-20.8} & 54.8 &  \textbf{-11.6} & 58.9 & \textbf{-16.1} & 45.3 & \textbf{-9.2} \\
\bottomrule 
\end{tabular}
}
\caption{$\Delta$ U denotes the proportion of datasets whose uncertainty decreases/increases compared to $4$-shot settings, with the first line for each dataset giving the ratio of decreased uncertainty questions and the second line for each dataset giving the ratio of increased uncertainty questions. $\Delta Acc$ represents the performance changes caused by the corresponding part of examples.}
\label{tab:main_tab3}
\end{table*}

\newpage
\section{Interprebility for k-shot ICL}\label{Interprebility for k-shot IC}
\subsection{Case Study} \label{E: case study}
\vspace{-.5cm}
\begin{figure*}[!h]
    \centering
    \begin{subfigure}[b]{0.45\textwidth}
        \centering
        \includegraphics[width=\textwidth]{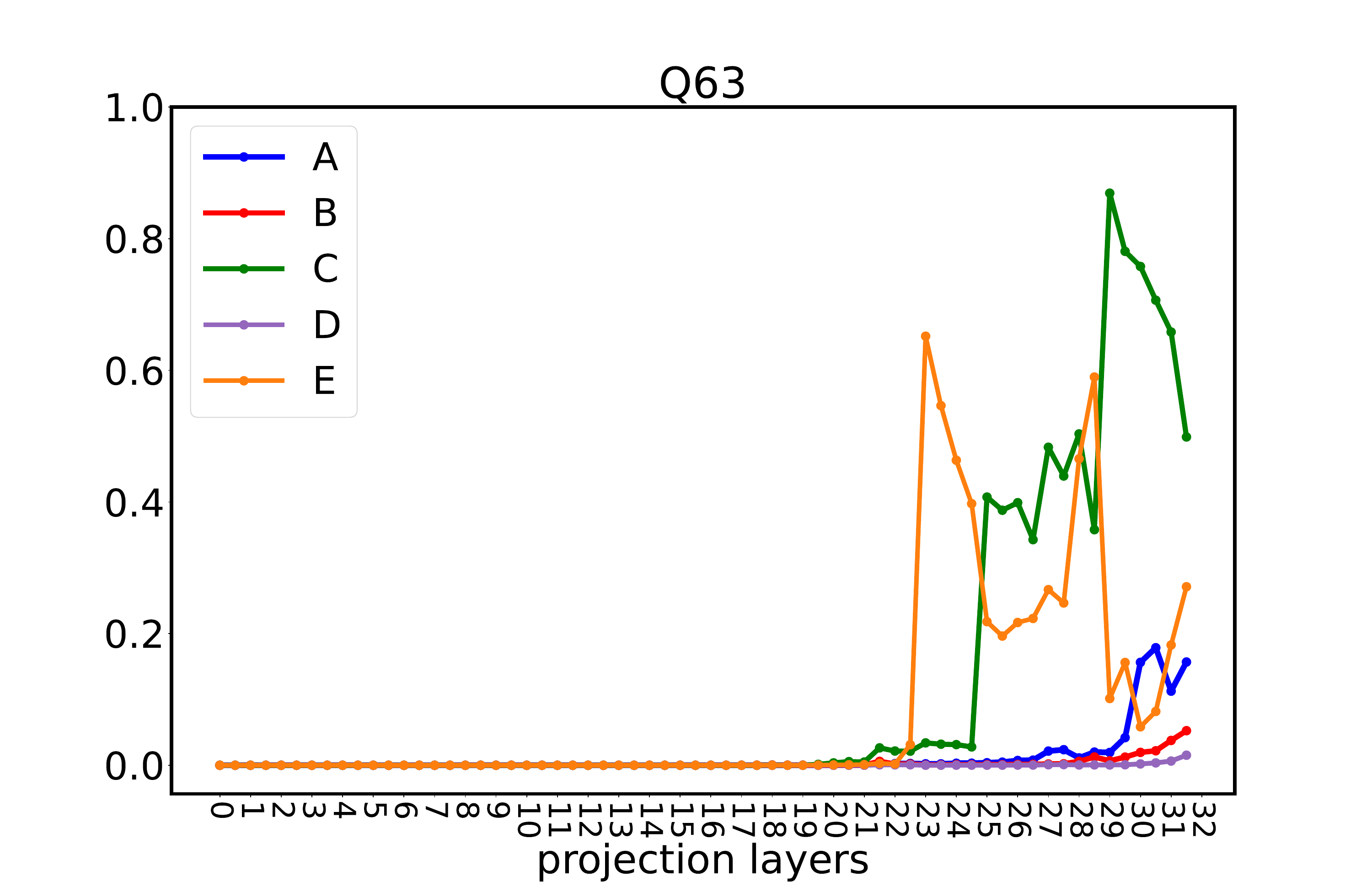}
        \vspace{-.1cm}
        \caption{4-shot}
    \end{subfigure}
    \hspace{-.5cm} 
    \begin{subfigure}[b]{0.45\textwidth}
        \centering
        \includegraphics[width=\textwidth]{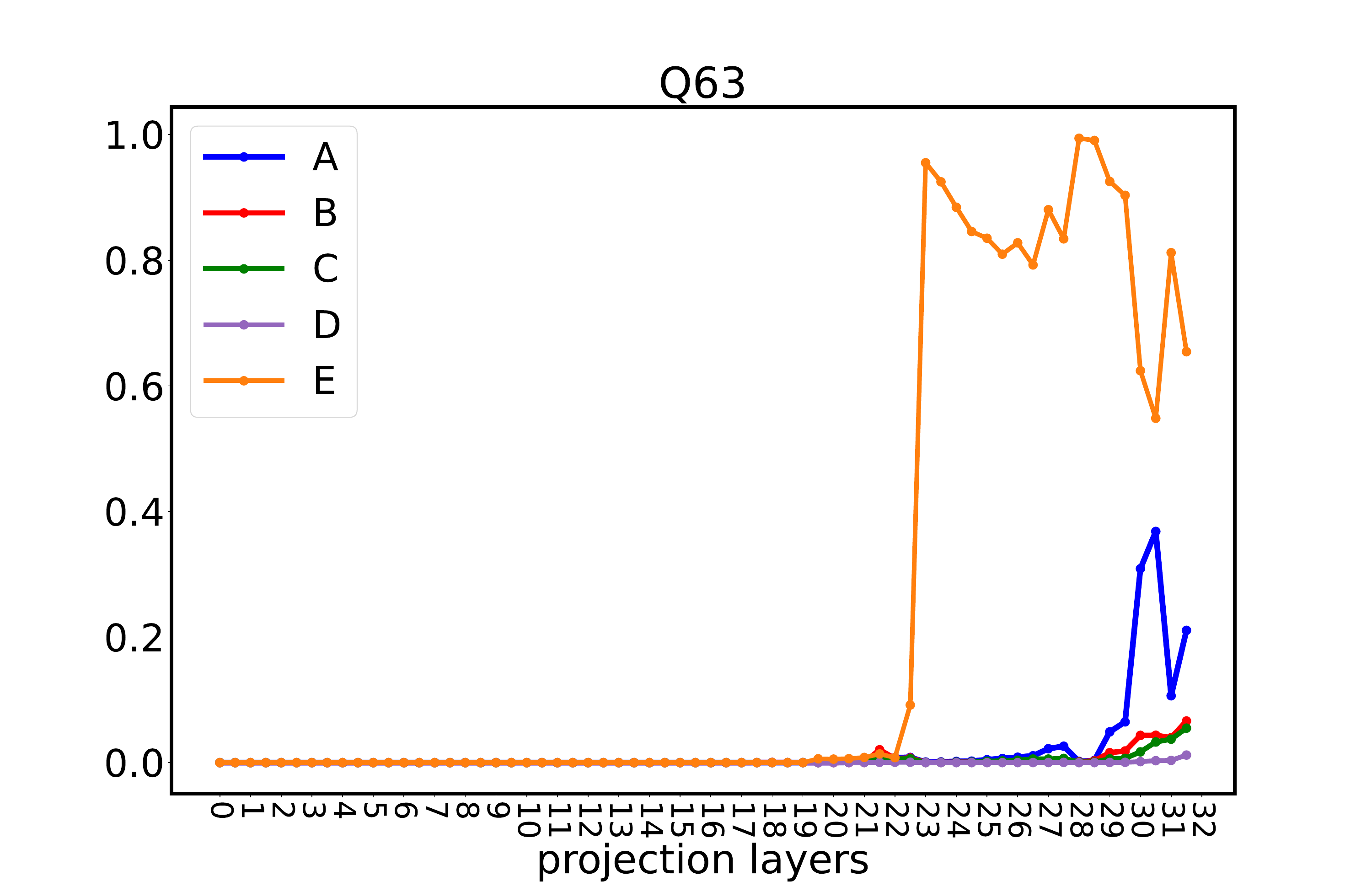}
        \vspace{-.1cm}
        \caption{32-shot}
    \end{subfigure}
    \vspace{-.3cm}
    \label{fig:case_1}
\end{figure*}
\vspace{-0.7cm}
\begin{figure*}[!h]
    \centering
    \begin{subfigure}[b]{0.45\textwidth}
        \centering
        \includegraphics[width=\textwidth]{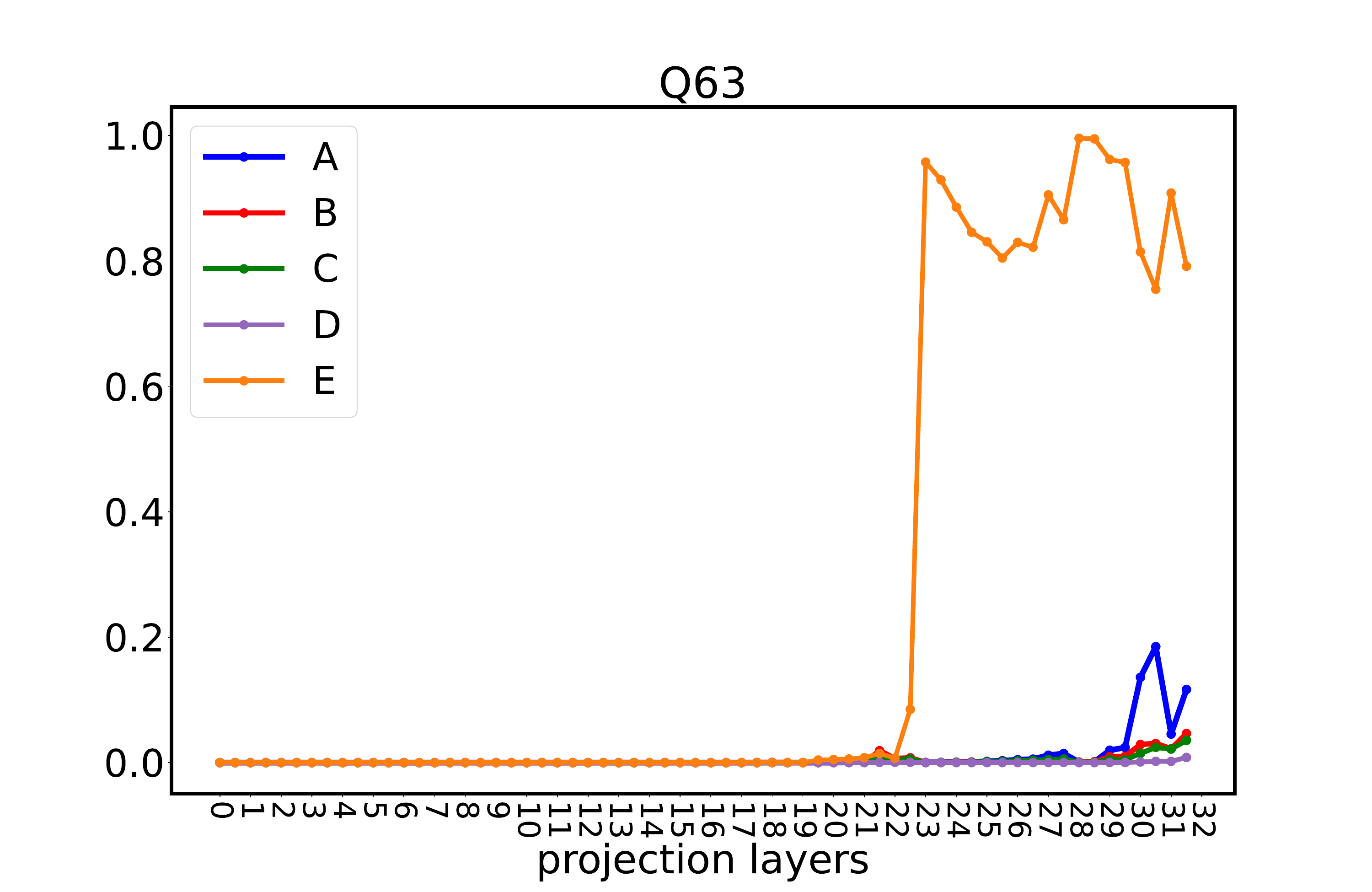}
        \vspace{-.1cm}
        \caption{64-shot}
    \end{subfigure}
    \hspace{-.5cm} 
    \begin{subfigure}[b]{0.45\textwidth}
        \centering
        \includegraphics[width=\textwidth]{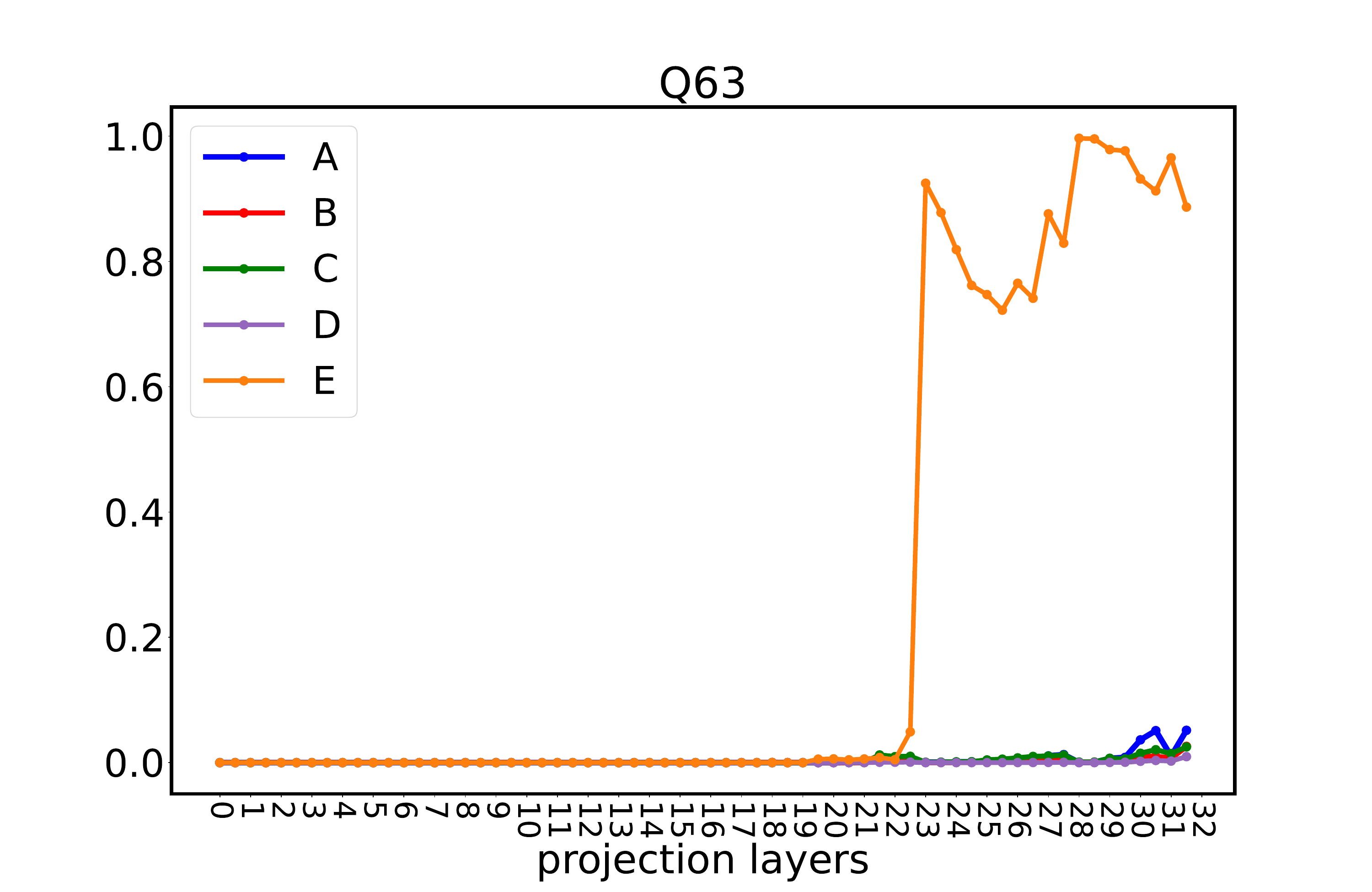}
        \vspace{-.1cm}
        \caption{128-shot}
    \end{subfigure}
    \vspace{-.1cm}
    \caption{The inner confidence changes (0-1 probability) of five options ["A", "B", "C", "D", "E"] for a specific question in Commonsense QA for Mistral-7B-v0.2 under 4-shot (a), 32-shot (b), 64-shot (c), and 128-shot(d) ICL. \textbf{The correct option is "E"} and LLMs only made a mistake under 4-shot ICL and got correct with more examples.}
    \label{fig:case_2}
\end{figure*}
\vspace{-.5cm}
\subsection{Additional results: Average logits and probabilities}\label{Additional results: Average logits and probabilities}
\vspace{-.5cm}
\begin{figure*}[!h]
    \centering
    \begin{subfigure}[b]{0.245\textwidth}
        \centering
        \includegraphics[width=\textwidth]{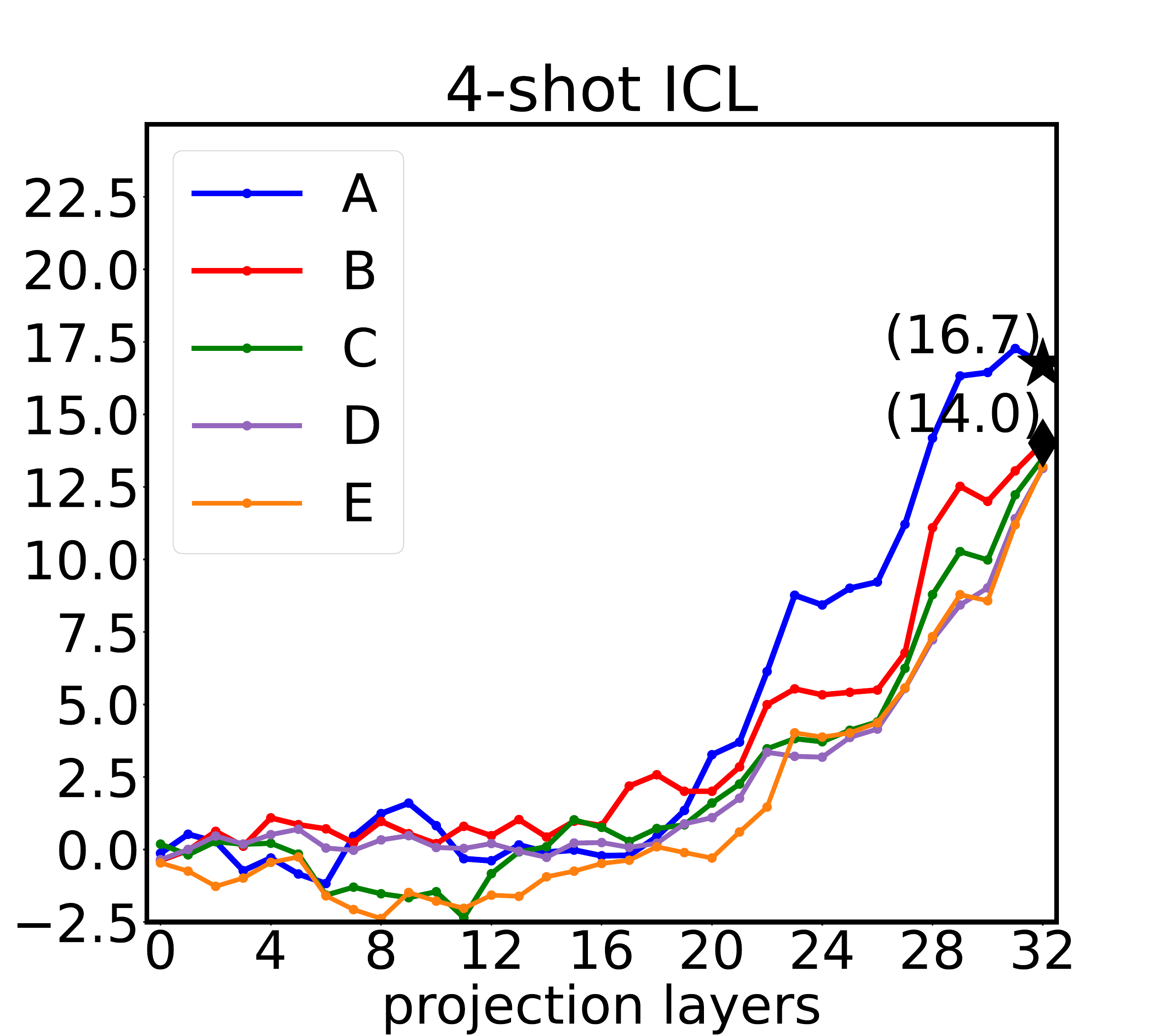}
    \end{subfigure}
    \hspace{-.2cm}
    \begin{subfigure}[b]{0.245\textwidth}
        \centering
        \includegraphics[width=\textwidth]{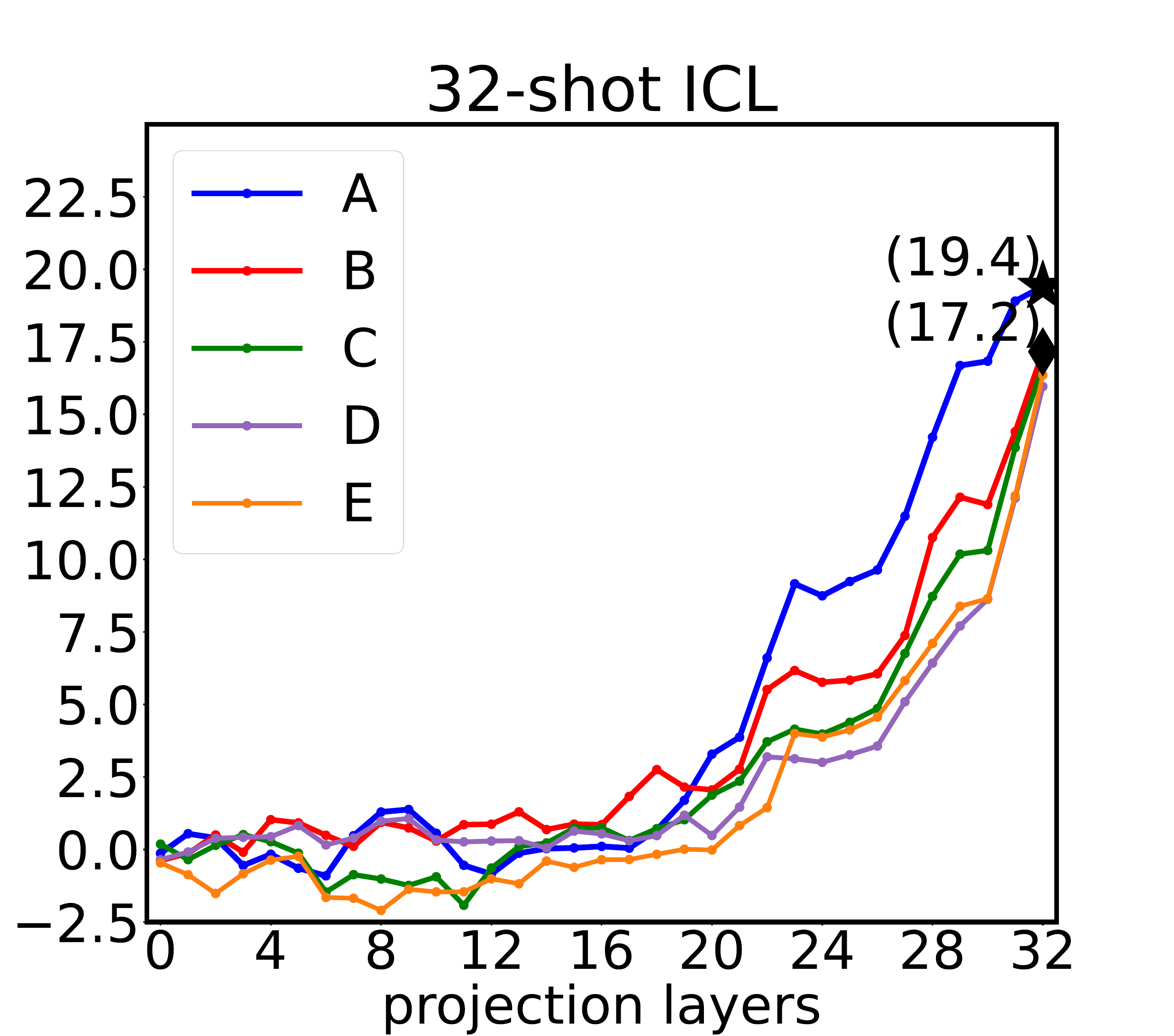}
    \end{subfigure}
    \hspace{-.2cm}
    \begin{subfigure}[b]{0.245\textwidth}
        \centering
        \includegraphics[width=\textwidth]{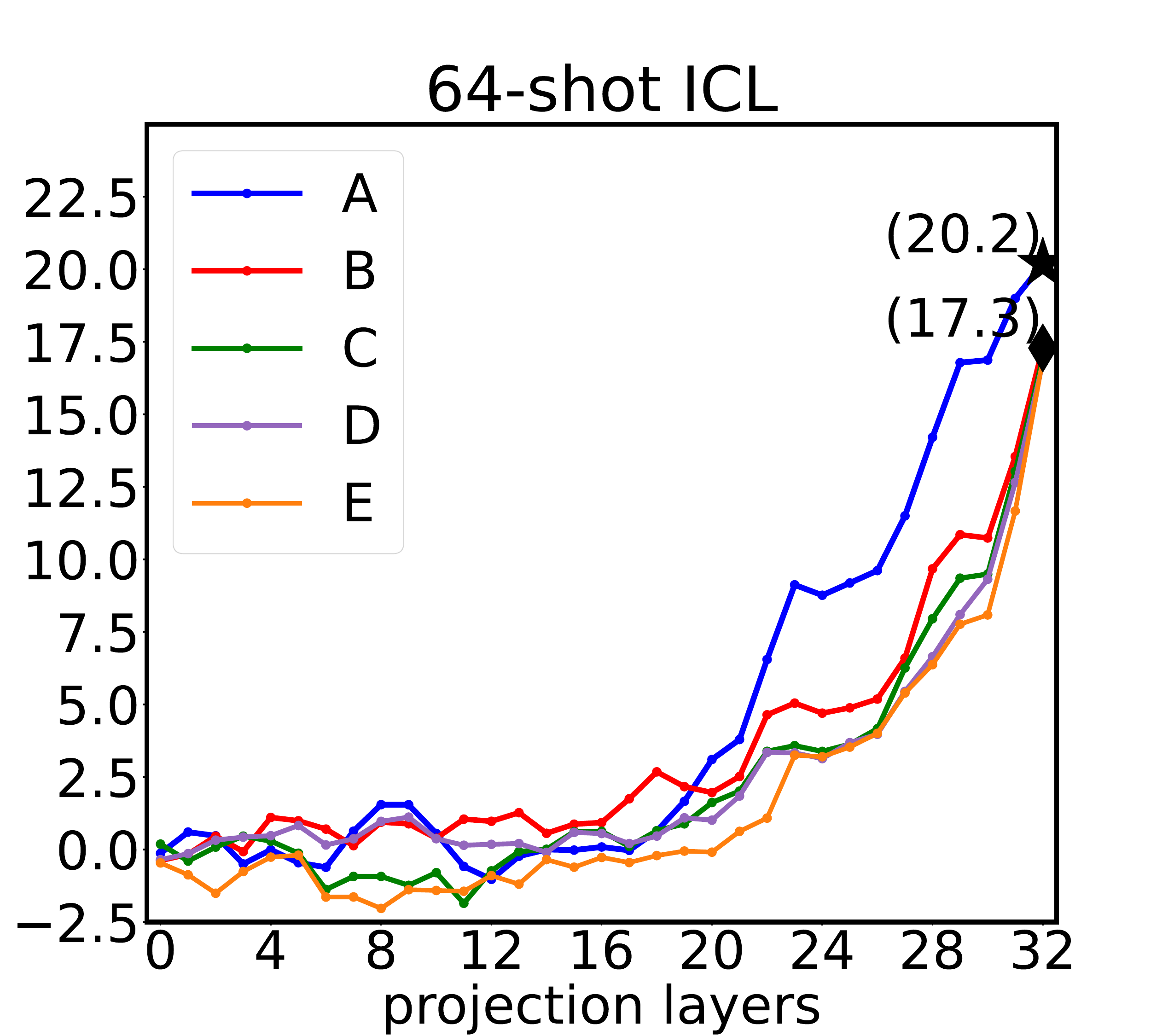}
    \end{subfigure}
    \hspace{-.2cm}
    \begin{subfigure}[b]{0.245\textwidth}
        \centering
        \includegraphics[width=\textwidth]{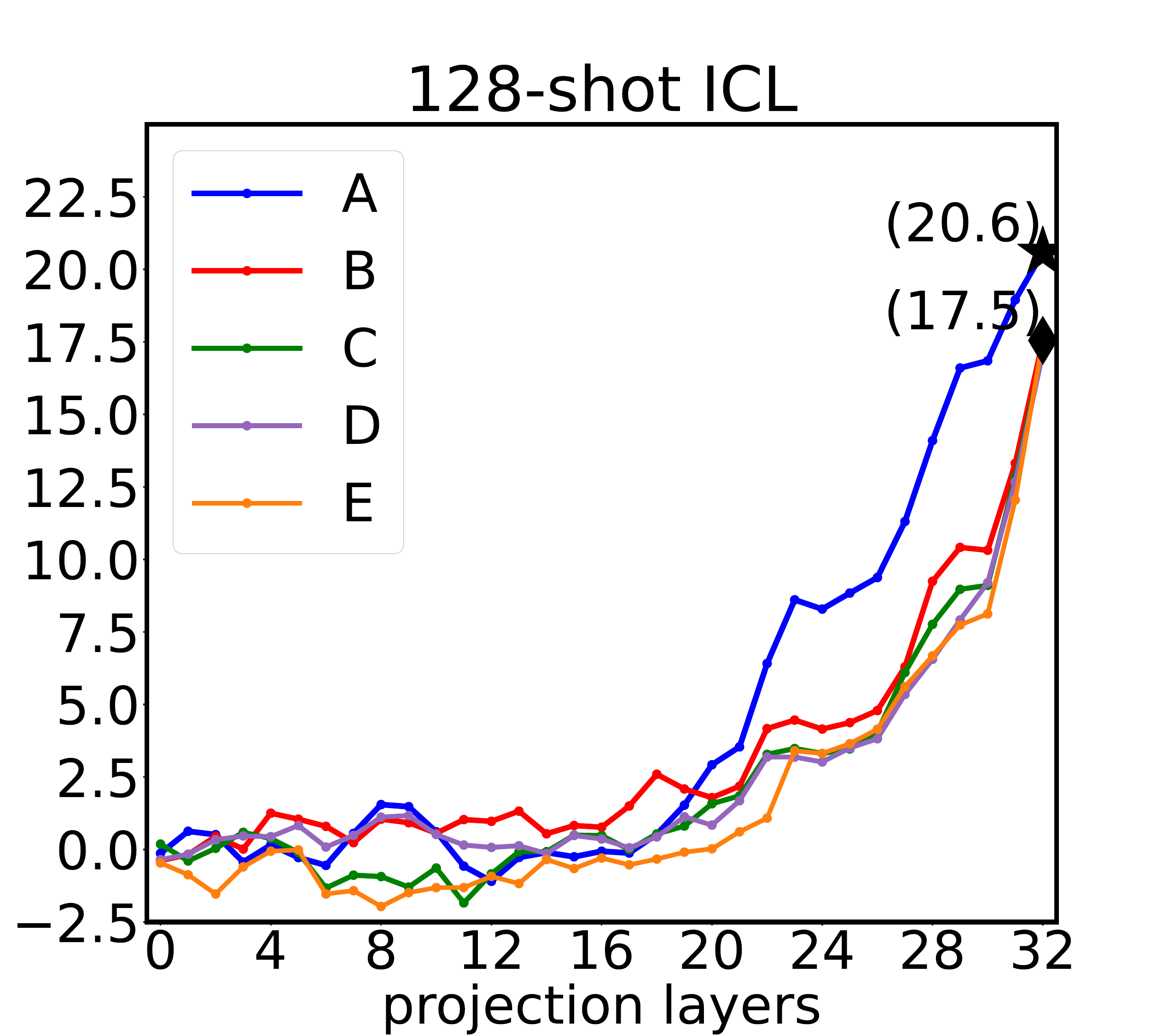}
    \end{subfigure}
    \label{fig: case study 2 logits}
\end{figure*}
\vspace{-.4cm}
\begin{figure*}[!h]
    \centering
    \begin{subfigure}[b]{0.245\textwidth}
        \centering
        \includegraphics[width=\textwidth]{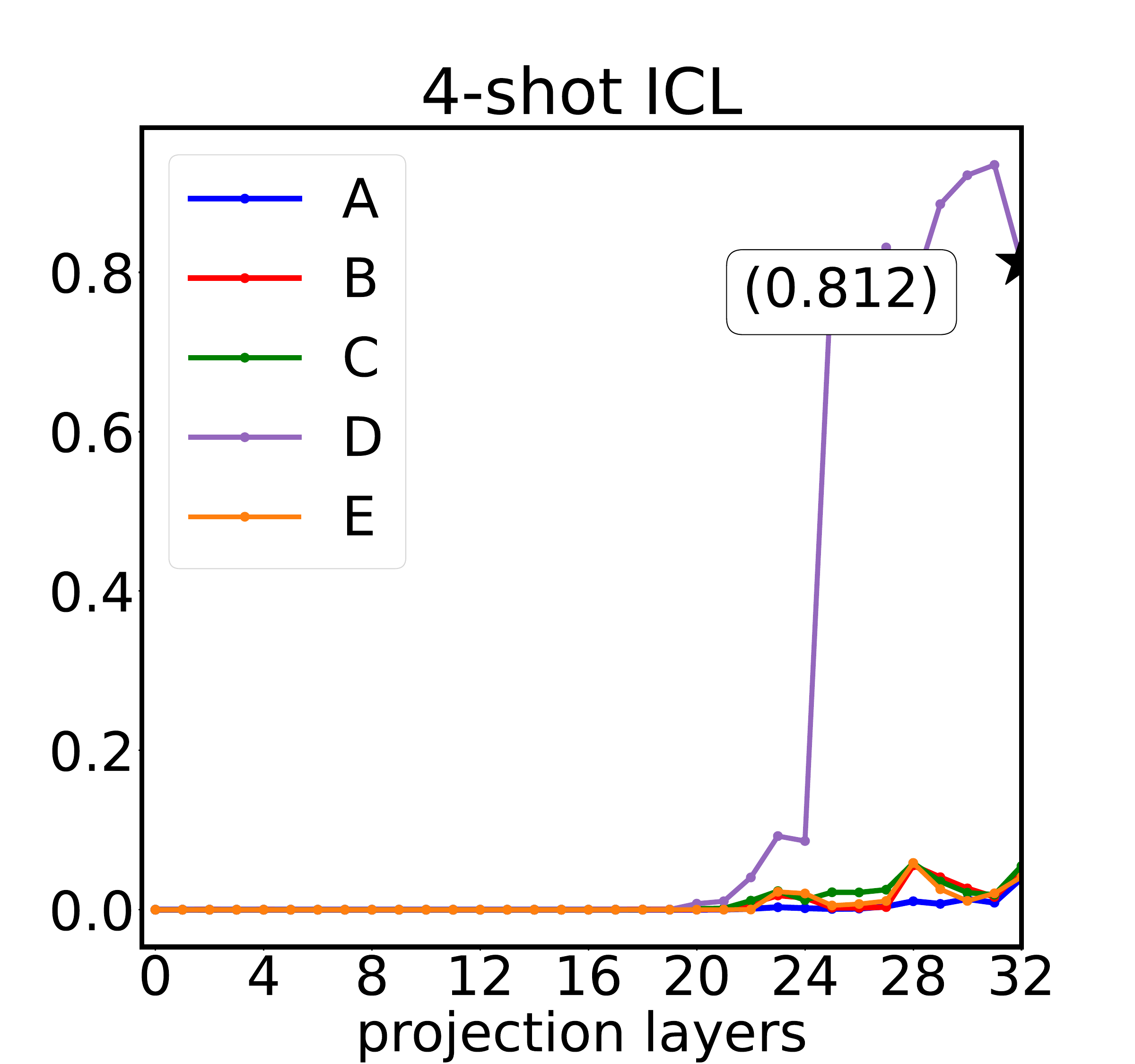}
    \end{subfigure}
    \hspace{-.2cm}
    \begin{subfigure}[b]{0.245\textwidth}
        \centering
        \includegraphics[width=\textwidth]{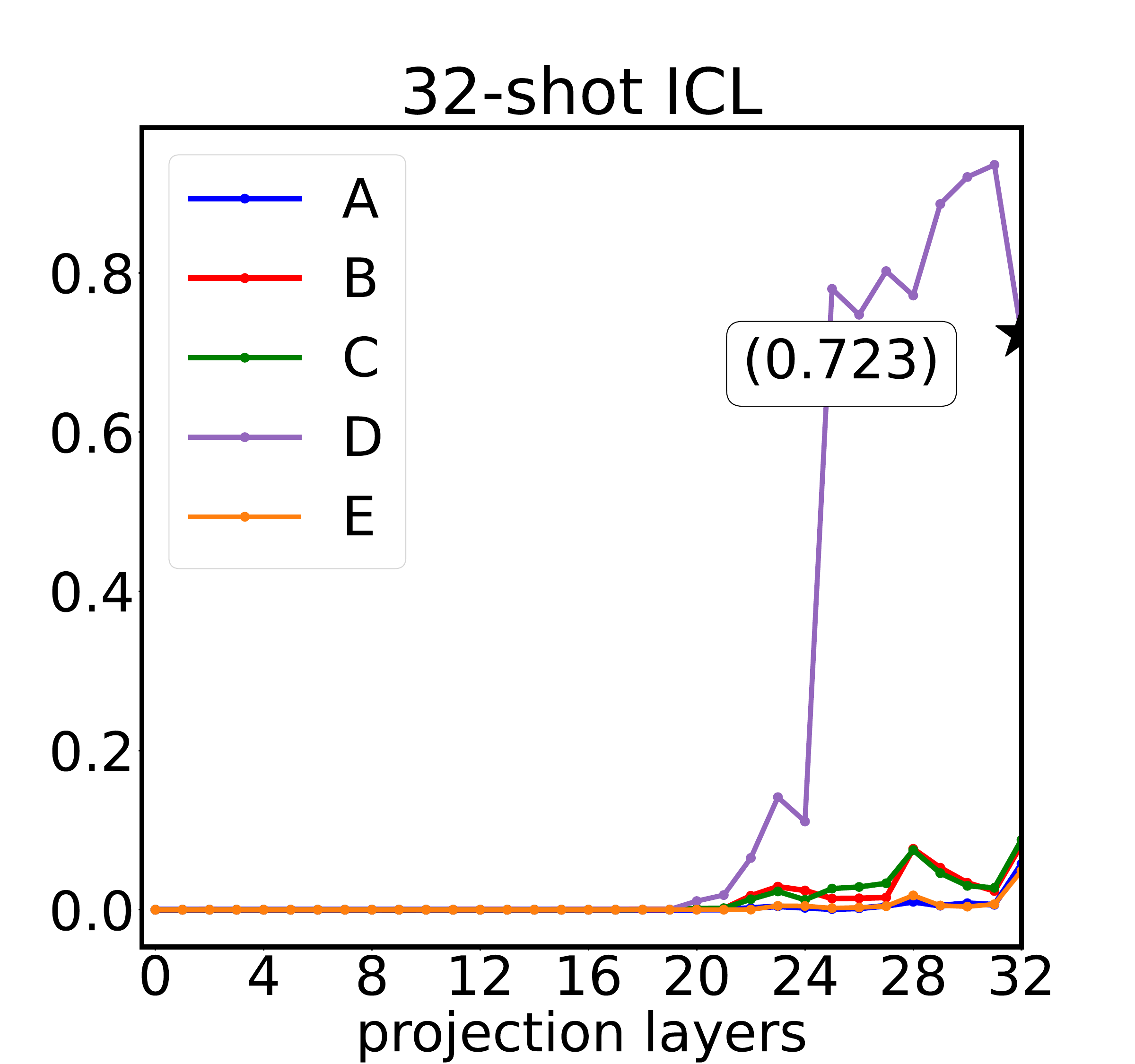}
    \end{subfigure}
    \hspace{-.2cm}
    \begin{subfigure}[b]{0.245\textwidth}
        \centering
        \includegraphics[width=\textwidth]{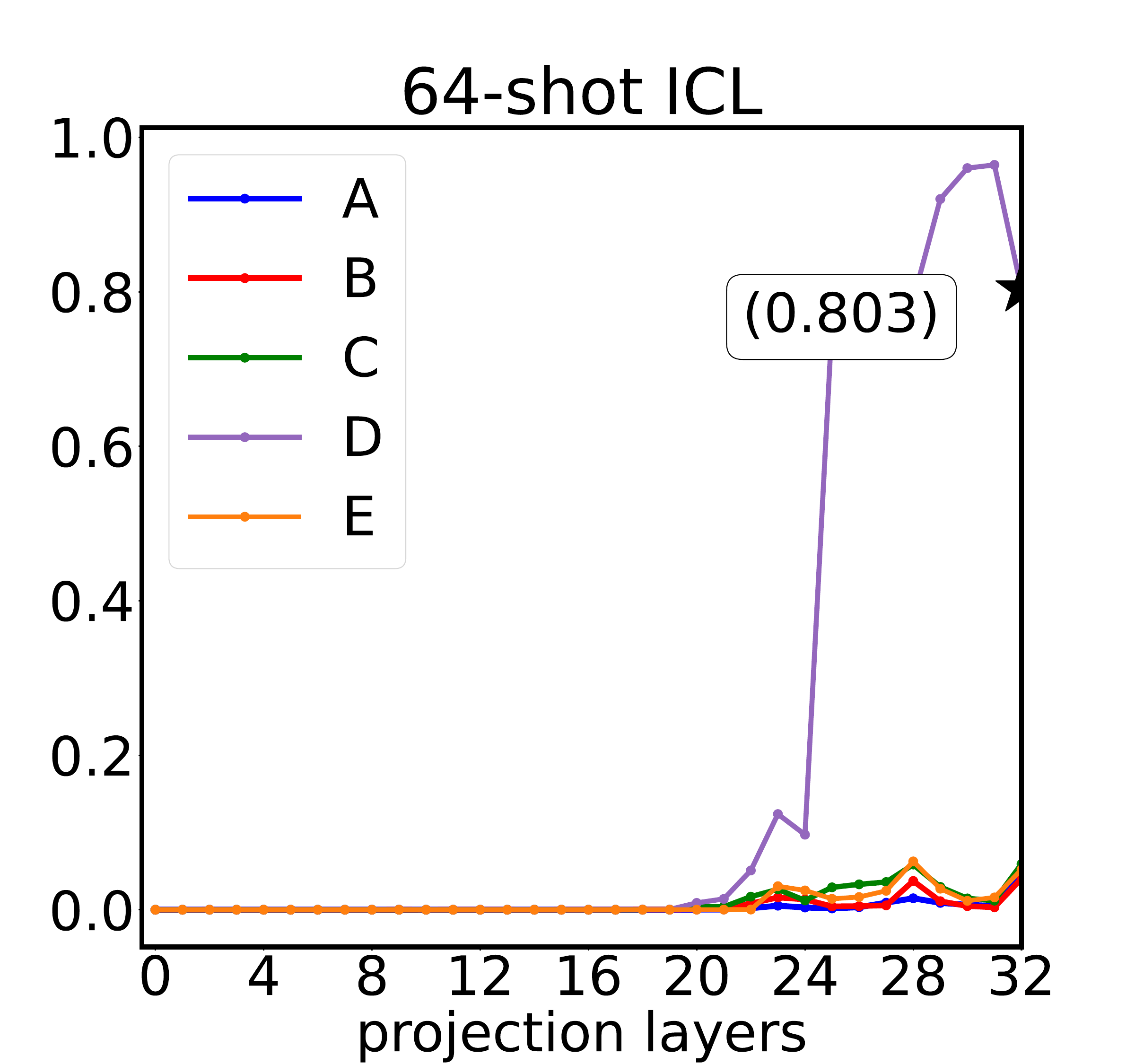}
    \end{subfigure}
    \hspace{-.2cm}
    \begin{subfigure}[b]{0.245\textwidth}
        \centering
        \includegraphics[width=\textwidth]{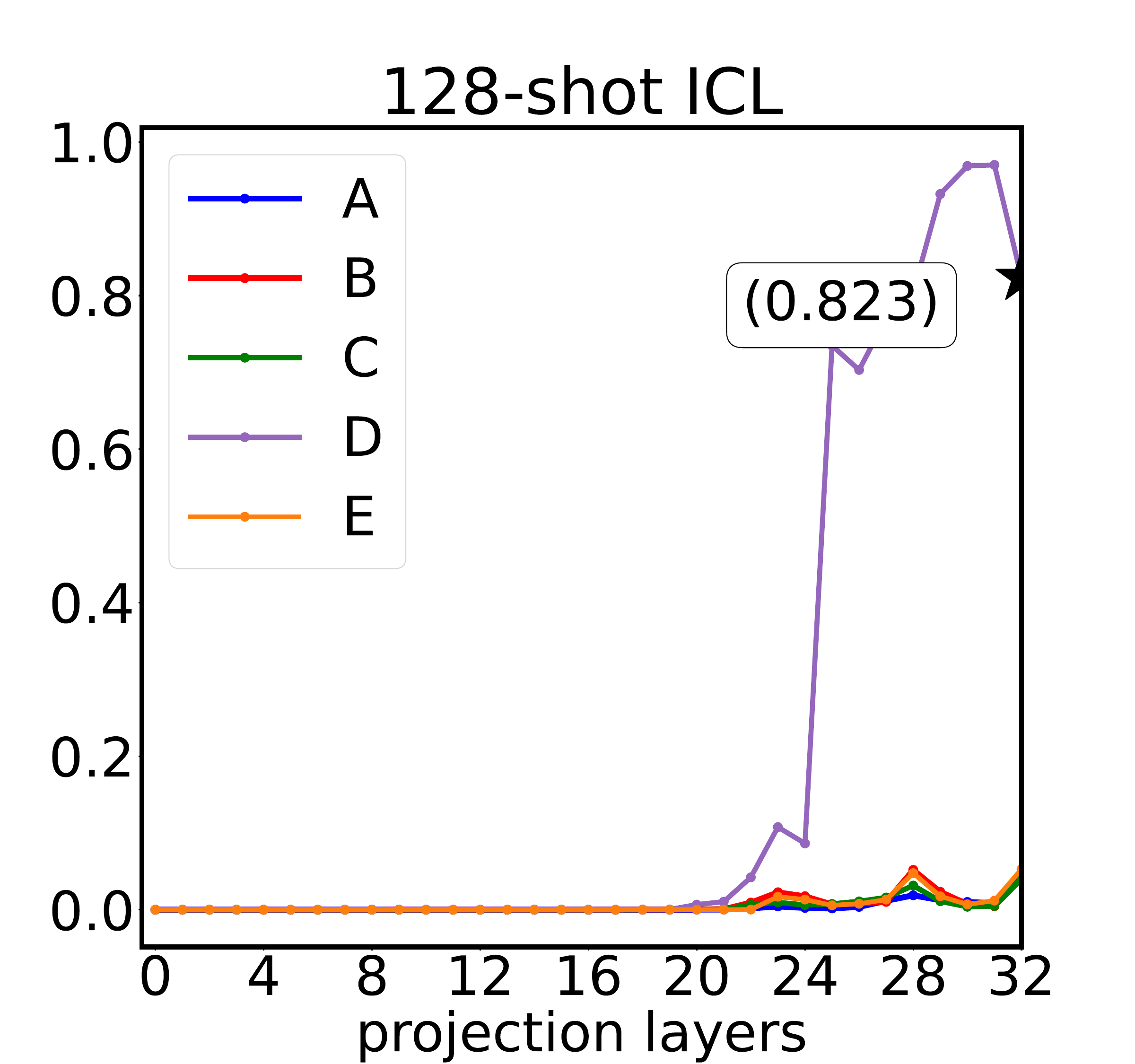}
    \end{subfigure}
    \caption{Average logits and probabilities of Mistral-7B-v0.2 on the Commonsense QA dataset for MCQA items where the correct answer is "A".}
    \label{fig: case study 2 probs}
\end{figure*}

\newpage
\section{AI Assistant Usage}
We used \textit{chatgpt} to assist with correcting spelling errors in writing .
\section{Experimental Details}
\subsection{Prompt templates}\label{appedix:prompt template}
\begin{figure*}[!h]
    \centering
    \includegraphics[width=.98\textwidth]{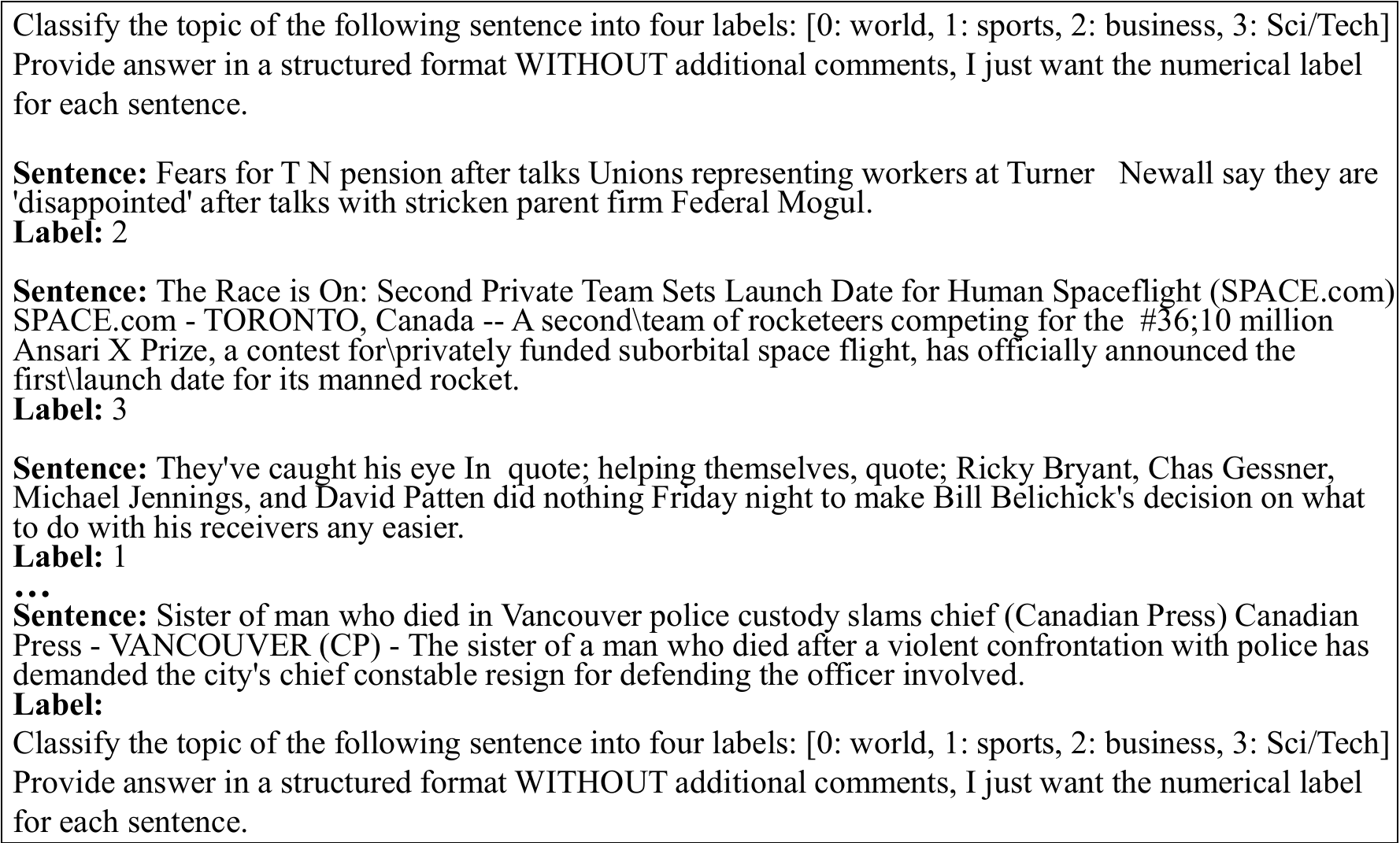} 
    \caption{Prompt template with a test input for AG News dataset.}
    \label{fig: agnews prompts}
\end{figure*}

\begin{figure*}[!h]
    \centering
    \includegraphics[width=.98\textwidth]{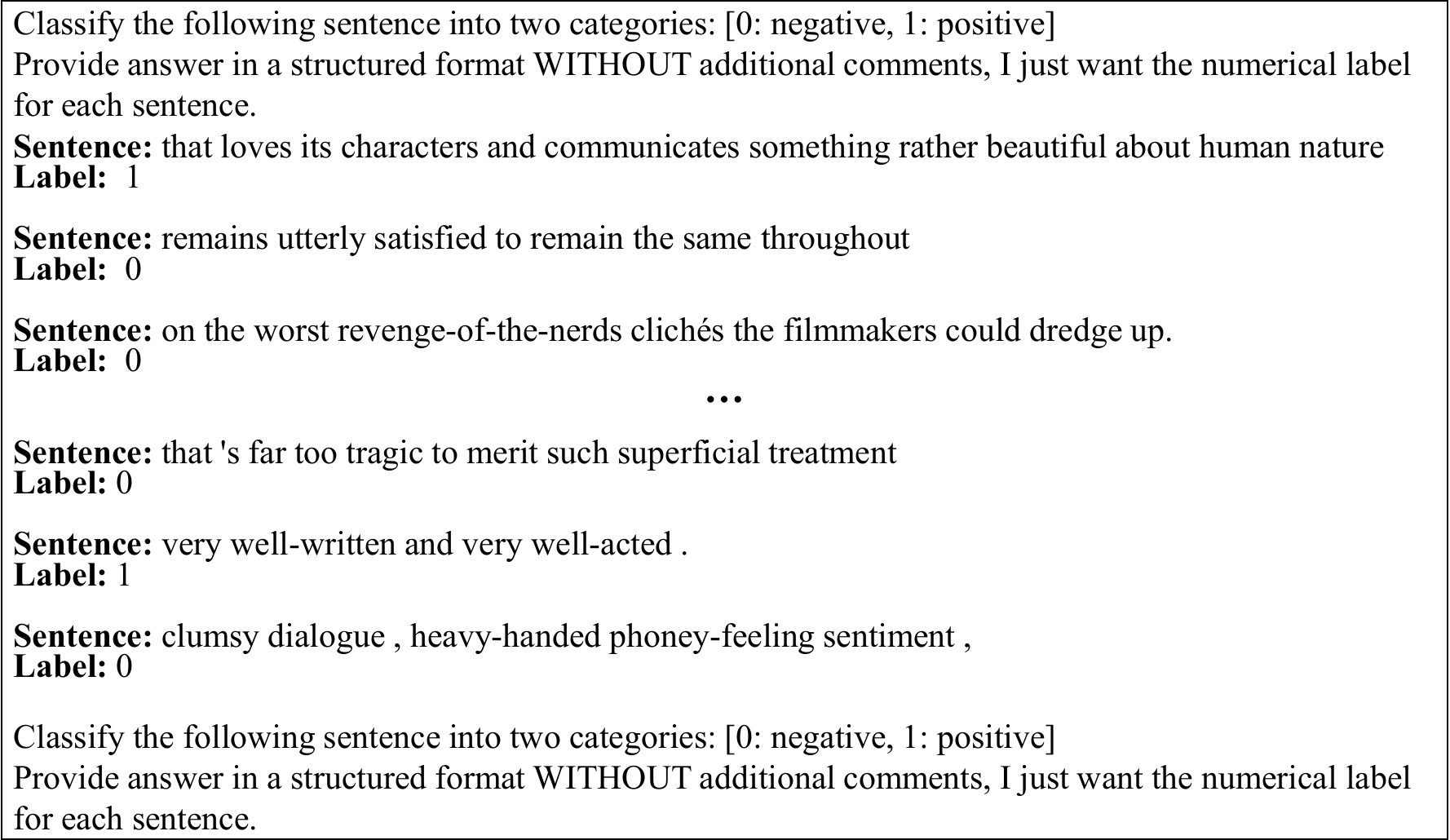} 
    \caption{Prompt template with a test input for SST-2 dataset.}
    \label{fig: sst2 prompts}
\end{figure*}
\begin{figure*}[!h]
    \centering
    \includegraphics[width=.98\textwidth]{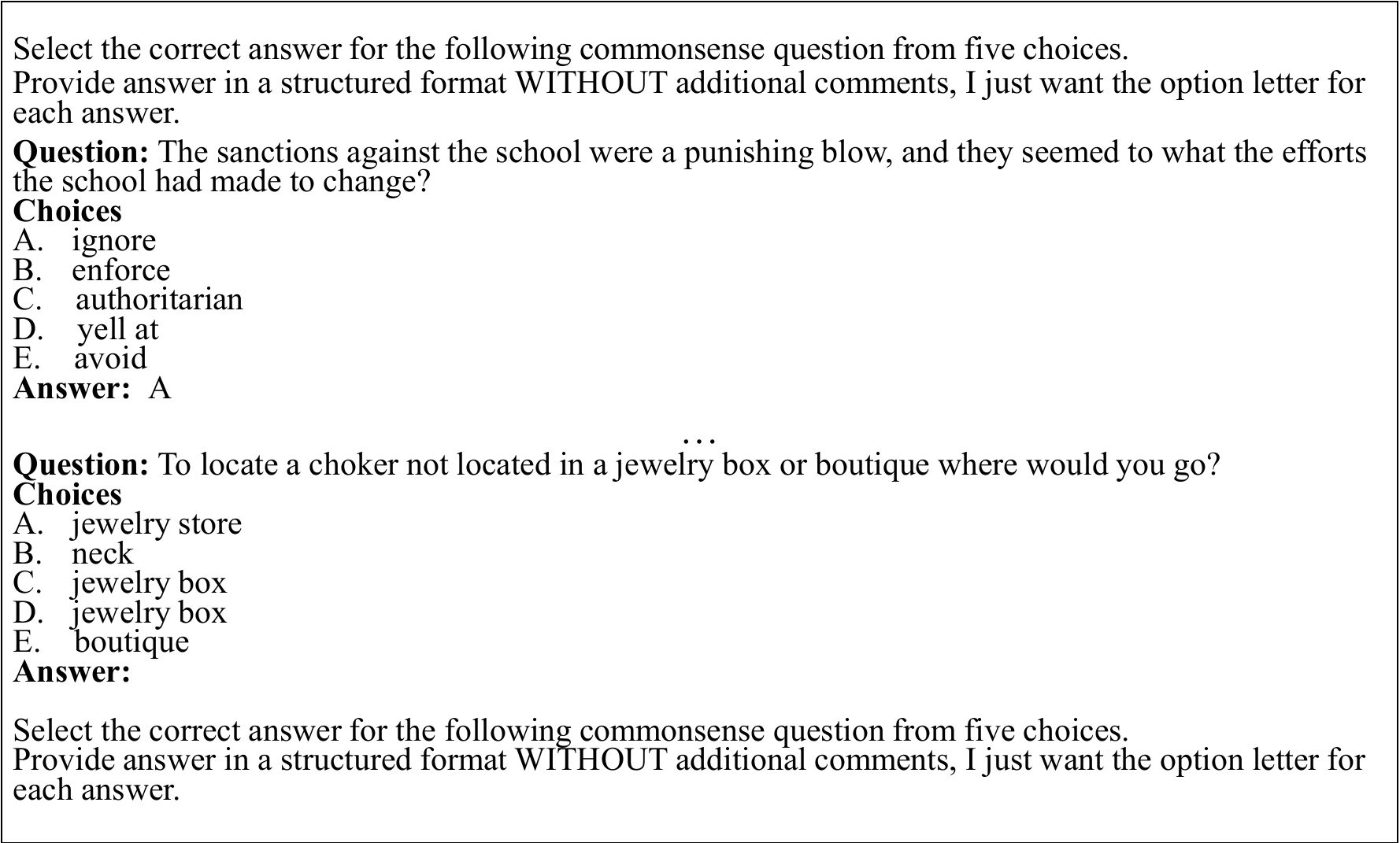} 
    \caption{Prompt template with a test input for Commonsense QA dataset.}
    \label{fig: cmqa prompts}
\end{figure*}
\begin{figure*}[!h]
    \centering
    \includegraphics[width=.98\textwidth]{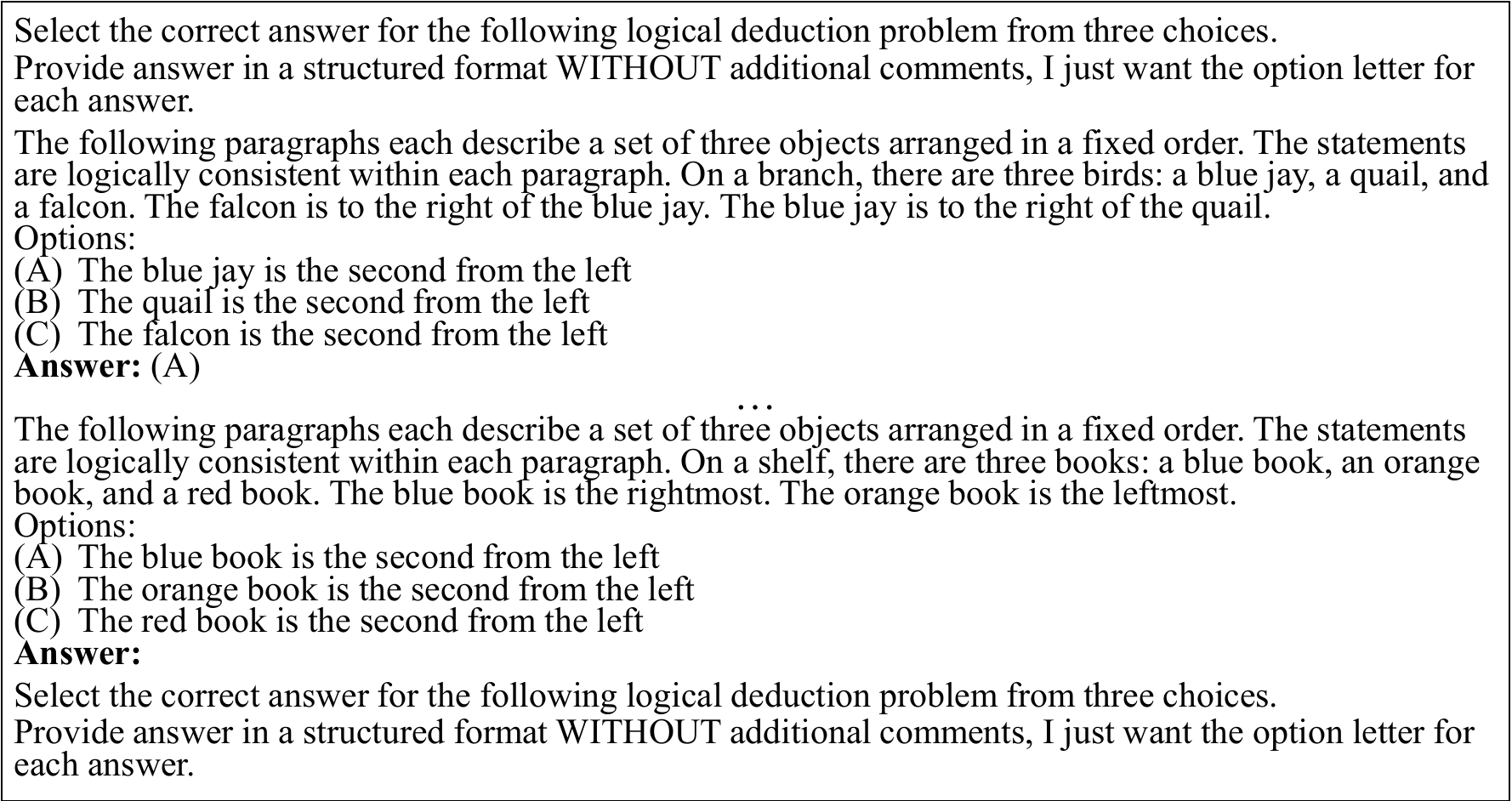} 
    \caption{Prompt template with a test input for logical deduction three objects dataset.}
    \label{fig: ld3 prompts}
\end{figure*}
\begin{figure*}[!h]
    \centering
    \includegraphics[width=.98\textwidth]{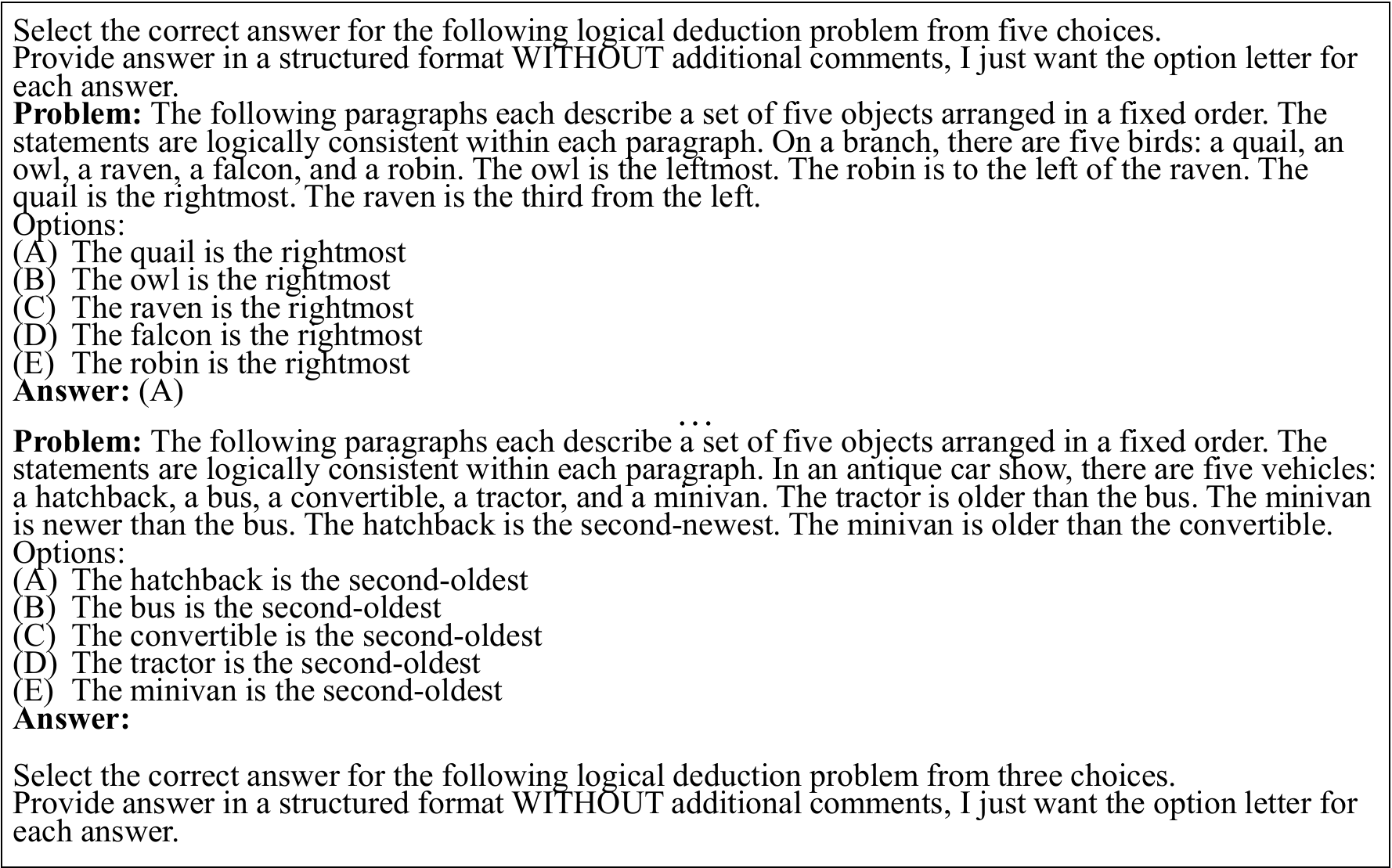} 
    \caption{Prompt template with a test input for logical deduction five objects dataset.}
    \label{fig: ld5 prompts}
\end{figure*}
\begin{figure*}[!ht]
    \centering
    \includegraphics[width=.98\textwidth]{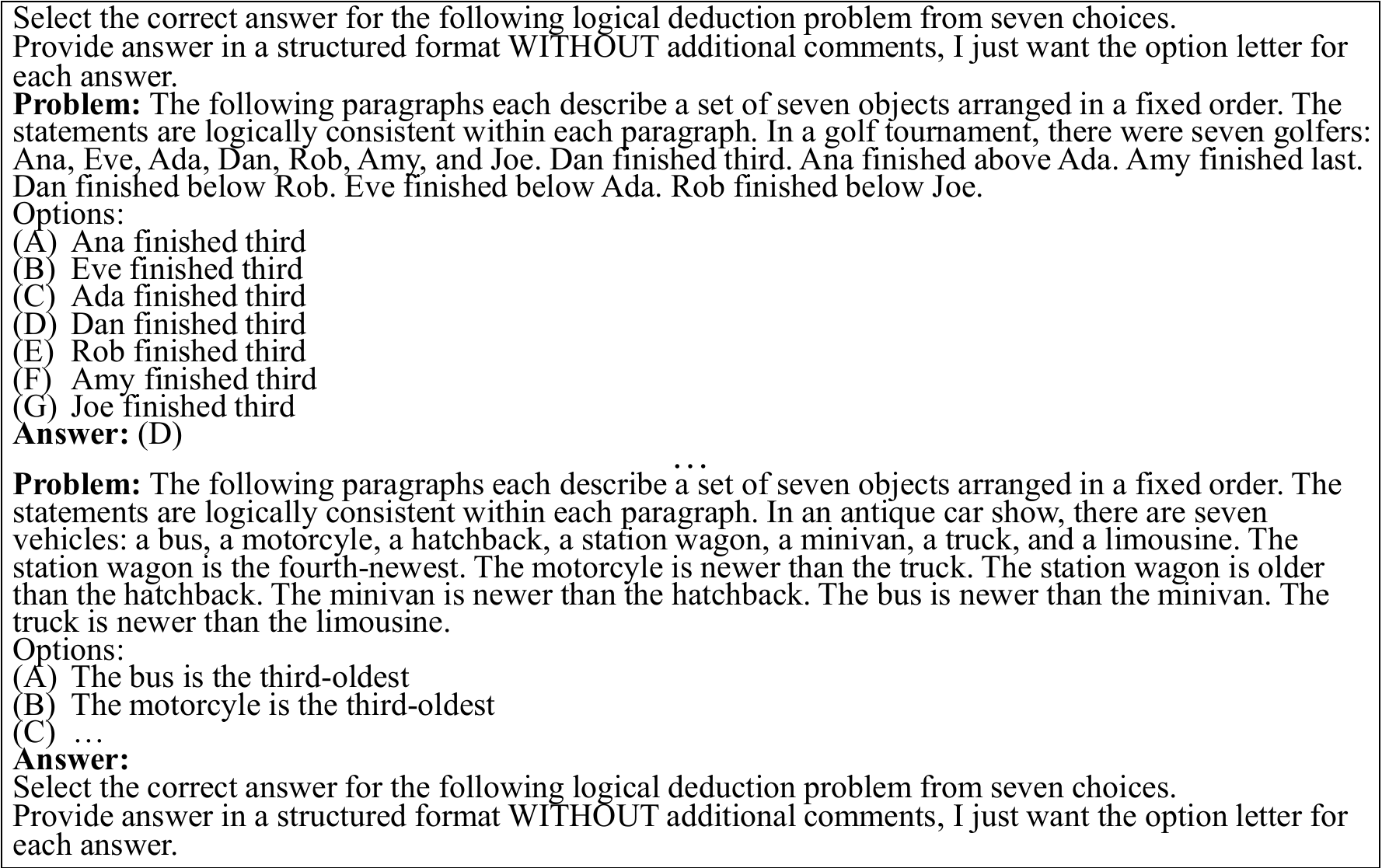} 
    \caption{Prompt template with a test input for logical deduction seven objects dataset.}
    \label{fig: ld7 prompts}
\end{figure*}

\end{document}